\documentclass[accepted]{uai2023}
\newif\ifinappendix

\usepackage[sort&compress, elide]{natbib}
\renewcommand{\cite}{\citep}

\usepackage{amsthm,amsmath,amssymb,amsfonts,bm,bbm,stmaryrd}

\def\eqref#1{equation~\ref{#1}}

\def\1{\bm{1}}

\DeclareMathAlphabet{\mathsfit}{\encodingdefault}{\sfdefault}{m}{sl}
\SetMathAlphabet{\mathsfit}{bold}{\encodingdefault}{\sfdefault}{bx}{n}

\newcommand{\E}{\mathbb{E}}

\newcommand{\R}{\mathbb{R}}

\newcommand{\range}[2]{\llbracket#1..#2\rrbracket}

\newcommand\independent{\protect\mathpalette{\protect\independenT}{\perp}}
\def\independenT#1#2{\mathrel{\rlap{$#1#2$}\mkern2mu{#1#2}}}

\newtheorem{theorem}{Theorem}[section]

\newtheorem{lemma}[theorem]{Lemma}

\newtheorem{definition}[theorem]{Definition}

\usepackage{microtype}
\usepackage{graphicx}
\usepackage{caption}
\usepackage{subcaption}
\usepackage{booktabs} %
\usepackage{paralist} %
\usepackage{enumitem}
\usepackage{multirow}
\usepackage{afterpage}
\usepackage{nicefrac}
\usepackage{tabularray}
\usepackage{graphbox}
\usepackage{wrapfig}
\usepackage{titletoc}
\usepackage{amssymb}%
\usepackage{pifont}%
\newcommand{\cmark}{\ding{51}}%
\newcommand{\xmark}{\ding{55}}%

\usepackage{tikz}
\usetikzlibrary{bayesnet}
\usetikzlibrary{arrows.meta}
\usetikzlibrary{arrows.meta,calc,decorations.markings,math,arrows.meta}
\usetikzlibrary{matrix}
\usepackage{pgfplots}
\pgfplotsset{compat=1.17}

\usepackage{hyperref}
\usepackage{xcolor, colortbl}
\usepackage[capitalize,noabbrev]{cleveref}
\usepackage{arydshln}

\newcommand{\OurApproach}{BISCUIT}
\newcommand{\OurApproachFull}{\textbf{B}inary \textbf{I}nteraction\textbf{s} for \textbf{C}a\textbf{u}sal \textbf{I}den\textbf{t}ifiability}

\newcommand{\robotstate}{regime variable}  %
\newcommand{\eg}{\emph{e.g.},}
\newcommand{\ie}{\emph{i.e.},}

\newcommand{\MLPI}{$\text{MLP}^{\hat{I}_i}_{\omega}$}
\newcommand{\MLPImath}{\text{MLP}^{\hat{I}_i}_{\omega}}
\newcommand{\MLPzi}{$\text{MLP}^{z_i}_{\omega}$}

\newcommand{\intvdiv}{\Delta}
\newcommand{\pseudoparagraph}[1]{\vspace{3mm}\textbf{#1}\hspace{3mm}}
\usepackage{comment}
\usetikzlibrary{backgrounds}
\usetikzlibrary{calc}

\definecolor{instantcolor}{HTML}{117733}
\definecolor{tempcolor}{HTML}{004488}
\definecolor{intvcolor}{HTML}{c83c04}
\definecolor{confcolor}{HTML}{886600}
\definecolor{obscolor}{HTML}{7a1f5c}
\definecolor{extraintvcolor}{HTML}{117733}
\definecolor{gaussianplotIone}{HTML}{BB5566}
\definecolor{gaussianplotItwo}{HTML}{004488}
\definecolor{gaussianplotIonetwo}{HTML}{BB9911}
\definecolor{gaussianplotObs}{HTML}{808080}
\definecolor{dark-green}{HTML}{00ba16}

\newcommand{\ourtitle}{\OurApproach{}: Causal Representation Learning from Binary Interactions}

\usepackage{titletoc}
\newcommand{\smallparagraph}[1]{\textbf{#1}\hspace{2mm}}
\begin{document}

\title{\ourtitle}

\author[1]{\href{mailto:p.lippe@uva.nl?Subject=UAI2023-BISCUIT}{Phillip Lippe}{}}
\author[2,3]{Sara Magliacane}
\author[2]{Sindy L\"owe}
\author[1]{Yuki M. Asano}
\author[4]{Taco Cohen}
\author[1]{Efstratios Gavves}

\affil[1]{%
    QUVA Lab, University of Amsterdam
}
\affil[2]{%
    AMLab, University of Amsterdam
}
\affil[3]{%
    MIT-IBM Watson AI Lab
}
\affil[4]{
    Qualcomm AI Research\thanks{
    \resizebox{0.9\linewidth}{!}{
    Qualcomm AI Research is an initiative of Qualcomm Technologies, Inc.
    }
    }\\
    Amsterdam, Netherlands
}

\maketitle

\begin{abstract}
Identifying the causal variables of an environment and how to intervene on them is of core value in applications such as robotics and embodied AI.
While an agent can commonly interact with the environment and may implicitly perturb the behavior of some of these causal variables, often the targets it affects remain unknown.
In this paper, we show that causal variables can still be identified for many common setups, e.g., additive Gaussian noise models, if the agent's interactions with a causal variable can be described by an unknown binary variable. This happens when each causal variable has two different mechanisms, e.g., an observational and an interventional one.
Using this identifiability result, we propose \OurApproach{}, a method for simultaneously learning causal variables and their corresponding binary interaction variables.
On three robotic-inspired datasets, \OurApproach{} accurately identifies causal variables and can even be scaled to complex, realistic environments for embodied AI.
Project page: \href{https://phlippe.github.io/BISCUIT/}{phlippe.github.io/BISCUIT/}.
\end{abstract}

\section{Introduction}
\label{sec:introduction}

\begin{figure}[t!]
    \centering
    \begin{tikzpicture}
        \node[inner sep=0pt, outer sep=0pt] (img) at (0,0)
        {\includegraphics[width=\linewidth]{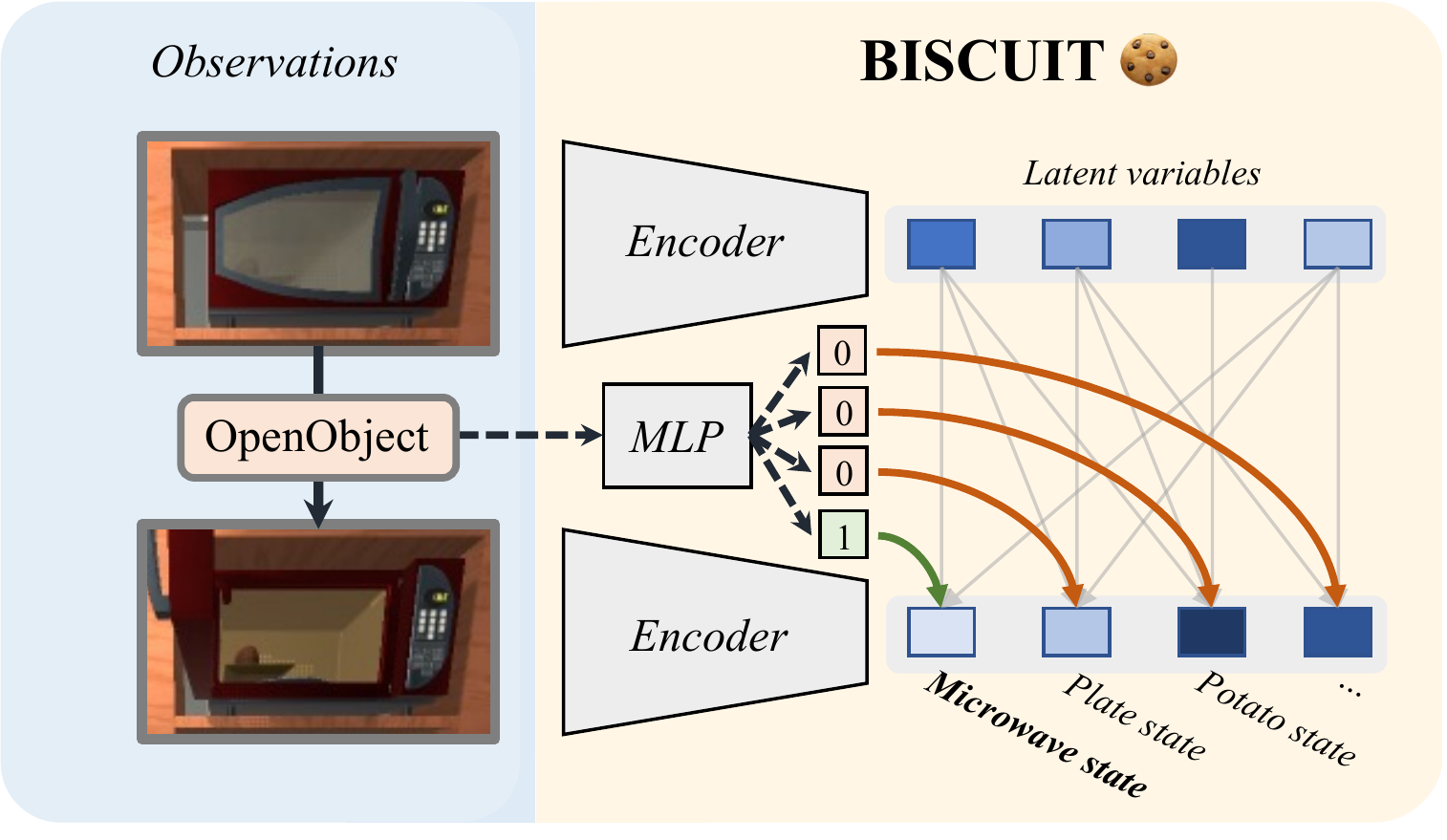}};
        \node[inner sep=0pt] (Xtn1) at (-0.455\linewidth, 0.13\linewidth) {\scriptsize $X^{t-1}$};
        \node[inner sep=0pt] (Xt) at (-0.455\linewidth, -0.14\linewidth) {\scriptsize $X^{t}$};
        \node[inner sep=0pt] (Rt) at (-0.455\linewidth, 0.0\linewidth) {\scriptsize $R^{t}$};
    \end{tikzpicture}
    \caption{
    \OurApproach{} identifies causal variables from images $X^{t-1}$ and $X^{t}$, by learning to encode an observable \robotstate{} $R^t$, \eg{} an action, as binary variables. 
    Conditioning each latent on one of these binary variables identifies causal variables in environments like iTHOR \cite{kolve2017ai}.
    }
    \label{fig:introduction_figure_1}
\end{figure}
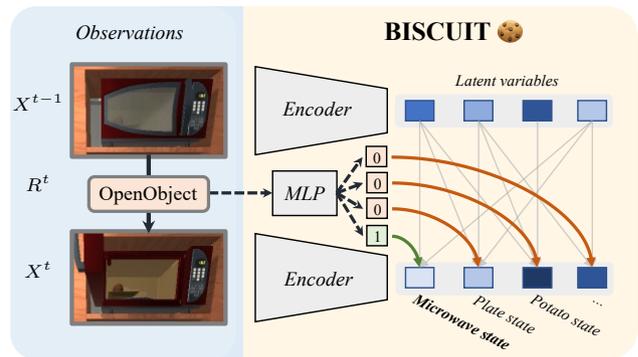

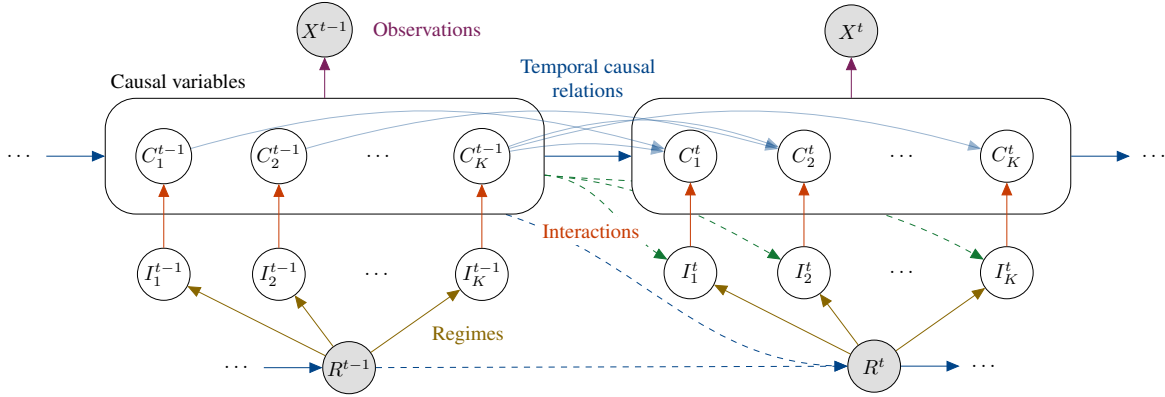
\begin{figure*}[t!]
    \centering
    \resizebox{0.9\textwidth}{!}{
        \begin{tikzpicture}
            \node (Ct) at (2.75,0) [draw,rounded corners=15pt,minimum width=7.5cm,minimum height=2cm,fill=white] {};
    	\node (Ct1) at (11.75,0) [draw,rounded corners=15pt,minimum width=7.5cm,minimum height=2cm,fill=white] {};
    	
    	\node[latent, minimum size=.9cm] (C1t) {$C^{t-1}_1$} ; %
    	\node[latent, minimum size=.9cm, right=of C1t] (C2t) {$C^{t-1}_2$} ; %
    	\node[const, right=of C2t] (dotdott) {$\cdots$} ; %
    	\node[latent, minimum size=.9cm, right=of C2t, xshift=1.5cm] (CKt) {$C^{t-1}_K$} ; %
    	\node[obs, above=of Ct, minimum size=.9cm, yshift=-.3cm] (Xt) {$X^{t-1}$} ; %
    	
    	\node[latent, xshift=9cm, minimum size=.9cm] (C1t1) {$C^{t}_1$} ; %
    	\node[latent, right=of C1t, xshift=9cm, minimum size=.9cm] (C2t1) {$C^{t}_2$} ; %
    	\node[const, right=of C2t1] (dotdott1) {$\cdots$} ; %
    	\node[latent, right=of C2t, xshift=10.5cm, minimum size=.9cm] (CKt1) {$C^{t}_K$} ; %
    	\node[obs, above=of Ct1, minimum size=.9cm, yshift=-.3cm] (Xt1) {$X^{t}$} ; %
    	
    	\node[latent, below=of C1t1, yshift=-.1cm, minimum size=.9cm] (I1t1) {$I^{t}_1$} ; %
    	\node[latent, below=of C2t1, yshift=-.1cm, minimum size=.9cm] (I2t1) {$I^{t}_2$} ; %
    	\node[const, right=of I2t1] (dotdotIt1) {$\cdots$} ; %
    	\node[latent, below=of CKt1, yshift=-.1cm, minimum size=.9cm] (IKt1) {$I^{t}_K$} ; %
    	\node[obs, below=of I2t1, xshift=1.2cm, minimum size=.9cm, yshift=.3cm] (Rt1) {$R^{t}$} ; %
    	
    	\node[latent, below=of C1t, yshift=-.1cm, minimum size=.9cm] (I1t) {$I^{t-1}_1$} ; %
    	\node[latent, below=of C2t, yshift=-.1cm, minimum size=.9cm] (I2t) {$I^{t-1}_2$} ; %
    	\node[const, right=of I2t] (dotdotIt) {$\cdots$} ; %
    	\node[latent, below=of CKt, yshift=-.1cm, minimum size=.9cm] (IKt) {$I^{t-1}_K$} ; %
    	\node[obs, below=of I2t, xshift=1.2cm, minimum size=.9cm, yshift=.3cm] (Rt) {$R^{t-1}$} ; %
    	
    	\node[const, left=of Ct] (dotdotCt) {$\cdots$\hspace{2mm}\phantom{}} ;
    	\node[const, left=of Rt] (dotdotRt) {$\cdots$\hspace{2mm}\phantom{}} ;
    	\node[const, right=of Ct1] (dotdotCt1) {\hspace{2mm}$\cdots$} ;
    	\node[const, right=of Rt1] (dotdotRt1) {\hspace{2mm}$\cdots$} ;
    	
    	\node[align=center,tempcolor] at (7.25,1.35) {Temporal causal\\relations};
    	\node[align=center,intvcolor,fill=white] at (7.3,-1.3) {Interactions};
    	\node[align=center,obscolor] at (4.5,2.2) {Observations};
    	\node[align=center,confcolor] at (5.2,-3.1) {Regimes};
    	\node[align=center,black] at (0.25,1.3) {Causal variables};
    	
    	\edge[obscolor]{Ct}{Xt} ;
    	\edge[obscolor]{Ct1}{Xt1} ;
     
            \begin{scope}[on background layer]
                \draw[->,dashed,extraintvcolor] (Ct) to [out=355, in=150] (I1t1);
                \draw[->,dashed,extraintvcolor] (Ct) to [out=355, in=150] (I2t1);
                \draw[->,dashed,extraintvcolor] (Ct) to [out=355, in=150] (IKt1);
                \draw[->,dashed,tempcolor] (Ct) to [out=342, in=180] (Rt1);
    	\end{scope}
    	
    	\edge[tempcolor,dashed]{Rt}{Rt1} ;
    	\edge[tempcolor]{Ct}{Ct1} ;
    	\edge[tempcolor]{dotdotCt}{Ct} ;
    	\edge[tempcolor]{dotdotRt}{Rt} ;
    	\edge[tempcolor]{Ct1}{dotdotCt1} ;
    	\edge[tempcolor]{Rt1}{dotdotRt1} ;
    	
    	\edge[intvcolor]{I1t1}{C1t1} ;
    	\edge[intvcolor]{I2t1}{C2t1} ;
    	\edge[intvcolor]{IKt1}{CKt1} ;
    	
    	\edge[confcolor]{Rt1}{I1t1} ;
    	\edge[confcolor]{Rt1}{I2t1} ;
    	\edge[confcolor]{Rt1}{IKt1} ;
    	
    	\edge[intvcolor]{I1t}{C1t} ;
    	\edge[intvcolor]{I2t}{C2t} ;
    	\edge[intvcolor]{IKt}{CKt} ;
    	
    	\edge[confcolor]{Rt}{I1t} ;
    	\edge[confcolor]{Rt}{I2t} ;
    	\edge[confcolor]{Rt}{IKt} ;
    
            \draw[->,tempcolor,opacity=0.4] (C1t) to [out=16, in=164] (C1t1);
            \draw[->,tempcolor,opacity=0.4] (C2t) to [out=16, in=164] (C2t1);
            \draw[->,tempcolor,opacity=0.4] (CKt) to [out=16, in=164] (CKt1);
            \draw[->,tempcolor,opacity=0.4] (CKt) to [out=10, in=170] (C1t1);
            \draw[->,tempcolor,opacity=0.4] (CKt) to [out=20, in=160] (C2t1);
        \end{tikzpicture}
    }
    \caption{A representation of our assumptions. Observed variables are shown in gray ($X^{\tau}$ and $R^{\tau}$) and latent variables in white. Optional causal edges are shown as dashed lines. A latent causal variable $C^{t}_i$ has as parents a subset of the causal factors at the \textcolor{tempcolor}{previous time step} $C^{t-1}=\{C_1^{t-1}, \dots, C_{K}^{t-1}\}$, and its latent \textcolor{intvcolor}{binary interaction variable} $I^{t}_i$. The interaction variables are determined by an observed \textcolor{confcolor}{\robotstate{}} $R^{t}$ and potentially by the variables from the \textcolor{extraintvcolor}{previous time step} $C^{t-1}$ (\eg{} in a collision). The \robotstate{} can be a dynamical process over time as well, for example, by depending on the previous time step. The \textcolor{obscolor}{observation} $X^{\tau}$ is a high-dimensional entangled representation of all causal variables $C^{\tau}$ at time step $\tau$.}
    \label{fig:main_causal_graph_full}
\end{figure*}

Learning a low-dimensional representation of an environment is a crucial step in many applications, \eg{} robotics \cite{lesort2018state}, embodied AI \cite{kolve2017ai} and reinforcement learning \cite{hafner2021mastering, traeuble2022the}.
A promising direction for learning robust and actionable representations is \textit{causal representation learning} \cite{schoelkopf2021towards}, which aims to identify the underlying causal variables and their relations in a given environment from high-dimensional observations, \eg{} images.
However, learning causal variables from high-dimensional observations is a considerable challenge and may not always be possible, since multiple underlying causal systems could generate the same data distribution \cite{hyvarinen1999nonlinear}.
To overcome this, several works make use of additional information, \eg{} by using counterfactual observations \cite{brehmer2022weakly, ahuja2022weakly, locatello2020weakly}, observed intervention targets \cite{lippe2022citris, lippe2022icitris}.
Alternatively, one can restrict the distributions of causal variables, \eg{} by considering environments with non-stationary noise \cite{yao2021learning, yao2022temporally, khemakhem2020variational} or sparse causal relations \cite{lachapelle2021disentanglement, lachapelle2022partial}.

In this paper, instead, we focus on interactive environments, where an agent can perform actions which may have an effect on the underlying causal variables.
We will assume that these interactions between the agent and the causal variables can be described by \emph{binary} variables, \ie{} that with the agent's actions, we can switch between two mechanisms, or distributions, of a causal variable, similarly to performing soft interventions.
Despite being binary, these interactions include a wide range of common scenarios, such as a robot pressing a button, opening/closing a door, or even colliding with a moving object and alternating its course.

In this setup, we prove that causal variables are identifiable if the agent interacts with each causal variable in a distinct pattern, \ie{} does not always interact with any two causal variables at the same time.
We show that for $K$ variables, we can in many cases fulfill this by having as few as $\lfloor \log_2 K \rfloor +2$ actions with sufficiently diverse effects, allowing identifiability even for a limited number of actions.
The binary nature of the interactions permits the identification of a wider class of causal models than previous work in a similar setup, including the common, challenging additive Gaussian noise model \cite{hyvarinen1999nonlinear}.

Based on these theoretical results, we propose \OurApproach{} (\OurApproachFull{}).
\OurApproach{} is a variational autoencoder \cite{kingma2014auto} which learns the causal variables and the agent's binary interactions with them in an unsupervised manner (see \cref{fig:introduction_figure_1}).
In experiments on robotic-inspired datasets, \OurApproach{} identifies the causal variables and outperforms previous methods.
Furthermore, we apply \OurApproach{} to the realistic 3D embodied AI environment iTHOR \cite{kolve2017ai}, and show that \OurApproach{} is able to generate realistic renderings of unseen causal states in a controlled manner.
This highlights the potential of causal representation learning in the challenging task of embodied AI.
In summary, our contributions are:
\begin{compactitem}
    \item We show that under mild assumptions, binary interactions with unknown targets identify the causal variables from high-dimensional observations over time.
    \item We propose \OurApproach{}, a causal representation learning framework that learns the causal variables and their binary interactions simultaneously.
    \item We empirically show that \OurApproach{} identifies both the causal variables and the interaction targets on three robotic-inspired causal representation learning benchmarks, and allows for controllable generations.
\end{compactitem}

\section{Preliminaries}
\label{sec:preliminaries}

In this paper, we consider a causal model $\mathcal{M}$ as visualized in \cref{fig:main_causal_graph_full}.
The model $\mathcal{M}$ consists of $K$ latent causal variables $C_1,...,C_K$ which interact with each other over time, like in a dynamic Bayesian Network (DBN) \cite{DBN, Murphy_DBN}.
In other words, at each time step $t$, we instantiate the causal variables as $C^t=\{C_1^t,...,C_K^t\}\in\mathcal{C}$, where $\mathcal{C}\subseteq\mathbb{R}^K$ is the domain.
In terms of the causal graph, each variable $C^t_i$ may be caused by a subset of variables in the previous time step $\{C_1^{t-1},...,C_K^{t-1}\}$.
For simplicity, we restrict the temporal causal graph to only model dependencies on the previous time step. 
Yet, as we show in \cref{sec:proof_longer_temporal_deps}, our results can be trivially extended to longer dependencies, \eg{} $(C^{t-2},C^{t-1})\to C^t$, since $C^{t-1}$ is only used for ensuring conditional independence.
As in DBNs, we consider the graph structure to be time-invariant.

Besides the intra-variable dynamics, we assume that the causal system is affected by a \robotstate{} $R^t$ with arbitrary domain $\mathcal{R}$, which can be continuous or discrete of arbitrary dimensionality.
This \robotstate{} can model any known external causes on the system, which, for instance, could be a robotic arm interacting with an environment.
For the causal graph, we assume that the effect of the \robotstate{} $R^t$ on a causal variable $C^t_i$ can be described by a latent \emph{binary interaction} variable $I^t_i\in\{0,1\}$.
This can be interpreted as each causal variable having two mechanisms/distributions, \eg{} an observational and an interventional mechanism, which has similarly been assumed in previous work \cite{brehmer2022weakly, lippe2022citris, lippe2022icitris}. 
Thereby, the role of the interaction variable $I^t_i$ is to select the mechanism, \ie{} observational or interventional, at time step $t$.
For example, a collision between an agent and an object is an interaction that switches the dynamics of the object from its natural course to a perturbed one.
In this paper, we consider the interaction variable $I^t_i$ to be an unknown function of the \robotstate{} and the previous causal variables, \ie{} $I^t_i=f_i(R^t,C^{t-1})$.
The dependency on the previous time step allows us to model interactions that only occur in certain states of the system, \eg{} a collision between an agent (modeled by $R^t$) with an object with position $C_i^{t-1}$ will only happen for certain positions of the agent and the object.

We consider the causal graph of \cref{fig:main_causal_graph_full} to be causally sufficient, \ie{} we assume there are no other unobserved confounders except the ones we have described in the previous paragraphs and represented in the Figure, and that the causal variables within the same time step are independent of each other, conditioned on the previous time step and their interaction variables.
We summarize the dynamics as $p(C^t|C^{t-1},R^t)=\prod_{i=1}^K p(C^t_i|C^{t-1},I^t_i)$.
Although $C^t_i$ only depends on a subset of $C^{t-1}$, w.l.o.g. we model it as depending on all causal variables from the previous time step.

In causal representation learning, the task is to identify causal variables from an entangled, potentially higher-dimensional representation, \eg{} an image.
We consider an injective observation function $g$, mapping the causal variables $C^t$ to an observation $X^t=g(C^t)$.
Following \citet{klindt2021towards,yao2021learning}, we assume $g$ to be defined everywhere for $C^t$ and differentiable almost everywhere.
In our setting, once we identify the causal variables, the causal graph can be trivially learned by testing for conditional independence, since the causal graph is limited to edges following the temporal dimension, \ie{} from $C^{t-1}$ to $C^t$.
We provide further details on the graph discovery and an example on learned causal variables in \cref{sec:proof_causal_graph}. 

\section{Identifying Causal Variables}
\label{sec:theory}

Our goal in this paper is to identify the causal variables $C_1,...,C_K$ of a causal system from sequences of observations $(X^t,R^t)$.
We first define the identifiability class that we consider.
We then provide an intuition on how binary interactions enable identifiability, before presenting our two identifiability results.
The practical algorithm based on these results, \OurApproach{}, is presented in \cref{sec:method}.

\subsection{Identifiability Class and Definitions}

Intuitively, we seek to estimate an observation function $\hat{g}$, which maps a latent space $\hat{\mathcal{C}}$ to observations $X$, and models each true causal variable $C_i$ in a different dimension of the latent space $\hat{\mathcal{C}}$. 
This observation function should be equivalent to the true observation function $g$, up to permuting and transforming the variables individually, \eg{} through scaling. 
Several previous works \cite{yao2021learning, yao2022temporally, lachapelle2021disentanglement, khemakhem2020variational} have considered equivalent identifiability classes, which we define as:

\begin{definition}
    \label{def:identifiability}
    Consider a model $\mathcal{M}=\langle g,f,\omega,\mathcal{C} \rangle$ with an injective function $g(C)=X$ with $C\in\mathcal{C}$ and a latent distribution $p_{\omega}(C^{t}|C^{t-1},R^{t})$, parameterized by $\omega$ and defined:
    \begin{equation*}
        p_{\omega}(C^{t}|C^{t-1},R^{t})=\prod_{i=1}^{K} p_{\omega,i}\big(C^t_i|C^{t-1},f_i(R^t,C^{t-1})\big),
    \end{equation*}
    where $f_i:\mathcal{R}\times\mathcal{C}\to\{0,1\}$ outputs a binary variable for the variable $C^t_i$. 
    We call $\mathcal{M}$ identifiable iff for any other model $\mathcal{\widetilde{M}}=\langle \tilde{g},\tilde{f},\tilde{\omega},\tilde{\mathcal{C}} \rangle$ with the same observational distribution $p(X^t|X^{t-1},R^t)$, $g$ and $\tilde{g}$ are equivalent up to a component-wise invertible transformation $T$ and a permutation $\pi$:
    \begin{equation*}
        p_{\mathcal{M}}(X^t|X^{t-1}\!,R^t) = p_{\mathcal{\widetilde{M}}}(X^t|X^{t-1}\!,R^t) \Rightarrow g = \tilde{g} \circ T \circ \pi
    \end{equation*}
\end{definition}

To achieve this identifiability, we rely on the interaction variables $I^t_i$ being binary and having \textit{distinct interaction patterns}, a weaker form of faithfulness on the interaction variables.
Intuitively, we do not allow that any two causal variables to have identical interaction variables $I^t_i,I^t_j$ across the whole dataset, \ie{} being always interacted with at the same time.
Similarly, if all $I^t_i$ are always zero ($\forall t,i\colon I^t_i=0$), then we fall back into the well-known unidentifiable setting of non-linear ICA \cite{hyvarinen1999nonlinear}.
Since interaction variables can also be functions of the previous state, we additionally assume that for all possible previous states, the interaction variables cannot be deterministic functions of any other. 
Thus, we assume that all causal variables have \textit{distinct interaction patterns}, which we formally define as:

\begin{definition}
    \label{def:theory_distinct_intvs}
    A causal variable $C_i$ in $\mathcal{M}=\langle g,f,\omega,\mathcal{C}\rangle$ has a \textbf{distinct interaction pattern} if for all values of $C^{t-1}$, its interaction variable $I^t_i=f_i(R^t,C^{t-1})$ is not a deterministic function $b\!:\!\{0,1\}\!\to\!\{0,1\}$ of any other $I^t_j$:
    \begin{equation*}
        \label{eq:theory_assum_distinct_intvs}
        \forall C^{t-1}\!,\forall j\!\neq\!i, \nexists b, \forall R^t\colon f_i(R^t,C^{t-1})=b(f_j(R^t,C^{t-1})).
    \end{equation*}
\end{definition}

This assumption generalizes the intervention setup of \citet{lippe2022intervention}, which has a similar condition on its binary intervention variables, but assumed them to be independent of the previous time step.
This implies that we can create a distinct interaction pattern for each of the $K$ causal variables by having as few as $\lfloor \log_2 K\rfloor + 2$ different values for $R^t$, if the interaction variables are independent of $C^{t-1}$.
In contrast, other methods in similar setups that also exploit an external, temporally independent, observed variable \citep{yao2022temporally, yao2021learning, khemakhem2020variational} require the number of regimes to scale linearly with the number of causal variables.
If the interaction variables depend on $C^{t-1}$, the lower bound of the number of different values for $R^t$ depends on the causal model $\mathcal{M}$, more specifically its interaction functions $f_i$. Concretely, the lower bound for a causal model $\mathcal{M}$ is the smallest set of values of $R^t$ that ensure different interaction patterns for all $C^{t-1}$ in $\mathcal{M}$. In the worst case, each $C^{t-1}$ may require different values of $R^t$ to fulfill the condition of \cref{def:theory_distinct_intvs}, such that $R^t$ would need to be of the same domain as $C^{t-1}$ (for instance being continuous). At the same time, for models $\mathcal{M}$ in which the condition of \cref{def:theory_distinct_intvs} can be fulfilled by the same values of $R^t$ for all $C^{t-1}$, we again recover the lower bound of $\lfloor \log_2 K + 2 \rfloor$ different values of $R^t$.

\subsection{Intuition: Additive Gaussian Noise}
\label{sec:theory_intuition}

\begin{figure}[t!]
    \centering
    \begin{tabular}{cc}
        \begin{tikzpicture}
            \begin{axis}[
              width=0.6\columnwidth,
              height=0.6\columnwidth,
              axis lines=middle,
              axis line style={draw=none},
              tick style={draw=none},
              xmin=-0.99,xmax=1.49,ymin=-0.99,ymax=1.49,
              xtick distance=0.5,
              ytick distance=0.5,
              yticklabels={,,},
              xticklabels={,,},
              title={True variables},
              grid=major,
              grid style={thin,densely dotted,black!20}
              ]
              \draw[black,-Stealth] (-1.0,0.0) -- (1.5,0.0);
              \draw[black,-Stealth] (0.0,-1.0) -- (0.0,1.5);
              
              \coordinate (Obs) at (axis cs: 0.0,0.0);
              \coordinate (I1) at (axis cs: 1.0,0.0);
              \coordinate (I2) at (axis cs: 0.0,1.0);
              \coordinate (I1I2) at (axis cs: 1.0,1.0);
              \draw[thick,color=gaussianplotIone,->](Obs)-- node[yshift=-3mm,xshift=0mm]{$I_1$}(I1);
              \draw[thick,color=gaussianplotItwo,->](Obs)--  node[yshift=0mm,xshift=-3mm]{$I_2$}(I2);
              \draw[color=gaussianplotIonetwo,dashed,opacity=0.5](I1)--(I1I2);
              \draw[color=gaussianplotIonetwo,dashed,opacity=0.5](I2)--(I1I2);
              \draw[thick,color=gaussianplotIonetwo,->](Obs)-- node[yshift=-2mm,xshift=2mm] {$I_{12}$}(I1I2);
              
              \draw[line width=0.7mm,opacity=0.8,gaussianplotItwo] (axis cs: -0.07,1.0) -- (axis cs: 0.0,1.0);
              \draw[line width=0.7mm,opacity=0.8,gaussianplotIonetwo] (axis cs: 0.0,1.0) -- (axis cs: 0.07,1.0);
              \draw[line width=0.7mm,opacity=0.8,gaussianplotObs] (axis cs: 0.07,0.0) -- (axis cs: -0.07,0.0);
            
              \draw[line width=0.7mm,opacity=0.8,gaussianplotIone] (axis cs: 1.0,-0.07) -- (axis cs: 1.0,0.0);
              \draw[line width=0.7mm,opacity=0.8,gaussianplotIonetwo] (axis cs: 1.0,0.0) -- (axis cs: 1.0,0.07);
              \draw[line width=0.7mm,opacity=0.8,gaussianplotObs] (axis cs: 0.0,0.07) -- (axis cs: 0.0,-0.07);
            \end{axis}
            \node (C2) at (1.7,3.2) {$C_2$};
            \node (C1) at (3.2,1.7) {$C_1$};
        \end{tikzpicture}
        & 
        \begin{tikzpicture}
            \begin{axis}[
              anchor=origin,
              width=0.6\columnwidth,
              height=0.6\columnwidth,
              axis lines=middle,
              axis line style={draw=none},
              tick style={draw=none},
              xmin=-1.0,xmax=1.5,ymin=-1.0,ymax=1.5,
              xtick distance=0.5,
              ytick distance=0.5,
              yticklabels={,,},
              xticklabels={,,},
              title={Rotated variables},
              grid=major,
              grid style={draw=none}
              ]
              \foreach \x in {-3,-2,-1,0,1,2,3,4}
                {   
                  \edef\temp{\noexpand
                  \draw[thin,densely dotted,black!20] (axis cs: \x*0.5*0.707+10,\x*0.5*0.707-10) -- (axis cs: \x*0.5*0.707-10,\x*0.5*0.707+10);
                  }
                  \temp
                  \edef\temp{\noexpand
                  \draw[thin,densely dotted,black!20] (axis cs: -\x*0.5*0.707+10,\x*0.5*0.707+10) -- (axis cs: -\x*0.5*0.707-10,\x*0.5*0.707-10);
                  }
                  \temp
                }
              
              \draw[black,-Stealth] (-1.0,1.0) -- (1.0,-1.0);
              \draw[black,-Stealth] (-1.0,-1.0) -- (1.5,1.5);
              
              \coordinate (Obs) at (axis cs: 0.0,0.0);
              \coordinate (I1) at (axis cs: 1.0,0.0);
              \coordinate (I2) at (axis cs: 0.0,1.0);
              \coordinate (I1I2) at (axis cs: 1.0,1.0);
              \draw[thick,color=gaussianplotIone,->](Obs)-- node[yshift=-3mm,xshift=0mm]{$I_1$}(I1);
              \draw[thick,color=gaussianplotItwo,->](Obs)--  node[yshift=0mm,xshift=-3mm]{$I_2$}(I2);
              \draw[color=gaussianplotItwo,dashed,opacity=0.5](I2)--(axis cs: 0.5,0.5);
              \draw[color=gaussianplotItwo,dashed,opacity=0.5](I2)--(axis cs: -0.5,0.5);
              \draw[color=gaussianplotIone,dashed,opacity=0.5](I1)--(axis cs: 0.5,0.5);
              \draw[color=gaussianplotIone,dashed,opacity=0.5](I1)--(axis cs: 0.5,-0.5);
              \draw[thick,color=gaussianplotIonetwo,->](Obs)-- node[yshift=2mm,xshift=7mm] {$I_{12}$}(I1I2);
    
              \draw[line width=0.7mm,opacity=0.8,gaussianplotIone] (axis cs: 0.5+0.05,-0.5+0.05) -- (axis cs: 0.5-0.05,-0.5-0.05);
              \draw[line width=0.7mm,opacity=0.8,gaussianplotItwo] (axis cs: -0.5+0.05,0.5+0.05) -- (axis cs: -0.5-0.05,0.5-0.05);
              \draw[line width=0.7mm,opacity=0.8,gaussianplotObs] (axis cs: 0.05,0.05) -- (axis cs: -0.05,-0.05);
              
              \draw[line width=0.7mm,opacity=0.8,gaussianplotIone] (axis cs: 0.5+0.05,0.5-0.05) -- (axis cs: 0.5,0.5);
              \draw[line width=0.7mm,opacity=0.8,gaussianplotItwo] (axis cs: 0.5,0.5) -- (axis cs: 0.5-0.05,0.5+0.05);
              \draw[line width=0.7mm,opacity=0.8,gaussianplotObs] (axis cs: -0.05,0.05) -- (axis cs: 0.05,-0.05);
              \draw[line width=0.7mm,opacity=0.8,gaussianplotIonetwo] (axis cs: 1.0+0.05,1.0-0.05) -- (axis cs: 1.0-0.05,1.0+0.05);
            \end{axis}
            \node (Chat2) at (2.1,1.7) {$\hat{C}_2$};
            \node (Chat1) at (1.5,-1.0) {$\hat{C}_1$};
        \end{tikzpicture}
    \end{tabular}
    \caption{
    Binary interactions identify the additive Gaussian noise model in \cref{eq:theory_additive_gaussian_model}. 
    The plots show the change of the mean for each variable for a fixed $C^{t-1}$ under interactions affecting only one variable (\textcolor{gaussianplotIone}{$I_1=1$} or \textcolor{gaussianplotItwo}{$I_2=1$}), and under joint interactions \textcolor{gaussianplotIonetwo}{$I_{12}$} (\textcolor{gaussianplotIonetwo}{$I_{1}=I_{2}=1$}). 
    Ticks on axes show mean values for each variable, where the mean for $(I_1=0, I_2=0)$ lies at the origin. The colors of the ticks match the interaction color. 
    \textbf{Left}: true causal variables $C_1$ and $C_2$. 
    Each variable has two possible means after any of the possible interactions. 
    The effect of the interactions can be described by a binary variable per axis. 
    \textbf{Right}: a rotated representation. 
    Both $\hat{C}_1$ and $\hat{C}_2$ have three possible means, which cannot be described by a binary variable per axis anymore.
    }
    \label{fig:main_gaussian_example}
\end{figure}
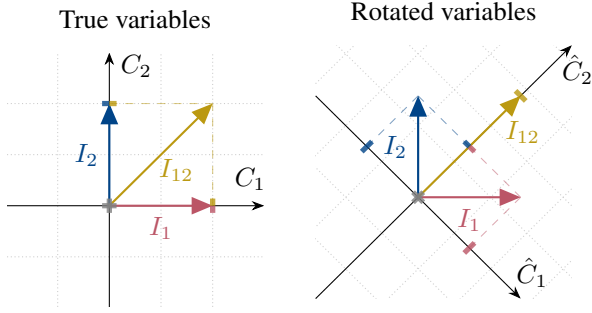

We first provide some intuition on how binary interactions, \ie{} knowing that each variable has exactly two potential mechanisms, enable identifiability, even when we do not know which variables are interacted with at each time step.
We take as an example an additive Gaussian noise model with two variables $C_1,C_2$, each described by the equation:
\begin{equation}
    \label{eq:theory_additive_gaussian_model}
    C^{t}_i = \mu_i(C^{t-1}, I^{t}_i) + \epsilon_i, \hspace{3mm} \epsilon_i \sim \mathcal{N}(0, \sigma^2),
\end{equation}
where $\epsilon_i$ is additive noise with variance $\sigma^2$, and $\mu_i$ a function for the mean with $\mu_i(C^{t-1}, I^{t}_i=0)\neq \mu_i(C^{t-1}, I^{t}_i=1)$.
Due to the rotational invariance of Gaussians, the true causal variables $C_1,C_2$ and their rotated counterparts $\hat{C}_1, \hat{C}_2$ model the same distribution with the same factorization: $\prod_{i=1}^2 p_i(C^{t}_i|C^{t-1},R^{t}) = \prod_{i=1}^2 \hat{p}_i(\hat{C}^{t}_i|\hat{C}^{t-1},R^{t})$.
This property makes the model unidentifiable in many cases \cite{yao2022temporally, khemakhem2020variational, hyvaerinen2019nonlinear, lachapelle2021disentanglement}. 
However, when the effect of the \robotstate{} on a causal variable $C_i$ can be described by a \textit{binary} variable, \ie{} $I_i\in\{0,1\}$, the two representations become distinguishable.
In \cref{fig:main_gaussian_example}, we visualize the two representations by showing the means of the different variables under interactions, which we detail in \cref{sec:proof_additive_gaussian_noise} and provide intuition here. %
For the original representation $C_1,C_2$, each variable's mean takes on only two different values for any $R^t$. For example, for \robotstate{}s where $I_1=0$, the variable $C_1$ takes a mean that is in the center of the coordinate system. Similarly, when $I_1=1$, the variable $C_1$ will take a mean that is represented as a pink (for $I_1=1, I_2=0$) or yellow tick (for $I_1=1, I_2=1$).  
In contrast, for the rotated variables, both $\hat{C}_1$ and $\hat{C}_2$ have three different means depending on the interactions, making it impossible to model them with individual binary variables.
Intuitively, the only alternative representations to $C_1,C_2$ which can be described by binary variables are permutations and/or element-wise transformations, effectively identifying the causal variables according to our identifiability class.

\subsection{Identifiability Result}
\label{sec:theory_identifiability_result}

When extending this intuition to more than two variables, we find that systems may become unidentifiable when the two distributions of each causal variable, \ie{} interacted and not interacted, always differ in the same manner.
Formally, we denote the log-likelihood difference between the two distributions of a causal variable $C^t_i$ as $\intvdiv(C^{t}_i|C^{t-1}) :=\log p(C^{t}_i|C^{t-1},I^t_i=1) - \log p(C^{t}_i|C^{t-1},I^t_i=0)$.
If this difference or its derivative w.r.t. $C^t_i$ is constant for all values of $C^t_i$, the effect of the interactions could be potentially modeled in fewer than $K$ dimensions, giving rise to models that do not identify the causal model $\mathcal{M}$.

To prevent this, we consider two possible setups for ensuring sufficient variability of $\intvdiv(C^{t}_i|C^{t-1})$: \emph{dynamics variability}, and \emph{time variability}.
We present our identifiability result below and provide the proofs in \cref{sec:appendix_proofs}.

\begin{theorem}
    \label{theo:main_theorem_no_gaus}
    \label{theo:main_theorem_time_var}
    \label{theo:main_theorem}
    An estimated model $\mathcal{\widehat{M}}=\langle \hat{g},\hat{f},\hat{\omega},\hat{\mathcal{C}} \rangle$ identifies the true causal model $\mathcal{M}=\langle g,f,\omega,\mathcal{C} \rangle$ if:
    \begin{compactenum}
        \item (\textbf{Observations}) $\mathcal{\widehat{M}}$ and $\mathcal{M}$ model the same likelihood:
        $$
            p_{\widehat{\mathcal{M}}}(X^t|X^{t-1},R^t)=p_{\mathcal{M}}(X^t|X^{t-1},R^t);
        $$
        \item (\textbf{Distinct Interaction Patterns}) Each variable $C_i$ in $\mathcal{M}$ has a distinct interaction pattern (Definition \ref{def:theory_distinct_intvs});
    \end{compactenum}
    and one of the following two conditions holds for $\mathcal{M}$:
    \begin{compactenum}
        \item[A.] (\textbf{Dynamics Variability}) Each variable's log-likelihood difference is twice differentiable and not always zero:
        $$
            \forall C^t_i ,\exists C^{t-1} \colon \frac{\partial^2 \intvdiv(C^{t}_i|C^{t-1})}{\partial (C^t_i)^2} \neq 0;
        $$
        \item[B.] (\textbf{Time Variability}) For any $C^{t}\in\mathcal{C}$, there exist $K+1$ different values of $C^{t-1}$ denoted with $c^1,...,c^{K+1}\in\mathcal{C}$, for which the vectors $v_1,...,v_{K}\in\R^{K+1}$ with
        \ifinappendix
        $$
            v_{i} = \begin{bmatrix}
                \frac{\partial \intvdiv\left(C^{t}_i|C^{t-1}=c^1\right)}{\partial C^{t}_i} & 
                \frac{\partial \intvdiv\left(C^{t}_i|C^{t-1}=c^2\right)}{\partial C^{t}_i} & 
                \cdots &
                \frac{\partial \intvdiv\left(C^{t}_i|C^{t-1}=c^{K+1}\right)}{\partial C^{t}_i} \\ 
            \end{bmatrix}^T \in \mathbb{R}^{K+1}
        $$
        \else
        $$
            v_{i} = \begin{bmatrix}
                \frac{\partial \intvdiv\left(C^{t}_i|C^{t-1}=c^1\right)}{\partial C^{t}_i} & 
                \cdots &
                \frac{\partial \intvdiv\left(C^{t}_i|C^{t-1}=c^{K+1}\right)}{\partial C^{t}_i} \\ 
            \end{bmatrix}^T
        $$
        \fi
        are linearly independent.
    \end{compactenum}
\end{theorem}

Intuitively, \cref{theo:main_theorem} states that we can identify a causal model $\mathcal{M}$ by maximum likelihood optimization, if we have distinct interaction patterns (Definition~\ref{def:theory_distinct_intvs}) and $\intvdiv(C^{t}_i|C^{t-1})$ varies sufficiently, either in dynamics or in time.
\emph{Dynamics variability} can be achieved by the difference $\intvdiv(C^{t}_i|C^{t-1})$ being non-linear for all causal variables.
This assumption is common in previous ICA-based works \cite{yao2021learning, yao2022temporally, hyvaerinen2019nonlinear} and, for instance, allows for Gaussian distributions with variable standard deviations.
While allowing for a variety of distributions, it excludes additive Gaussian noise models.
We can include this challenging setup by considering the \emph{time variability} assumption, which states that the effect of the interaction depends on the previous time step, and must do so differently for each variable.
As an example, consider a dynamical system with several moving objects, where an interaction is a collision with a robotic arm.
The time variability condition is commonly fulfilled by the fact that the trajectory of each object depends on its own velocity and position.

In comparison to previous work, our identifiability results cover a larger class of causal models by exploiting the binary nature of the interaction variables. We provide a detailed comparison in \cref{sec:proof_relation_to_prev_results}. 
In short, closest to our setup, \citet{khemakhem2020variational} and \citet{yao2021learning} require a stronger form of both our dynamics and time variability assumptions, excluding common models like additive Gaussian noise models. 
\citet{lachapelle2021disentanglement} requires that no two causal variables share the same parents, limiting the allowed temporal graph structures. 
Meanwhile, our identifiability results allow for arbitrary temporal causal graphs.
Further, the two conditions of \cref{theo:main_theorem} complement each other well by covering different underlying distributions for the same general setup.
Thus, in the next section, we can develop one joint learning algorithm for identifying the causal variables based on both conditions in \cref{theo:main_theorem}.

\section{\OurApproach{}}
\label{sec:method}

Using the results of \cref{sec:theory}, we propose \OurApproach{} (\OurApproachFull{}), a neural-network based approach to identify causal variables and their interaction variables.
In short, \OurApproach{} is a variational autoencoder (VAE) \cite{kingma2014auto}, which aims at modeling each of the causal variables $C_1,...,C_K$ in a separate latent dimension by enforcing the latent structure of \cref{fig:main_causal_graph_full}.
We first give an overview of \OurApproach{} and then detail the design choices for the model prior.

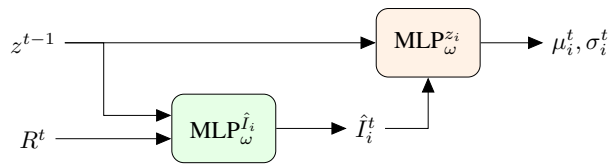
\begin{figure}[t!]
    \centering
    \resizebox{\columnwidth}{!}{
        \begin{tikzpicture}
            \node (zt) at (0,0) {$z^{t-1}$};
            \node (Rt1) at (0,-1.4) {$R^{t}$};
            \node (MLPI) at (2.75,-1.25) [draw,minimum width=1.5cm,minimum height=1cm,fill=green!10,rounded corners=5pt] {\MLPI{}};
            \node (Ihat) at (4.8,-1.25) {$\hat{I}^{t}_i$};
            \node (MLPz) at (5.7,0.0) [draw,minimum width=1.5cm,minimum height=1cm,fill=orange!10,rounded corners=5pt] {\MLPzi{}};
            \node (zt1) at (7.9,0) [align=center] {$\mu^{t}_i,\sigma^{t}_i$};
        
            \coordinate (MLPIinter) at ($(MLPI.west)!0.5!(Rt1.east)$);
            \coordinate (MLPIztconnect) at ($(MLPIinter)!0.2!(zt.east)$);
            \draw[->] (zt) -| (MLPIztconnect) -- (MLPI.west |- MLPIztconnect);
            \draw[->] (Rt1) -- (MLPI.west |- Rt1);
            \draw[->] (MLPI) -- (Ihat);
            \draw[->] (zt) -- (MLPz);
            \draw[->] (Ihat) -| (MLPz.south |- Ihat.east) -- (MLPz);
            \draw[->] (MLPz) -- (zt1);
        \end{tikzpicture}
    }
    \caption{
    The prior structure of \OurApproach{}. Based on the previous latents $z^{t-1}$ and the observed \robotstate{} $R^{t}$, the \MLPI{} predicts the interaction variable $\hat{I}_i^{t}$. Then, \MLPzi{} outputs the distribution for the next time step $p(z^{t}_i|z^{t-1},\hat{I}_i^{t})$, which can be parameterized by, \eg{} a mean $\mu^{t}_i$ and std $\sigma^{t}_i$. 
    }
    \label{fig:model_prior}
\end{figure}

\subsection{Overview}
\label{sec:method_overview}

\OurApproach{} consists of three main elements: the encoder $q_{\phi}$, the decoder $p_{\theta}$, and the prior $p_{\omega}$.
The decoder and encoder implement the observation function $g$ and its inverse $g^{-1}$ (Definition~\ref{def:identifiability}), respectively, and act as a map between observations $x^t$ and a lower-dimensional latent space $z^t \in\mathbb{R}^{M}$, in which we learn the causal variables $C_1^t,...,C_K^t$.
The goal of the model is to learn each causal variable $C_i^t$ in a different latent dimension, \eg{} $z^t_j$, effectively separating and hence identifying the causal variables according to \cref{def:identifiability}.
Thus, we need the latent space to have at least $K$ dimensions.
In practice, since the number of causal variables is not known a priori, we commonly overestimate the latent dimensionality, \ie{} $M\gg K$.
Still, we expect the model to only use $K$ dimensions actively, with the redundant dimensions not containing any information after training.

On this latent space, the prior $p_{\omega}$ learns a distribution that follows the structure in Definition~\ref{def:identifiability}, modeling the dynamics in the latent space.
As an objective, we maximize the data likelihood of observation triplets $\{X^t,X^{t-1},R^t\}$ from the true causal model, as stated in \cref{theo:main_theorem_no_gaus,theo:main_theorem_time_var}.
The loss function for \OurApproach{} is:
\begin{equation}
    \label{eq:method_vae_objective}
    \begin{split}
        &\mathcal{L}^{t} = -\E_{q_{\phi}\left(z^{t}|x^{t}\right)}\left[\log p_{\theta}(x^{t}|z^{t})\right] + \\ 
        &\E_{q_{\phi}\left(z^{t-1}|x^{t-1}\right)}\left[\text{KL}\left(q_{\phi}(z^{t}|x^{t})||p_{\omega}(z^{t}|z^{t-1},R^{t})\right)\right]
    \end{split}
\end{equation}
with learnable parameter sets $\phi$ (encoder), $\theta$ (decoder), and $\omega$ (prior), and KL being the Kullback-Leibler divergence.

For visually complex datasets, the VAE commonly has to perform a trade-off between reconstruction quality and prior modeling, which may cause poorer identification of the causal variables.
To circumvent that, we follow \citet{lippe2022citris} by separating the reconstruction and prior modeling stage by training an autoencoder and a normalizing flow \cite{rezende2015variational} in separate stages.
In this setup, an autoencoder is first trained to map the observations $x^t$ into a lower-dimensional space.
Afterward, we learn a normalizing flow on the autoencoder's representations to transform them into the desired causal representation, using the same prior structure as in the VAE.
In experiments, we refer to this approach as \OurApproach{}-NF, and the previously described VAE-based approach as \OurApproach{}-VAE.

\subsection{Model Prior}
\label{sec:method_prior}
Our prior follows the distribution structure of Definition~\ref{def:identifiability}, which has two elements per latent variable: a function to model the binary interaction variable, and a conditional distribution.
We integrate this into \OurApproach{}'s prior by learning both via multi-layer perceptrons (MLPs):
\begin{equation}
    p_{\omega}(z^{t}|z^{t-1}\!\!,R^{t})\!=\!\prod_{i=1}^{M} p_{\omega,i}\big(z_i^{t}|z^{t-1}\!\!,\MLPImath{}(R^{t}, z^{t-1})\big).\!
\end{equation}
Here, \MLPI{} is an MLP that maps the \robotstate{} $R^t$ and the latents of the previous time step $z^{t-1}$ to a binary output $\hat{I}^{t}_i$, as shown in \cref{fig:model_prior} .
This MLP aims to learn the interaction variable for the latent variable $z_i$, simply by optimizing \cref{eq:method_vae_objective}.
The variable $\hat{I}^t_i$ is then used as input for predicting the distribution over $z^{t}_i$.
For simplicity, we model $p_{\omega,i}$ as a Gaussian distribution, which is parameterized by one MLP per variable, \MLPzi{}, predicting the mean and standard deviation.
To allow for more complex distributions, $p_{\omega,i}$ can alternatively be modeled by a conditional normalizing flow \cite{winkler2019learning}.

In early experiments, we found that enforcing $\hat{I}^{t}_i$ to be a binary variable and backpropagating through it with the straight-through estimator \cite{bengio2013estimating} leads to suboptimal performances.
Instead, we model $\hat{I}^{t}_i$ as a continuous variable during training by using a temperature-scaled $\tanh{}$ as the output activation function of \MLPI{}.
By gradually decreasing the temperature, we bring the activation function closer to a discrete step function towards the end of training.

\section{Related Work}
\label{sec:related_work}

\smallparagraph{Causal Discovery from Unknown Targets}
Learning (equivalence classes of) causal graphs from observational and interventional data, even with unknown intervention targets, is a common setting in causal discovery \cite{jaber2020causal, brouillard2020differentiable, squires2020permutation, mooij2020joint}. 
In recent work, this is even extended to the case in which we have unknown mixtures of interventional data \cite{kumar2021disentangling,faria2022differentiable,mian2023information}, which for example can happen if the \robotstate{} is not observed.
In our paper, we assume that we observe the \robotstate{} and then reconstruct the latent interaction variables, which resemble the observed context variables by \citet{mooij2020joint}. 
Moreover, our work is on a different task, causal representation learning, in which we try to learn the causal variables from high-dimensional data.

\smallparagraph{Causal Representation Learning}
A common basis for causal representation learning is Independent Component Analysis (ICA) \cite{comon1994independent, hyvaerinen2001independent}, which aims to identify independent latent variables from observations.
Due to the non-identifiability for the general case of non-linear observation functions \cite{hyvarinen1999nonlinear}, additional auxiliary variables are often considered in this setting \cite{hyvaerinen2019nonlinear, hyvarinen2016unsupervised}.
Ideas from ICA have been integrated into neural networks \cite{khemakhem2020variational, khemakhem2020ice, reizinger2022embrace} and applied to causality \cite{shimizu2006linear,gresele2021independent, monti2019causal} for identifying causal variables. 

Recently, several works in causal representation learning have exploited distribution shifts or interventions to identify causal variables.
Using counterfactual observations, \citet{brehmer2022weakly, ahuja2022weakly, locatello2020weakly} learn causal variables from pairs of images, between which only a subset of variables has changed via interventions with unknown targets.
For temporal processes, \citet{lachapelle2021disentanglement,lachapelle2022partial} can model interventions of unknown target via \emph{actions}, which is equivalent to the \robotstate{} in our setting, but require that each causal variable has a strictly unique parent set.
On the other hand, \citet{yao2021learning, yao2022temporally} consider observations from $m$ different regimes $u_1,...,u_m$, where, in our setting, the regime indicator $u$ is a discrete version of $R^t$. 
However, they require at least $2K+1$ different regimes compared to $\lfloor \log_2 K\rfloor + 2$ settings for ours, and have stronger conditions on the distribution changes over regimes (\eg{} no additive Gaussian noise models).
In temporal settings where the intervention targets are known, CITRIS \cite{lippe2022citris,lippe2022icitris} identifies scalar and multidimensional causal variables from high-dimensional images.
Nonetheless, observing the intervention targets requires additional supervision, which may not always be available.
To the best of our knowledge, we are the first to use unknown binary interactions to identify the causal variables from high-dimensional observations. 

\section{Experiments}
\label{sec:experiments}

To illustrate the effectiveness of \OurApproach{}, we evaluate it on a synthetic toy benchmark and two environments generated by 3D robotic simulators.
We publish our code at \url{https://github.com/phlippe/BISCUIT}, and detail the data generation and hyperparameters in \cref{sec:appendix_experimental_details}.

\subsection{Synthetic Toy Benchmark}

\begin{figure}[t!]
    \centering
    \begin{subfigure}[b]{0.32\columnwidth}
        \centering
        \includegraphics[width=\textwidth]{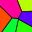}
        \caption{Voronoi}
        \label{fig:experiments_environments_voronoi}
    \end{subfigure}
    \begin{subfigure}[b]{0.32\columnwidth}
        \centering
        \includegraphics[width=\textwidth]{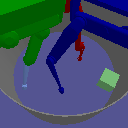}
        \caption{CausalWorld}
        \label{fig:experiments_environments_causalworld}
    \end{subfigure}
    \begin{subfigure}[b]{0.32\columnwidth}
        \centering
        \includegraphics[width=\textwidth]{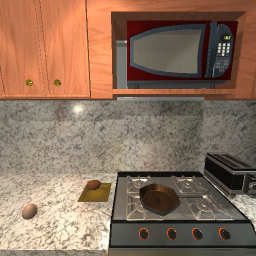}
        \caption{iTHOR}
        \label{fig:experiments_environments_ithor}
    \end{subfigure}
    \caption{Example figures of our three environments with increasing complexity: Voronoi \cite{lippe2022icitris}, CausalWorld \cite{ahmed2020causalworld}, and iTHOR \cite{kolve2017ai}.}
    \label{fig:experiments_environments}
\end{figure}

To evaluate \OurApproach{} on various graph structures, we extend the Voronoi benchmark \cite{lippe2022citris} by replacing observed intervention targets with unobserved binary interactions.
In this dataset, each causal variable follows an additive Gaussian noise model, where the mean is modeled by a randomly initialized MLP.
To determine the parent set, we randomly sample the causal graph with an edge likelihood of $0.4$.
Instead of observing the causal variables directly, they are first entangled by applying a two-layer randomly initialized normalizing flow before visualizing the outputs as colors in a Voronoi diagram of size $32\times 32$ (see \cref{fig:experiments_environments_voronoi}).
We extend the original benchmark by including a robotic arm that moves over the Voronoi diagram and interacts by touching individual color regions/tiles.
Each tile corresponds to one causal variable, allowing for both single- and multi-target interactions.
The models need to deduce these interactions from a \robotstate{} $R^t\in[0,1]^2$ which is the 2D location of the robotic arm on the image.
When the robotic arm interacts with a variable, its mean is set to zero, which resembles a stochastic perfect intervention.

\smallparagraph{Evaluation}
We generate five Voronoi systems with six causal variables, and five systems with nine variables.
We compare \OurApproach{} to iVAE \cite{khemakhem2020variational}, LEAP \cite{yao2021learning}, and Disentanglement via Mechanism Sparsity (DMS) \cite{lachapelle2021disentanglement}, since all use a \robotstate{}. We do not compare with CITRIS \citep{lippe2022citris,lippe2022icitris}, because it requires known intervention targets.
We follow \citet{lippe2022icitris} in evaluating the models on a held-out test set where all causal variables are independently sampled.
We calculate the coefficient of determination \cite{wright1921correlation}, also called the $R^2$ score, between each causal variable $C_i$ and each learned latent variable $z_j$, denoted by $R^2_{ij}$.
If a model identifies the causal variables according to Definition~\ref{def:identifiability}, then for each causal variable $C_i$, there exists one latent variable $z_j$ for which $R^2_{ij}=1$, while it is zero for all others.
Since the alignment of the learned latent variables to causal variables is not known, we report $R^2$ scores for the permutation $\pi$ that maximizes the diagonal of the $R^2$ matrix, \ie{} $R^2\text{-diag}=\nicefrac{1}{K}\sum_{i=1}^{K} R^{2}_{i,\pi(i)}$ (where 1 is optimal).
To account for spurious modeled correlation, we also report the maximum correlation besides this alignment: $R^2\text{-sep}=\nicefrac{1}{K}\sum_{i=1}^{K}\max_{j\neq\pi(i)} R^{2}_{ij}$ (optimal 0).

\begin{figure}[t!]
    \centering
    \begin{tabular}{cc}
        \multicolumn{2}{c}{\includegraphics[width=0.8\columnwidth]{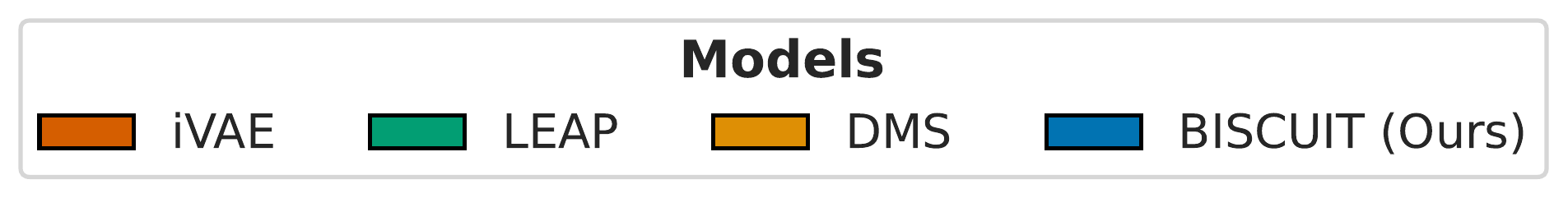}} \\[-2mm]
        \includegraphics[width=0.45\columnwidth]{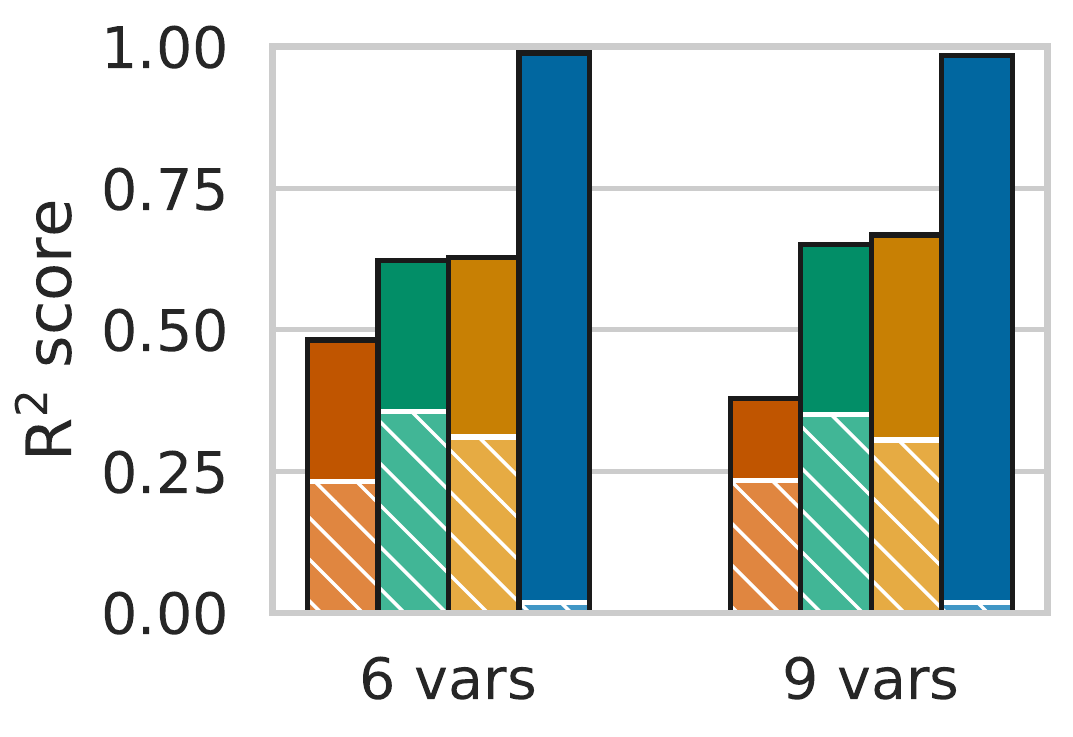} & 
        \includegraphics[width=0.45\columnwidth]{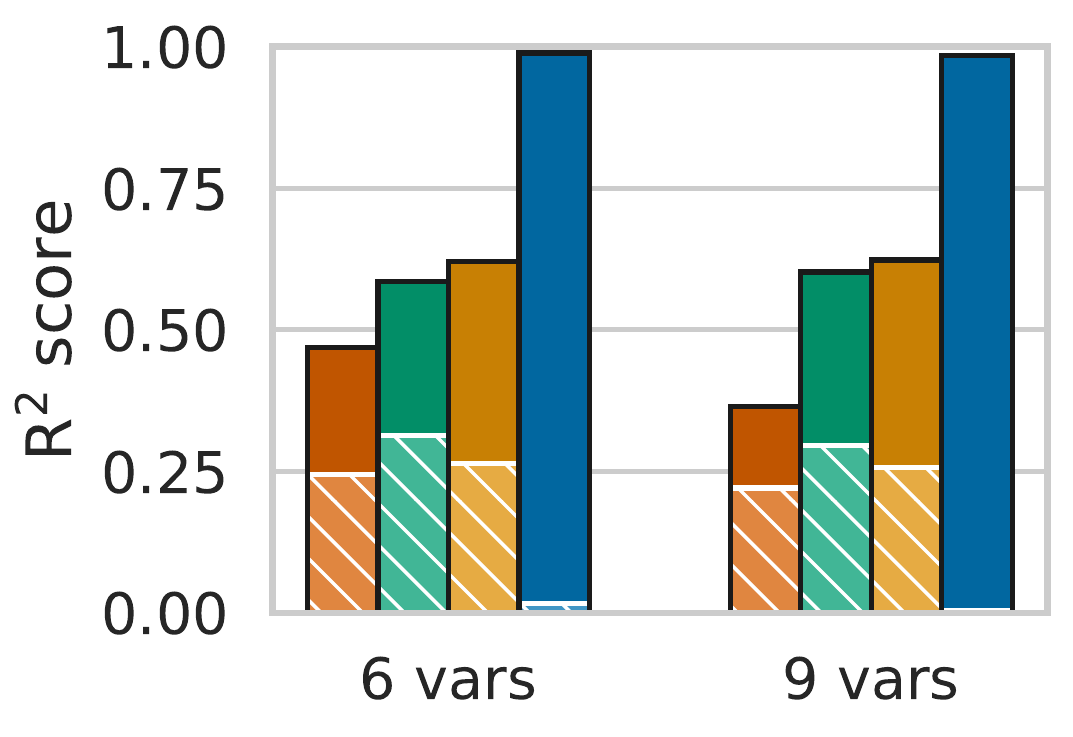} \\
        (a) Random Interactions & (b) Minimal Interactions \\ 
    \end{tabular}
    \caption{Results on the Voronoi benchmark averaged over 10 seeds. Solid bars show the mean $R^2$-diag score (higher is better), and striped bars the $R^2$-sep scores (lower is better, non-visible bars indicate close-to zero values). \OurApproach{} accurately identifies the causal variables across settings.}
    \label{fig:experiments_voronoi_bar_plot}
\end{figure}

\smallparagraph{Results}
The results in \cref{fig:experiments_voronoi_bar_plot}a show that \OurApproach{} identifies the causal variables with high accuracy for both graphs with six and nine variables.
In comparison, all baselines struggle to identify the causal variables, often falling back to modeling the colors as latent variables instead. 
While the assumptions of iVAE and LEAP do not hold for additive Gaussian noise models, the assumptions of DMS, including the graph sparsity, mostly hold.
Still, \OurApproach{} is the only method to consistently identify the true variables, illustrating that its stable optimization and robustness.

\smallparagraph{Minimal Number of Regimes}
To verify that \OurApproach{} only requires $\lfloor \log_2 K \rfloor + 2$ different regimes (\cref{theo:main_theorem}), we repeat the previous experiments with reducing the interaction maps to a minimum.
This results in four sets of interactions for six variables, and five for nine variables.
\cref{fig:experiments_voronoi_bar_plot}b shows that \OurApproach{} still correctly identifies causal variables in this setting, supporting our theoretical results.

\smallparagraph{Learned Intervention Targets} 
After training, we can use the interaction variables $\hat{I}_1,...,\hat{I}_M$ learned by \OurApproach{} to identify the regions in which the robotic arm interacts with a causal variable.
Based on our theoretical results, we expect that some of the learned variables are identical to the true interaction variables $I_1,...I_K$ up to permutations and sign-flips.
In all settings, we find that the learned binary variables match the true interaction variables with an average F1 score of $98\%$ for the same permutation of variables as in the $R^2$ evaluation. This shows that \OurApproach{} identified the true interaction variables.
Thus, in practice, one could use a few samples with labeled interaction variables to identify the learned permutation of the model.

\subsection{CausalWorld}
\label{sec:experiments_causalworld}

CausalWorld \cite{ahmed2020causalworld} is a robotic manipulation environment with a tri-finger robot, which can interact with objects in an enclosed space by touch (see \cref{fig:experiments_environments_causalworld}).
The environment also allows for interventions on various environment parameters, including the colors or friction parameters of individual elements.
We experiment on this environment by recording the robot's interactions with a cube.
Besides the cube position, rotation and velocity, the causal variables are the colors of the three fingertips, as well as the floor, stage and cube friction, which we visualize by the colors of the respective objects.
All colors and friction parameters follow an additive Gaussian noise model. 
When a robot finger touches the cube, we perform a stochastic perfect intervention on its color.
Similarly, an interaction with the friction parameters correspond to touching these objects with all three fingers.
The \robotstate{} $R^t$ is modeled by the angles of the three motors per robot finger from the current and previous time step, providing velocity information. 

This environment provides two new challenges.
Firstly, not all interactions are necessarily binary.
In particular, the collisions between the robot and the cube have different effects depending on the velocity and direction of the fingers of the robot, which are not part of the state of the causal variables at the previous time step.
Additionally, the robotic system is present in the observation/image, while our theoretical results assume that $R^t$ is not a direct cause of $X^t$.
We adapt \OurApproach{}-NF and the baselines to this case by adding $R^t$ as additional information to the decoder, effectively removing the need to model $R^t$ in the latent space.

\begin{table}[t!]
    \centering
    \caption{$R^2$ scores (diag $\uparrow$ / sep $\downarrow$) for the identification of the causal variables on CausalWorld and iTHOR.}
    \resizebox{\columnwidth}{!}{
    \begin{tabular}{lcc}
        \toprule
        \textbf{Models} & \textbf{CausalWorld} & \textbf{iTHOR} \\
        \midrule
        iVAE \cite{khemakhem2020variational} & 0.28 / 0.00 & 0.48 / 0.35 \\
        LEAP \cite{yao2021learning} & 0.30 / 0.00 & 0.63 / 0.45 \\
        DMS \cite{lachapelle2021disentanglement} & 0.32 / 0.00 & 0.61 / 0.40\\
        \OurApproach{}-NF (Ours) & \textbf{0.97} / 0.01 & \textbf{0.96} / \textbf{0.15}\\
        \bottomrule
    \end{tabular}
    }
    \label{tab:experiments_causalworld_ithor}
\end{table}

On this task, \OurApproach{} identifies the causal variables well, as seen in \cref{tab:experiments_causalworld_ithor}.
Because the cube position, velocity and rotation share the same interactions, in the evaluation we consider them as a multidimensional variable.
Although the true model cannot be fully described by binary interaction variables, \OurApproach{} still models the binary information of whether a collision happens or not for the cube, since it is the most important part of the dynamics.
We verify this in \cref{sec:appendix_causalworld_results} by measuring the F1 score between the predicted interaction variables and ground truth interactions/collisions.
\OurApproach{} achieves an F1 score of 50\% for all cube-arm interactions, which indicates a high similarity between the learned
interaction and the ground truth collisions considering that collisions only happen in approximately 5\%
of the frames.
The mismatches are mostly due to the learned interactions being more conservative, \ie{} being 1 already a frame too early sometimes. 
Meanwhile, none of the baselines are able to reconstruct the image sufficiently, missing the robotic arms and the cube (see \cref{sec:appendix_causalworld_results}).
While this might improve with significant tuning effort, \OurApproach{}-NF is not sensitive to the difficulty of the reconstruction due to its separate autoencoder training stage.

\subsection{iTHOR - Embodied AI}
\label{sec:experiments_ithor}

To illustrate the potential of causal representation learning in embodied AI, we apply \OurApproach{} to the iTHOR environment \cite{kolve2017ai}.
In this environment, an embodied AI agent can perform actions on various objects in an 3D indoor scene such as a kitchen.
These agent-object interactions can often be described by a binary variable, \eg{} pickup/put down an object, open/close a door, turn on/off an object, etc., which makes it an ideal setup for \OurApproach{}.

Our goal in this environment is to identify the causal variables, \ie{} the objects and their states, from sequences of interactions.
We perform this task on the kitchen environment shown in \cref{fig:experiments_environments_ithor}.
This environment contains two movable objects, \ie{} a plate and an egg, and seven static objects, \eg{} a microwave and a stove.
Overall, we have 18 causal variables, which include both continuous, \eg{} the location of the plate, and binary variables, \eg{} whether the microwave is on or off.
Causal variables influence each other by state changes, \eg{} the egg gets cooked when it is in the pan and the stove is turned on.
Further, the set of possible actions that can be performed depends on the previous time step, \eg{} only one object can be picked up at a time.
For training, we generate a dataset where we randomly pick a valid action at each time step.
We model the \robotstate{} $R^t$ as a two-dimensional pixel coordinate, which is the position of a pixel showing the interacted object in the image ($R^t\in[0,1]^2$).
This simulates iTHOR's web demo \cite{kolve2017ai}, where a user interacts with objects by clicking on them.

We train \OurApproach{}-NF and our baselines on this dataset, and compare the latent representation to the ground truth causal variables in terms of the $R^2$ score in \cref{tab:experiments_causalworld_ithor}.
Although the baselines reconstruct the image mostly well, the causal variables are highly entangled in their representations.
In contrast, \OurApproach{} identifies and separates most of the causal variables optimally, except for the two movable objects (egg/plate).
This is likely due to the high inherent correlation of the two objects, since their positions cannot overlap and only one of them can be picked up at a time.

\begin{figure}[t!]
    \centering
    \footnotesize
    \setlength{\tabcolsep}{1pt}
    \begin{tabular}{ccc}
        \includegraphics[width=0.3\linewidth]{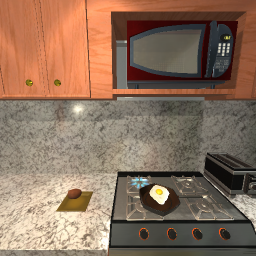} & 
        \includegraphics[width=0.3\linewidth]{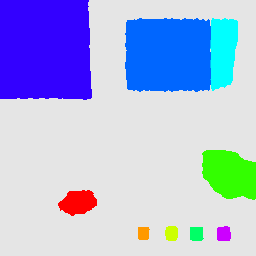} &
        \includegraphics[width=0.3\linewidth]{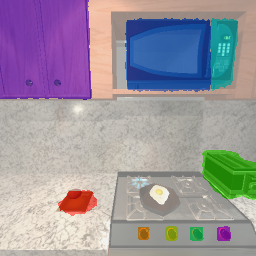} \\
        Input Image & Learned Interactions & Combined Image \\
    \end{tabular}
    \caption{Visualizing the learned interaction variables of \OurApproach{} for an example input image (left). We show the locations, \ie{} values of $R^t\in[0,1]^2$, for which each interaction variable is greater than zero/active as different colors. For readability, only nine interaction variables are shown. The right image is an overlay of both. \OurApproach{} accurately learns the interactions and adapts them to the input image.
    }
    \label{fig:experiments_ithor_interaction_maps}
\end{figure}

Besides evaluating the causal representation, we also visualize the learned interaction variables of \OurApproach{} in \cref{fig:experiments_ithor_interaction_maps}. 
Here, each color represents the region in which \OurApproach{} identified an interaction with a different causal variable.
\cref{fig:experiments_ithor_interaction_maps} shows that \OurApproach{} has identified the correct interaction region for each object.
Moreover, it allows for context-dependent interactions, as the location of the plate influences the region of its corresponding interaction variable.

\begin{figure}[t!]
    \centering
    \footnotesize
    \setlength{\tabcolsep}{1pt}
    \begin{tabular}{ccc}
        \includegraphics[width=0.3\linewidth]{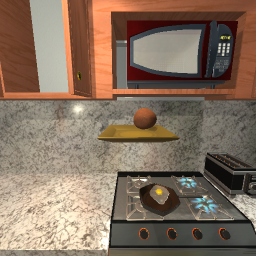} & 
        \includegraphics[width=0.3\linewidth]{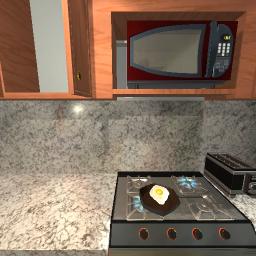} &
        \includegraphics[width=0.3\linewidth]{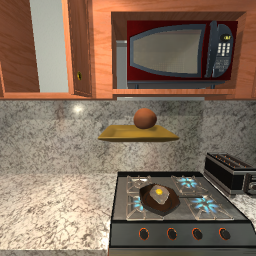} \\
        Input Image 1 & Input Image 2 & Generated Output \\
    \end{tabular}
    \caption{
    Performing interventions in the latent space. First, the two inputs images are encoded into latent space. Then, we replace the latents of the front-left stove and microwave in the first image by the corresponding latents of the second image. This corresponds to effectively turning on the stove and turning off the microwave in the first image. All remaining latents are kept fixed, aiming the other causal variables to remain unchanged from their state in the first image. Decoding these new latents in the right image shows that \OurApproach{} can model the effect of interventions accurately. Furthermore, it is able to create the novel, unseen scenario where the egg is uncooked, but the stove is turned on. This highlights the modularity of \OurApproach{}'s representations.
    }
    \label{fig:experiments_ithor_triplet}
\end{figure}

Finally, we can use the learned causal representation to perform interventions and create novel combinations of causal variables.
For this, we encode two images into the learned latent space of \OurApproach{}, and combine the latent representations of the causal variables to have a novel image decoded.
For example, in \cref{fig:experiments_ithor_triplet}, we replace the latents representing the front-left stove and the microwave state in the first image by the corresponding latents of the second image.
This aims to perform an intervention on the front-left stove (turning on) and the microwave state (turning off) while all remaining causal variables such as the egg state should stay unchanged.\footnote{This setup can also be interpreted as performing a perfect intervention on all causal variables with values picked from image 1 or 2 for individual variables. Causal relations between variables, \eg{} the stove and the egg, are actively broken in this setup.}
\OurApproach{} not only integrates these changes without influencing any of the other causal variables, but generates a completely novel state: even though in the iTHOR environment, the egg is instantaneously cooked when the stove turns on, \OurApproach{} correctly combines the state of the egg being raw with the stove burning.
This shows the capabilities of \OurApproach{} to model unseen causal interventions. 

\section{Conclusion}
\label{sec:conclusion}

We prove that under mild assumptions, causal variables become identifiable from high-dimensional observations, when their interactions with an external system can be described by unknown binary variables.
As a practical algorithm, we propose \OurApproach{}, which learns the causal variables and their interaction variables.
In experiments across three robotic-inspired datasets, \OurApproach{} outperforms previous methods in identifying the causal variables from images.

While in experiments, \OurApproach{} shows strong identification even for complex interactions, the presented theory is currently limited to binary interaction variables.
Although the first step may be to generalize the theory to interaction variables with more than two states, extensions to unknown domains or sparse, continuous interaction variables are other interesting future directions.
Instead of assuming distinct interaction patterns, future work can extend these results to partial identifiability, similar to \citet{lippe2022citris, lachapelle2022partial}.
Finally, our results open up the opportunity for empirical studies showing the benefits of causal representations for complex real-world tasks like embodied AI. 

\begin{contributions}
    P.~Lippe conceived the idea, developed the theoretical results, implemented the models and experiments, and wrote the paper.
    S.~Magliacane, S.~L\"owe, Y.~M.~Asano, T.~Cohen, E.~Gavves advised during the project and helped in writing the paper.
\end{contributions}

\begin{acknowledgements}
    We thank SURF for the support in using the National Supercomputer Snellius.
    This work is financially supported by Qualcomm Technologies Inc., the University of Amsterdam and the allowance Top consortia for Knowledge and Innovation (TKIs) from the Netherlands Ministry of Economic Affairs and Climate Policy.
\end{acknowledgements}

\bibliography{references}

\begin{thebibliography}{54}
\providecommand{\natexlab}[1]{#1}
\providecommand{\url}[1]{\texttt{#1}}
\expandafter\ifx\csname urlstyle\endcsname\relax
  \providecommand{\doi}[1]{doi: #1}\else
  \providecommand{\doi}{doi: \begingroup \urlstyle{rm}\Url}\fi

\bibitem[Ahmed et~al.(2020)Ahmed, Tr{\"a}uble, Goyal, Neitz, Bengio,
  Sch{\"o}lkopf, W{\"u}thrich, and Bauer]{ahmed2020causalworld}
Ahmed, O., Tr{\"a}uble, F., Goyal, A., Neitz, A., Bengio, Y., Sch{\"o}lkopf,
  B., W{\"u}thrich, M., and Bauer, S.
\newblock {CausalWorld: A robotic manipulation benchmark for causal structure
  and transfer learning}.
\newblock \emph{arXiv preprint arXiv:2010.04296}, 2020.

\bibitem[Ahuja et~al.(2022)Ahuja, Hartford, and Bengio]{ahuja2022weakly}
Ahuja, K., Hartford, J., and Bengio, Y.
\newblock {Weakly Supervised Representation Learning with Sparse
  Perturbations}.
\newblock In \emph{Advances in Neural Information Processing Systems 35,
  NeurIPS}, 2022.

\bibitem[Assaad et~al.(2022)Assaad, Devijver, and Gaussier]{assaad2022survey}
Assaad, C.~K., Devijver, E., and Gaussier, E.
\newblock Survey and evaluation of causal discovery methods for time series.
\newblock \emph{J. Artif. Int. Res.}, 73, may 2022.
\newblock ISSN 1076-9757.

\bibitem[Bengio et~al.(2013)Bengio, L{\'e}onard, and
  Courville]{bengio2013estimating}
Bengio, Y., L{\'e}onard, N., and Courville, A.
\newblock {Estimating or propagating gradients through stochastic neurons for
  conditional computation}.
\newblock \emph{arXiv preprint arXiv:1308.3432}, 2013.

\bibitem[{Brehmer} et~al.(2022){Brehmer}, {de Haan}, {Lippe}, and
  {Cohen}]{brehmer2022weakly}
{Brehmer}, J., {de Haan}, P., {Lippe}, P., and {Cohen}, T.
\newblock {Weakly supervised causal representation learning}.
\newblock In \emph{Advances in Neural Information Processing Systems 35,
  NeurIPS}, 2022.

\bibitem[Brouillard et~al.(2020)Brouillard, Lachapelle, Lacoste,
  Lacoste{-}Julien, and Drouin]{brouillard2020differentiable}
Brouillard, P., Lachapelle, S., Lacoste, A., Lacoste{-}Julien, S., and Drouin,
  A.
\newblock {Differentiable Causal Discovery from Interventional Data}.
\newblock In \emph{Advances in Neural Information Processing Systems 33,
  NeurIPS}, 2020.

\bibitem[Comon(1994)]{comon1994independent}
Comon, P.
\newblock {Independent component analysis, A new concept?}
\newblock \emph{Signal Processing}, 36\penalty0 (3), April 1994.

\bibitem[Dean et~al.(1989)Dean and Kanazawa]{DBN}
Dean, T. and Kanazawa, K.
\newblock {A model for reasoning about persistence and causation}.
\newblock \emph{Computational Intelligence}, 5\penalty0 (2), 1989.

\bibitem[Dinh et~al.(2017)Dinh, Sohl{-}Dickstein, and Bengio]{dinh2017density}
Dinh, L., Sohl{-}Dickstein, J., and Bengio, S.
\newblock {Density estimation using Real NVP}.
\newblock In \emph{5th International Conference on Learning Representations,
  {ICLR} 2017, Toulon, France, April 24-26, 2017, Conference Track
  Proceedings}, 2017.

\bibitem[Falcon et~al.(2019)Falcon and {The PyTorch Lightning
  team}]{Falcon_PyTorch_Lightning_2019}
Falcon, W. and {The PyTorch Lightning team}.
\newblock {PyTorch Lightning}, 2019.

\bibitem[Faria et~al.(2022)Faria, Martins, and
  Figueiredo]{faria2022differentiable}
Faria, G. R.~A., Martins, A., and Figueiredo, M. A.~T.
\newblock {Differentiable Causal Discovery Under Latent Interventions}.
\newblock In \emph{Proceedings of the First Conference on Causal Learning and
  Reasoning}, volume 177 of \emph{Proceedings of Machine Learning Research}.
  PMLR, 11--13 Apr 2022.

\bibitem[Gresele et~al.(2021)Gresele, von K{\"u}gelgen, Stimper, Sch{\"o}lkopf,
  and Besserve]{gresele2021independent}
Gresele, L., von K{\"u}gelgen, J., Stimper, V., Sch{\"o}lkopf, B., and
  Besserve, M.
\newblock {Independent mechanism analysis, a new concept?}
\newblock In \emph{Advances in Neural Information Processing Systems}, 2021.

\bibitem[Hafner et~al.(2021)Hafner, Lillicrap, Norouzi, and
  Ba]{hafner2021mastering}
Hafner, D., Lillicrap, T.~P., Norouzi, M., and Ba, J.
\newblock {Mastering Atari with Discrete World Models}.
\newblock In \emph{International Conference on Learning Representations}, 2021.

\bibitem[He et~al.(2016)He, Zhang, Ren, and Sun]{he2016deep}
He, K., Zhang, X., Ren, S., and Sun, J.
\newblock {Deep residual learning for image recognition}.
\newblock In \emph{Proceedings of the IEEE conference on computer vision and
  pattern recognition}, 2016.

\bibitem[Hyv\"{a}rinen et~al.(2016)Hyv\"{a}rinen and
  Morioka]{hyvarinen2016unsupervised}
Hyv\"{a}rinen, A. and Morioka, H.
\newblock {Unsupervised Feature Extraction by Time-Contrastive Learning and
  Nonlinear ICA}.
\newblock In \emph{Proceedings of the 30th International Conference on Neural
  Information Processing Systems}. Curran Associates Inc., 2016.

\bibitem[Hyv\"{a}rinen et~al.(1999)Hyv\"{a}rinen and
  Pajunen]{hyvarinen1999nonlinear}
Hyv\"{a}rinen, A. and Pajunen, P.
\newblock {Nonlinear Independent Component Analysis: Existence and Uniqueness
  Results}.
\newblock \emph{Neural Netw.}, 12\penalty0 (3), apr 1999.
\newblock ISSN 0893-6080.

\bibitem[Hyv{\"a}rinen et~al.(2001)Hyv{\"a}rinen, Karhunen, and
  Oja]{hyvaerinen2001independent}
Hyv{\"a}rinen, A., Karhunen, J., and Oja, E.
\newblock \emph{{Independent Component Analysis}}.
\newblock John Wiley \& Sons, June 2001.

\bibitem[Hyv{\"a}rinen et~al.(2019)Hyv{\"a}rinen, Sasaki, and
  Turner]{hyvaerinen2019nonlinear}
Hyv{\"a}rinen, A., Sasaki, H., and Turner, R.
\newblock {Nonlinear ICA Using Auxiliary Variables and Generalized Contrastive
  Learning}.
\newblock In \emph{Proceedings of the {Twenty-Second} International Conference
  on Artificial Intelligence and Statistics}, volume~89 of \emph{Proceedings of
  Machine Learning Research}. PMLR, 2019.

\bibitem[Jaber et~al.(2020)Jaber, Kocaoglu, Shanmugam, and
  Bareinboim]{jaber2020causal}
Jaber, A., Kocaoglu, M., Shanmugam, K., and Bareinboim, E.
\newblock {Causal Discovery from Soft Interventions with Unknown Targets:
  Characterization and Learning}.
\newblock In \emph{Advances in Neural Information Processing Systems 33,
  NeurIPS}, 2020.

\bibitem[Khemakhem et~al.(2020{\natexlab{a}})Khemakhem, Kingma, Monti, and
  Hyvarinen]{khemakhem2020variational}
Khemakhem, I., Kingma, D., Monti, R., and Hyvarinen, A.
\newblock {Variational Autoencoders and Nonlinear ICA: A Unifying Framework}.
\newblock In \emph{Proceedings of the Twenty Third International Conference on
  Artificial Intelligence and Statistics}, volume 108 of \emph{Proceedings of
  Machine Learning Research}. PMLR, 2020{\natexlab{a}}.

\bibitem[Khemakhem et~al.(2020{\natexlab{b}})Khemakhem, Monti, Kingma, and
  Hyvarinen]{khemakhem2020ice}
Khemakhem, I., Monti, R., Kingma, D., and Hyvarinen, A.
\newblock {ICE-BeeM: Identifiable Conditional Energy-Based Deep Models Based on
  Nonlinear ICA}.
\newblock In \emph{Advances in Neural Information Processing Systems 33,
  NeurIPS}, 2020{\natexlab{b}}.

\bibitem[Kingma et~al.(2015)Kingma and Ba]{kingma2015adam}
Kingma, D.~P. and Ba, J.
\newblock {Adam: A Method for Stochastic Optimization}.
\newblock In \emph{3rd International Conference on Learning Representations,
  {ICLR} 2015, San Diego, CA, USA, May 7-9, 2015, Conference Track
  Proceedings}, 2015.

\bibitem[Kingma et~al.(2018)Kingma and Dhariwal]{kingma2018glow}
Kingma, D.~P. and Dhariwal, P.
\newblock {Glow: Generative Flow with Invertible 1x1 Convolutions}.
\newblock In \emph{Advances in Neural Information Processing Systems},
  volume~31. Curran Associates, Inc., 2018.

\bibitem[Kingma et~al.(2014)Kingma and Welling]{kingma2014auto}
Kingma, D.~P. and Welling, M.
\newblock {Auto-Encoding Variational Bayes}.
\newblock In \emph{2nd International Conference on Learning Representations,
  {ICLR} 2014, Banff, AB, Canada, April 14-16, 2014, Conference Track
  Proceedings}, 2014.

\bibitem[Klindt et~al.(2021)Klindt, Schott, Sharma, Ustyuzhaninov, Brendel,
  Bethge, and Paiton]{klindt2021towards}
Klindt, D., Schott, L., Sharma, Y., Ustyuzhaninov, I., Brendel, W., Bethge, M.,
  and Paiton, D.
\newblock {Towards Nonlinear Disentanglement in Natural Data with Temporal
  Sparse Coding}.
\newblock In \emph{International Conference on Learning Representations
  ({ICLR})}, 2021.

\bibitem[Kolve et~al.(2017)Kolve, Mottaghi, Han, VanderBilt, Weihs, Herrasti,
  Gordon, Zhu, Gupta, and Farhadi]{kolve2017ai}
Kolve, E., Mottaghi, R., Han, W., VanderBilt, E., Weihs, L., Herrasti, A.,
  Gordon, D., Zhu, Y., Gupta, A., and Farhadi, A.
\newblock {AI2-THOR: An interactive 3d environment for visual ai}.
\newblock \emph{arXiv preprint arXiv:1712.05474}, 2017.
\newblock Web demo \url{https://ai2thor.allenai.org/demo/}.

\bibitem[Kumar et~al.(2021)Kumar and Sinha]{kumar2021disentangling}
Kumar, A. and Sinha, G.
\newblock {Disentangling mixtures of unknown causal interventions}.
\newblock In \emph{Proceedings of the Thirty-Seventh Conference on Uncertainty
  in Artificial Intelligence}, volume 161 of \emph{Proceedings of Machine
  Learning Research}. PMLR, 27--30 Jul 2021.

\bibitem[Lachapelle et~al.(2022{\natexlab{a}})Lachapelle and
  Lacoste-Julien]{lachapelle2022partial}
Lachapelle, S. and Lacoste-Julien, S.
\newblock {Partial Disentanglement via Mechanism Sparsity}.
\newblock In \emph{UAI 2022 Workshop on Causal Representation Learning},
  2022{\natexlab{a}}.

\bibitem[Lachapelle et~al.(2022{\natexlab{b}})Lachapelle, Rodriguez, Le,
  Sharma, Everett, Lacoste, and Lacoste-Julien]{lachapelle2021disentanglement}
Lachapelle, S., Rodriguez, P., Le, R., Sharma, Y., Everett, K.~E., Lacoste, A.,
  and Lacoste-Julien, S.
\newblock {Disentanglement via Mechanism Sparsity Regularization: A New
  Principle for Nonlinear ICA}.
\newblock In \emph{First Conference on Causal Learning and Reasoning},
  2022{\natexlab{b}}.

\bibitem[Lesort et~al.(2018)Lesort, Díaz-Rodríguez, Goudou, and
  Filliat]{lesort2018state}
Lesort, T., Díaz-Rodríguez, N., Goudou, J.-F., and Filliat, D.
\newblock State representation learning for control: An overview.
\newblock \emph{Neural Networks}, 108, 2018.

\bibitem[Lippe et~al.(2022{\natexlab{a}})Lippe, Cohen, and
  Gavves]{lippe2022enco}
Lippe, P., Cohen, T., and Gavves, E.
\newblock {Efficient Neural Causal Discovery without Acyclicity Constraints}.
\newblock In \emph{International Conference on Learning Representations},
  2022{\natexlab{a}}.

\bibitem[Lippe et~al.(2022{\natexlab{b}})Lippe, Magliacane, L{\"o}we, Asano,
  Cohen, and Gavves]{lippe2022citris}
Lippe, P., Magliacane, S., L{\"o}we, S., Asano, Y.~M., Cohen, T., and Gavves,
  E.
\newblock {CITRIS: Causal Identifiability from Temporal Intervened Sequences}.
\newblock In \emph{Proceedings of the 39th International Conference on Machine
  Learning, {ICML}}, 2022{\natexlab{b}}.

\bibitem[Lippe et~al.(2022{\natexlab{c}})Lippe, Magliacane, L{\"o}we, Asano,
  Cohen, and Gavves]{lippe2022intervention}
Lippe, P., Magliacane, S., L{\"o}we, S., Asano, Y.~M., Cohen, T., and Gavves,
  E.
\newblock {Intervention Design for Causal Representation Learning}.
\newblock In \emph{UAI 2022 Workshop on Causal Representation Learning},
  2022{\natexlab{c}}.

\bibitem[Lippe et~al.(2023)Lippe, Magliacane, L{\"o}we, Asano, Cohen, and
  Gavves]{lippe2022icitris}
Lippe, P., Magliacane, S., L{\"o}we, S., Asano, Y.~M., Cohen, T., and Gavves,
  E.
\newblock Causal representation learning for instantaneous and temporal effects
  in interactive systems.
\newblock In \emph{The Eleventh International Conference on Learning
  Representations}, 2023.

\bibitem[Locatello et~al.(2020)Locatello, Poole, R{\"{a}}tsch, Sch{\"{o}}lkopf,
  Bachem, and Tschannen]{locatello2020weakly}
Locatello, F., Poole, B., R{\"{a}}tsch, G., Sch{\"{o}}lkopf, B., Bachem, O.,
  and Tschannen, M.
\newblock {Weakly-Supervised Disentanglement Without Compromises}.
\newblock In \emph{Proceedings of the 37th International Conference on Machine
  Learning, {ICML}}, 2020.

\bibitem[Mian et~al.(2023)Mian, Kamp, and Vreeken]{mian2023information}
Mian, O., Kamp, M., and Vreeken, J.
\newblock Information-theoretic causal discovery and intervention detection
  over multiple environments.
\newblock In \emph{Proceedings of the AAAI Conference on Artificial
  Intelligence}, AAAI-23, 2023.

\bibitem[Monti et~al.(2019)Monti, Zhang, and Hyv{\"{a}}rinen]{monti2019causal}
Monti, R.~P., Zhang, K., and Hyv{\"{a}}rinen, A.
\newblock {Causal Discovery with General Non-Linear Relationships using
  Non-Linear ICA}.
\newblock In \emph{Proceedings of the Thirty-Fifth Conference on Uncertainty in
  Artificial Intelligence, {UAI}}, 2019.

\bibitem[Mooij et~al.(2020)Mooij, Magliacane, and Claassen]{mooij2020joint}
Mooij, J.~M., Magliacane, S., and Claassen, T.
\newblock {Joint Causal Inference from Multiple Contexts}.
\newblock \emph{Journal of Machine Learning Research}, 21\penalty0 (99), 2020.

\bibitem[Murphy(2002)]{Murphy_DBN}
Murphy, K.
\newblock {Dynamic Bayesian Networks: Representation, Inference and Learning}.
\newblock \emph{UC Berkeley, Computer Science Division}, 2002.

\bibitem[Paszke et~al.(2019)Paszke, Gross, Massa, Lerer, Bradbury, Chanan,
  Killeen, Lin, Gimelshein, Antiga, Desmaison, K{\"{o}}pf, Yang, DeVito,
  Raison, Tejani, Chilamkurthy, Steiner, Fang, Bai, and
  Chintala]{paszke2019pytorch}
Paszke, A., Gross, S., Massa, F., Lerer, A., Bradbury, J., Chanan, G., Killeen,
  T., Lin, Z., Gimelshein, N., Antiga, L., Desmaison, A., K{\"{o}}pf, A., Yang,
  E., DeVito, Z., Raison, M., Tejani, A., Chilamkurthy, S., Steiner, B., Fang,
  L., Bai, J., and Chintala, S.
\newblock {PyTorch: An Imperative Style, High-Performance Deep Learning
  Library}.
\newblock In \emph{Advances in Neural Information Processing Systems 32: Annual
  Conference on Neural Information Processing Systems 2019, NeurIPS 2019,
  December 8-14, 2019, Vancouver, BC, Canada}, 2019.

\bibitem[Peters et~al.(2013)Peters, Janzing, and
  Sch\"{o}lkopf]{peters2013causal}
Peters, J., Janzing, D., and Sch\"{o}lkopf, B.
\newblock Causal inference on time series using restricted structural equation
  models.
\newblock In Burges, C., Bottou, L., Welling, M., Ghahramani, Z., and
  Weinberger, K. (eds.), \emph{Advances in Neural Information Processing
  Systems}, volume~26. Curran Associates, Inc., 2013.

\bibitem[Ramachandran et~al.(2017)Ramachandran, Zoph, and
  Le]{ramachandran2017searching}
Ramachandran, P., Zoph, B., and Le, Q.~V.
\newblock {Searching for activation functions}.
\newblock \emph{arXiv preprint arXiv:1710.05941}, 2017.

\bibitem[Reizinger et~al.(2022)Reizinger, Gresele, Brady, von K{\"u}gelgen,
  Zietlow, Sch{\"o}lkopf, Martius, Brendel, and Besserve]{reizinger2022embrace}
Reizinger, P., Gresele, L., Brady, J., von K{\"u}gelgen, J., Zietlow, D.,
  Sch{\"o}lkopf, B., Martius, G., Brendel, W., and Besserve, M.
\newblock {Embrace the Gap: VAEs Perform Independent Mechanism Analysis}.
\newblock In \emph{Advances in Neural Information Processing Systems 35,
  NeurIPS}, 2022.

\bibitem[Rezende et~al.(2015)Rezende and Mohamed]{rezende2015variational}
Rezende, D.~J. and Mohamed, S.
\newblock {Variational Inference with Normalizing Flows}.
\newblock In \emph{Proceedings of the 32nd International Conference on Machine
  Learning, {ICML}}, 2015.

\bibitem[Sch{\"o}lkopf et~al.(2021)Sch{\"o}lkopf, Locatello, Bauer, Ke,
  Kalchbrenner, Goyal, and Bengio]{schoelkopf2021towards}
Sch{\"o}lkopf, B., Locatello, F., Bauer, S., Ke, N.~R., Kalchbrenner, N.,
  Goyal, A., and Bengio, Y.
\newblock {Toward causal representation learning}.
\newblock \emph{Proceedings of the IEEE}, 109\penalty0 (5), 2021.

\bibitem[Shimizu et~al.(2006)Shimizu, Hoyer, Hyv\"{a}rinen, and
  Kerminen]{shimizu2006linear}
Shimizu, S., Hoyer, P.~O., Hyv\"{a}rinen, A., and Kerminen, A.
\newblock {A Linear Non-Gaussian Acyclic Model for Causal Discovery}.
\newblock \emph{J. Mach. Learn. Res.}, 7, dec 2006.

\bibitem[Squires et~al.(2020)Squires, Wang, and Uhler]{squires2020permutation}
Squires, C., Wang, Y., and Uhler, C.
\newblock {Permutation-based causal structure learning with unknown
  intervention targets}.
\newblock In \emph{Conference on Uncertainty in Artificial Intelligence}. PMLR,
  2020.

\bibitem[Tr{\"a}uble et~al.(2022)Tr{\"a}uble, Dittadi, Wuthrich, Widmaier,
  Gehler, Winther, Locatello, Bachem, Sch{\"o}lkopf, and
  Bauer]{traeuble2022the}
Tr{\"a}uble, F., Dittadi, A., Wuthrich, M., Widmaier, F., Gehler, P.~V.,
  Winther, O., Locatello, F., Bachem, O., Sch{\"o}lkopf, B., and Bauer, S.
\newblock {The Role of Pretrained Representations for the OOD Generalization of
  RL Agents}.
\newblock In \emph{International Conference on Learning Representations}, 2022.

\bibitem[Winkler et~al.(2019)Winkler, Worrall, Hoogeboom, and
  Welling]{winkler2019learning}
Winkler, C., Worrall, D., Hoogeboom, E., and Welling, M.
\newblock {Learning likelihoods with conditional normalizing flows}.
\newblock \emph{arXiv preprint arXiv:1912.00042}, 2019.

\bibitem[Wright(1921)]{wright1921correlation}
Wright, S.
\newblock {Correlation and causation}.
\newblock \emph{Journal of agricultural research}, 20\penalty0 (7), 1921.

\bibitem[Wu et~al.(2018)Wu and He]{wu2018group}
Wu, Y. and He, K.
\newblock Group normalization.
\newblock In \emph{Proceedings of the European conference on computer vision
  (ECCV)}, pp.\  3--19, 2018.

\bibitem[Yao et~al.(2022{\natexlab{a}})Yao, Chen, and Zhang]{yao2022temporally}
Yao, W., Chen, G., and Zhang, K.
\newblock {Temporally Disentangled Representation Learning}.
\newblock In \emph{Advances in Neural Information Processing Systems 35,
  NeurIPS}, 2022{\natexlab{a}}.

\bibitem[Yao et~al.(2022{\natexlab{b}})Yao, Sun, Ho, Sun, and
  Zhang]{yao2021learning}
Yao, W., Sun, Y., Ho, A., Sun, C., and Zhang, K.
\newblock {Learning Temporally Causal Latent Processes from General Temporal
  Data}.
\newblock In \emph{International Conference on Learning Representations},
  2022{\natexlab{b}}.

\bibitem[Zheng et~al.(2018)Zheng, Aragam, Ravikumar, and Xing]{zheng2018dags}
Zheng, X., Aragam, B., Ravikumar, P., and Xing, E.~P.
\newblock {DAGs with NO TEARS: Continuous Optimization for Structure Learning}.
\newblock In \emph{Advances in Neural Information Processing Systems 31: Annual
  Conference on Neural Information Processing Systems 2018, NeurIPS 2018,
  December 3-8, 2018, Montr{\'{e}}al, Canada}, 2018.

\end{thebibliography}
\bibliographystyle{uai2023}

\newpage

\onecolumn %
{
\hrule height4pt
\vskip .25in
\centering
\Large\bfseries
\ourtitle{}
\vskip .25in
\hrule height1pt
\vskip .25in
}

\appendix

\textbf{\LARGE Appendix}
\vskip 8mm
\startcontents[sections]\vbox{\sc\Large Table of Contents}\vspace{4mm}\hrule height .5pt\vspace{4mm}
\printcontents[sections]{l}{1}{\setcounter{tocdepth}{2}}
\newpage

\section{Reproducibility Statement}
\label{sec:appendix_reproducibility}

For reproducibility, we publish the code of \OurApproach{} and the generations of the datasets (Voronoi, CausalWorld, iTHOR) as well as the datasets itself at \url{https://github.com/phlippe/BISCUIT}.
All models were implemented using PyTorch \cite{paszke2019pytorch} and PyTorch Lightning \cite{Falcon_PyTorch_Lightning_2019}.
The hyperparameters and dataset details are described in \cref{sec:experiments} and \cref{sec:appendix_experimental_details}.
All experiments have been repeated for at least three seeds.
We provide an overview of the standard deviation, as well as additional insights to the results in \cref{sec:appendix_experimental_details}.

In terms of computational resources, the experiments of the Voronoi dataset were performed on a single NVIDIA A5000 GPU, with a training time of below 1 hour per model. 
The experiments of the CausalWorld and iTHOR dataset were performed on an NVIDIA A100 GPU (autoencoder training: 1-day training time; variational autoencoders: 1 to 2-day training time) and an NVIDIA A5000 GPU (normalizing flow training, 1-hour training time).

\section{Proofs}
\label{sec:appendix_proofs}

In this section, we prove the main theoretical results in this paper, namely \cref{theo:main_theorem}.
We start with a glossary of the used notation in \cref{sec:appendix_proofs_glossary}, which is the same as in the main paper.
All assumptions for the proof are described in \cref{sec:preliminaries} and \cref{sec:theory} of the main paper.
\cref{sec:proof_steps} contains the proofs for \cref{theo:main_theorem}.
Lastly, we provide further discussion on extensions of the proof, \eg{} to longer temporal dependencies (\cref{sec:proof_longer_temporal_deps}), discovering the causal graph (\cref{sec:proof_causal_graph}), and comparing its results to previous work (\cref{sec:proof_relation_to_prev_results}).

\subsection{Glossary}
\label{sec:appendix_proofs_glossary}

\cref{tab:appendix_glossary} provides an overview of the main notation used in the paper and the following proof.
Additional notation for individual proof steps is introduced in the respective sections.

\begin{table}[h!]
    \centering
    \caption{Glossary of used functions and variables.}
    \begin{tabular}{ll}
        \toprule
        \textbf{Function/Variable} & \textbf{Description} \\
        \midrule
        $\mathcal{M}$ & The true causal model $\mathcal{M}=\langle g, f, \omega, \mathcal{C} \rangle$ \\
        $K$ & Number of causal variables \\
        $C_1,...,C_K$ & The causal variables of $\mathcal{M}$ \\
        $C^t_1,...,C^t_K$ & Instantiations of causal variables of $\mathcal{M}$ at time t\\
        $\mathcal{C}$ & Domain of the causal variables, \ie{} $C^t\in\mathcal{C}$ \\
        $R$ & The \robotstate{} \\
        $I_1,...,I_K$ & Binary interaction variables \\
        $f_1,...,f_K$ & Functions determining the interaction variables, $f_i(R^t,C^{t-1})=I^t_i$ \\
        $p(C^t|C^{t-1},I^{t})$ & Conditional distribution of causal variables \\
        $\omega$ & Parameters of the conditional distribution \\
        $\intvdiv(C^{t}_i|C^{t-1})$ & Interaction effect for the causal variable $C_i$: $\intvdiv(C^{t}_i|C^{t-1})=\log\frac{p(C^{t}_i|C^{t-1},I^t_i=1)}{p(C^{t}_i|C^{t-1},I^t_i=0)}$\\
        $X^t$ & The observation, \eg{} an image \\
        $g$ & The observation function $g(C^t)=X^t$ of $\mathcal{M}$ \\
        $\widehat{\mathcal{M}}$ & Causal model $\widehat{\mathcal{M}}=\langle \hat{g}, \hat{f}, \hat{\omega}, \widehat{\mathcal{C}} \rangle$ with the same data distribution over observations $X$ as $\mathcal{M}$ \\
        $\hat{C}_1,...,\hat{C}_K$ & Causal variables modeled by a model $\widehat{\mathcal{M}}$ with domain $\widehat{\mathcal{C}}$ \\
        $\hat{I}_1,...,\hat{I}_K$ & Binary interaction variables modeled by a model $\widehat{\mathcal{M}}$ \\
        $\hat{g}$ & (Estimated) observation function of the model $\widehat{\mathcal{M}}$, $g(\hat{C}^t)=X^t$ \\
        \bottomrule
    \end{tabular}
    \label{tab:appendix_glossary}
\end{table}

\subsection{Proof Steps}
\label{sec:proof_steps}

The proof consists of four main steps:
\begin{compactenum}
    \item We show that for any $\hat{g}$ in $\widehat{\mathcal{M}}$, there must exist an invertible transformation between the latent space of $\widehat{\mathcal{M}}$ and the true causal model $\mathcal{M}$ (\cref{sec:proof_steps_setup}).
    \item We show that the distributions of different interaction variable values must strictly be different, starting with two variables (\cref{sec:proof_steps_same_intv_cases}) and then moving to the general case (\cref{sec:proof_steps_same_intv_cases_multivar}).
    \item Based on the previous step, we show that $\hat{g}$ must model the same interaction variable patterns for the individual interaction cases, starting with two variables \cref{sec:proof_steps_align_intv_cases}.\cref{sec:proof_steps_align_intv_cases_multivar_no_gaus} discusses it under the assumption of dynamics variability in \cref{theo:main_theorem_no_gaus}, and \cref{sec:proof_steps_align_intv_cases_multivar_time_var} for time variability in \cref{theo:main_theorem_time_var}.
    \item Given that both $\widehat{\mathcal{M}}$ and $\mathcal{M}$ model the same interaction variables, the invertible transformation must be equal to a set of component-wise transformations (\cref{sec:proof_steps_invertible_transformations}).
\end{compactenum}
In \cref{sec:proof_steps_final}, we combine these four steps to prove the full theorem. For step 2 and 3, we first start by discussing the proof idea for a system with only two variables, to give a better intuition behind the proof strategy.

\subsubsection{Existence of an invertible transformation between learned and true representations}
\label{sec:proof_steps_setup}

As a first step, we start with discussing the relation between the true observation function, $g$, and a potentially learned observation function, $\hat{g}$ and show that there exists an invertible transformation between the causal variables that are extracted from an image for each of these two functions.
Throughout this proof, we will use $C_i$ to refer to a causal variable from the original, true causal model $\mathcal{M}$, and we use $\hat{C}_i$ to refer to a latent variable modeled by $\widehat{\mathcal{M}}$, \ie{} an alternative representation of the environment.
Under this setup, we consider the following statement:

\begin{lemma}
    \label{lemma:proof_setup}
    Consider a model $\widehat{\mathcal{M}}=\langle \hat{g}, \hat{f}, \hat{\omega}, \widehat{\mathcal{C}} \rangle$ with an injective observation function $g(\hat{C^t})=X^t$ with $\hat{C^t}\in\widehat{\mathcal{C}}$ and a latent distribution $p_{\omega}(\hat{C}^{t}|\hat{C}^{t-1},R^{t})$, parameterized by $\omega$, which models the same data likelihood as the true causal model $\mathcal{M}$: $p_{\mathcal{M}}(X^t|X^{t-1},R^t)=p_{\widehat{\mathcal{M}}}(X^t|X^{t-1},R^t)$. Then, there must exist an invertible transformation $T$ such that for all $C^t,\hat{C^t}$:
    \begin{equation}
        C^t = T(\hat{C^t}).
    \end{equation}
\end{lemma}

\begin{proof}
    Since $g$ and $\hat{g}$ are injective functions with $\hat{g}(\hat{C^t})=X^t$ and $g(C^t)=X^t$, we have that $X^t=(g \circ g^{-1})(X^t)=(\hat{g} \circ \hat{g}^{-1})(X^t)=\text{id}_X(X^t)$ with $\text{id}$ being the identity function, when restricting the injective functions to their ranges to turn them to invertible functions with $C^t=(g^{-1} \circ g)(C^t)=\text{id}_{C}(C^t)$ and $\hat{C}^t=(\hat{g}^{-1} \circ \hat{g})(\hat{C}^t)=\text{id}_{\hat{C}}(\hat{C}^t)$.
    Combining the two results in: 
    \begin{equation}
        C^t =g^{-1}(X^t) = (g^{-1} \circ \hat{g} \circ \hat{g}^{-1})(X^t) = T(\hat{g}^{-1}(X^t)) = T(\hat{C^t})
    \end{equation}
    where $T=g^{-1} \circ \hat{g}$.
    This function has the inverse $T^{-1}=\hat{g}^{-1} \circ g$ with:
    \begin{equation}
        T^{-1} \circ T = \hat{g}^{-1} \circ \underbrace{g \circ g^{-1}}_{\text{id}_X} \circ \hat{g} = \hat{g}^{-1} \circ \hat{g} = \text{id}_{\hat{C}}
    \end{equation}
    Hence, there exists an invertible function $T=g^{-1} \circ \hat{g}$ between the two spaces of $\hat{C^t}$ and $C^t$.
\end{proof}

Since there exists an invertible, differentiable transformation between the two spaces, we can also express the relation between the two spaces via the change-of-variables:
\begin{equation}
    p_{\omega}(C^{t}|C^{t-1},R^{t})=p_{\hat{\omega}}(\hat{C}^{t}|\hat{C}^{t-1},R^{t}) |\det \bm{J}|
\end{equation}
where $\bm{J}$ is the Jacobian with $\bm{J}_{ij}=\frac{\partial \hat{C}^{t}_i}{\partial C^{t}_j}$.
Further, since there exists an invertible transformation also between $C^{t-1}$ and $\hat{C}^{t-1}$, we can align the conditioning set:
\begin{equation}
    p_{\omega}(C^{t}|C^{t-1},R^{t})=p_{\hat{\omega}}(\hat{C}^{t}|C^{t-1},R^{t}) |\det \bm{J}|
\end{equation}
For readability, we will drop $\omega$ and $\hat{\omega}$ from the index of $p$, since the difference is clear from the context (distribution over $C$/$\hat{C}$).
The following proof steps will take a closer look at aligning these two spaces.

\subsubsection{Any representation requires the same interaction cases - 2 variables}
\label{sec:proof_steps_same_intv_cases}

\paragraph{Setup} 
The idea of this proof step is to show that for any two causal variables $C_1,C_2$, any representation that models the same data likelihood, \eg{} $\hat{C_1}, \hat{C_2}$, must have an invertible transformation between the interaction variables $I^{t}_1,I^{t}_2$ (and $C_1^{t-1}, C_2^{t-1}$) and the learned interaction variables $\hat{I}^{t}_1,\hat{I}^{t}_2$ (and $\hat{C}_1^{t-1}, \hat{C}_2^{t-1}$).
In other words, disentanglement requires distinguishing between the same scenarios of interactions.

We start with considering the four possible interaction cases that we may encounter:
\begin{align}
    p(C_1^{t},C_2^{t}|C^{t-1},I^{t}_1=0,I^{t}_2=0) &= p(C^{t}_1|C^{t-1},I^{t}_1=0) \cdot p(C^{t}_2|C^{t-1},I^{t}_2=0) \\
    p(C_1^{t},C_2^{t}|C^{t-1},I^{t}_1=1,I^{t}_2=0) &= p(C^{t}_1|C^{t-1},I^{t}_1=1) \cdot p(C^{t}_2|C^{t-1},I^{t}_2=0) \\
    p(C_1^{t},C_2^{t}|C^{t-1},I^{t}_1=0,I^{t}_2=1) &= p(C^{t}_1|C^{t-1},I^{t}_1=0) \cdot p(C^{t}_2|C^{t-1},I^{t}_2=1) \\
    p(C_1^{t},C_2^{t}|C^{t-1},I^{t}_1=1,I^{t}_2=1) &= p(C^{t}_1|C^{t-1},I^{t}_1=1) \cdot p(C^{t}_2|C^{t-1},I^{t}_2=1)
\end{align}
Our goal is to show that all these four distributions must be strictly different for any $C^{t-1}$.
These inequalities generalize to any alternative representation that entangles the two variables $C_1,C_2$, since the alternative representation must model the same distributions $p(X^{t}|X^{t-1},...)=p(C^{t}|...)=p(C_1^{t},C_2^{t}|...)\cdot ...$.

\paragraph{Implications of theorem assumptions}
Before comparing the distributions, we first simplify what the assumptions of \cref{theo:main_theorem_no_gaus} imply for the individual variable's distributions.
The theorem assumes that $\intvdiv(C^t_i|C^{t-1}) = \log \nicefrac{p(C^{t}_i|C^{t-1},I^t_i=1)}{p(C^{t}_i|C^{t-1},I^t_i=0)}$ is differentiable and cannot be a constant.
Otherwise, the derivatives $\intvdiv(C^t_i|C^{t-1})$ would have to be constantly zero, which violates both condition (A) and (B) of the theorem.
Therefore, we can deduce that:
\begin{compactitem}
    \item For each variable $C_i$, there must exist at least one value of $C^t_i$ for which $p(C^{t}_i|C^{t-1},I^t_i=1)\neq p(C^{t}_i|C^{t-1},I^t_i=0)$, \ie{} $p(C^{t}_i|C^{t-1},I^t_i=1), p(C^{t}_i|C^{t-1},I^t_i=0)$ must strictly be different distributions.
    \item The distributions $p(C^{t}_i|C^{t-1},I^t_i=1), p(C^{t}_i|C^{t-1},I^t_i=0)$ must share the same support, since otherwise $\intvdiv(C^t_i|C^{t-1})=\pm\infty$ for some $C^t_i$ and thus not differentiable.
\end{compactitem}

\paragraph{Single-target vs Joint} 
We start with comparing single-target interactions versus the observational case.
Since the interactional distribution is strictly different from the observational, we obtain that $p(C^{t}_i|C^{t-1},I^{t}_i=0) \neq p(C^{t}_i|C^{t-1},I^{t}_i=1)$. 
With this inequality, we can deduce that:
\begin{align}
    p(C_1^{t},C_2^{t}|C^{t-1},I^{t}_1=0,I^{t}_2=0) &\neq p(C_1^{t},C_2^{t}|C^{t-1},I^{t}_1=1,I^{t}_2=0) \\
    p(C_1^{t},C_2^{t}|C^{t-1},I^{t}_1=0,I^{t}_2=0) &\neq p(C_1^{t},C_2^{t}|C^{t-1},I^{t}_1=0,I^{t}_2=1)
\end{align}
This is because these distributions only differ in one sub-distribution (\ie{} either $C_1$ or $C_2$ intervened versus passively observed), which must be strictly different due to our assumption.
A similar reasoning can be used to derive the same inequalities for the joint interaction case:
\begin{align}
    p(C_1^{t},C_2^{t}|C^{t-1},I^{t}_1=1,I^{t}_2=1) &\neq p(C_1^{t},C_2^{t}|C^{t-1},I^{t}_1=1,I^{t}_2=0) \\
    p(C_1^{t},C_2^{t}|C^{t-1},I^{t}_1=1,I^{t}_2=1) &\neq p(C_1^{t},C_2^{t}|C^{t-1},I^{t}_1=0,I^{t}_2=1)
\end{align}
With these, there are two relations yet to show.

\paragraph{Joint Interactions vs Observational}
First, consider the joint interaction ($I^t_1=I^t_2=1$) versus the pure observational regime ($I^t_1=I^t_2=0$). 
We prove that these two distributions must be different by contradiction. We first assume that they are equal and show that a contradiction strictly follows.
With both equations equal, we can write:
\begin{align}
    p(C^{t}_1|C^{t-1},I^{t}_1=0) \cdot p(C^{t}_2|C^{t-1},I^{t}_2=0) &= p(C^{t}_1|C^{t-1},I^{t}_1=1) \cdot p(C^{t}_2|C^{t-1},I^{t}_2=1) \\
    \label{eq:1942049}
    \Leftrightarrow \frac{p(C^{t}_1|C^{t-1},I^{t}_1=0)}{p(C^{t}_1|C^{t-1},I^{t}_1=1)} &= \frac{p(C^{t}_2|C^{t-1},I^{t}_2=1)}{p(C^{t}_2|C^{t-1},I^{t}_2=0)}
\end{align}
Note that the third step is possible since $p(C^{t}_i|C^{t-1},I^t_i=1), p(C^{t}_i|C^{t-1},I^t_i=0)$ share the same support. 
Further, since $C^{t}_1$ and $C^{t}_2$ are conditionally independent, the equality above must hold for any values of $C^{t}_1,C^{t}_2$. 
This implies that, for a given $C^{t}_2$, the fraction of $C^{t}_1$ must be constant, and vice versa.
Denoting this constant factor with $c$, we can rewrite the previous equation as:
\begin{align}
    \frac{p(C^{t}_1|C^{t-1},I^{t}_1=0)}{p(C^{t}_1|C^{t-1},I^{t}_1=1)} &= c \\
    \Leftrightarrow p(C^{t}_1|C^{t-1},I^{t}_1=0) &= c \cdot p(C^{t}_1|C^{t-1},I^{t}_1=1) \\
    \Leftrightarrow \int p(C^{t}_1|C^{t-1},I^{t}_1=0) dC^{t}_1 &= \int c \cdot p(C^{t}_1|C^{t-1},I^{t}_1=1) dC^{t}_1 \\
    \Leftrightarrow 1 &= c
\end{align}
In the last step, the two integrals disappear since both $p(C^{t}_1|C^{t-1},I^{t}_1=0)$ and $p(C^{t}_1|C^{t-1},I^{t}_1=1)$ are valid probability density functions.
Hence, the equality can only be valid if $c=1$, which implies $p(C^{t}_1|C^{t-1},I^{t}_1=1)=p(C^{t}_1|C^{t-1},I^{t}_1=0)$. 
However, this equality of distributions is out ruled by the assumptions of \cref{theo:main_theorem_no_gaus} as discussed in the beginning of this section, and thus causes a contradiction.
In other words, this shows that the joint interaction ($I^t_1=I^t_2=1$) and the pure observational regime ($I^t_1=I^t_2=0$) must be strictly different, \ie{}:
\begin{equation}
    p(C_1^{t},C_2^{t}|C^{t-1},I^{t}_1=0,I^{t}_2=0) \neq 
    p(C_1^{t},C_2^{t}|C^{t-1},I^{t}_1=1,I^{t}_2=1)
\end{equation}

\paragraph{Single-target vs Single-target}
The final step is to show that the distribution for interacting on $C_1$ versus the distribution of interacting on $C_2$ must be different.
For this, we can use a similar strategy as for the previous comparison and perform a proof of contradiction.
If both of the distributions are equal, the following equation follows:
\begin{align}
    \Leftrightarrow p(C^{t}_1|C^{t-1},I^{t}_1=0) \cdot 
    p(C^{t}_2|C^{t-1},I^{t}_2=1) &= p(C^{t}_1|C^{t-1},I^{t}_1=1) \cdot 
    p(C^{t}_2|C^{t-1},I^{t}_2=0) \\
    \Leftrightarrow \frac{p(C^{t}_1|C^{t-1},I^{t}_1=0)}{p(C^{t}_1|C^{t-1},I^{t}_1=1)} &= \frac{p(C^{t}_2|C^{t-1},I^{t}_2=0)}{p(C^{t}_2|C^{t-1},I^{t}_2=1)}
\end{align}
This is almost identical to \cref{eq:1942049}, besides the flipped fraction for $C_2$. 
Note, however, that the same implications hold, namely that both fractions need to be constant and constant with value 1.
This again contradicts our assumptions, and proves that the two distributions must be different:
\begin{equation}
    p(C_1^{t},C_2^{t}|C^{t-1},I^{t}_1=0,I^{t}_2=1) \neq 
    p(C_1^{t},C_2^{t}|C^{t-1},I^{t}_1=1,I^{t}_2=0)
\end{equation}

\paragraph{Conclusion} 
In summary, we have shown that the four possible cases of interactions strictly model different distributions.
Further, this distinction between the four cases can only be obtained by information from $R^{t}$ through $I^{t}$, since $I^{t}$ cannot be a deterministic function of the previous time step.
Thus, any possible representation of the variables $C_1,C_2$ must model the same four (or at least three) possible interaction settings.

\subsubsection{Any representation requires the same interaction cases - multi-variable case}
\label{sec:proof_steps_same_intv_cases_multivar}

So far, we have discussed the interaction cases for two variables.
This discussion can be easily extended to cases of three or more variables.
Before doing so, we formally state the lemma we are proving in this step.
\newcommand{\intvvala}{a^I}
\newcommand{\intvvalb}{b^I}

\begin{lemma}
    \label{lemma:proof_intervention_cases}
    For the interaction variables $I^{t}=\{I_1^{t},...,I_K^{t}\}$ in a causal model $\mathcal{M}$, any two values $\intvvala,\intvvalb\in\{0,1\}^K$ with $\intvvala\neq \intvvalb$ must strictly model different distributions:
    $$\forall C^t\colon p(C^{t}|C^{t-1},I^{t}=\intvvala)\neq p(C^{t}|C^{t-1},I^{t}=\intvvalb)$$
\end{lemma}

\begin{proof}
For $K$ variables, we can write the overall joint distribution $p(C^{t}|C^{t-1},I^{t})$ as:
\begin{align}
    p(C^{t}|C^{t-1},I^{t}) = p(C^{t}_1|C^{t-1},I^{t}_1) \cdot p(C^{t}_2|C^{t-1},I^{t}_2) \cdot ... \cdot p(C^{t}_K|C^{t-1},I^{t}_K)
\end{align}
Consider now the distributions for two different, arbitrary interaction values $\intvvala, \intvvalb$ ($\intvvala\neq \intvvalb$): 
\begin{align}
    \label{eq:same_intv_cases_genK_aI_bI}
    p(C^{t}|C^{t-1},I^{t}=\intvvala) &\stackrel{?}{=} 
    p(C^{t}|C^{t-1},I^{t}=\intvvalb) \\
    p(C^{t}_1|C^{t-1},I^{t}_1=a^I_1) \cdot ... \cdot p(C^{t}_K|C^{t-1},I^{t}_K=a^I_K) &= p(C^{t}_1|C^{t-1},I^{t}_1=b^I_1) \cdot ... \cdot p(C^{t}_K|C^{t-1},I^{t}_K=b^I_K)
\end{align}
We can rewrite this equation as:
\begin{align}
    \label{eq:same_intv_cases_genK_fractions}
    \frac{p(C^{t}_1|C^{t-1},I^{t}_1=\intvvala_1)}{p(C^{t}_1|C^{t-1},I^{t}_1=\intvvalb_1)}\cdot ... \cdot \frac{p(C^{t}_K|C^{t-1},I^{t}_K=\intvvala_K)}{p(C^{t}_K|C^{t-1},I^{t}_K=\intvvalb_K)} & = 1
\end{align}
Similar to our discussion on two variables, we can now analyze this equation under the situation where we keep all variables fixed up to $C^{t+1}_i$.
This is a valid scenario since all variables are independent based on their conditioning set $C^{t-1}$ and $a^I$/$b^I$.
This implies that the fraction of $C_i$ in \cref{eq:same_intv_cases_genK_fractions} must be equals to one divided by the multiplication of the remaining fractions, which we considered constant.
With this, we have the following equation:
\begin{align}
    \frac{p(C^{t}_i|C^{t-1},I^{t}_i=\intvvala_i)}{p(C^{t}_i|C^{t-1},I^{t}_i=\intvvalb_i)} & = c
\end{align}
where $c$ again summarizes all constant terms.
As shown earlier in this section, this equality can only hold if $\intvvala_i=\intvvalb_i$.
In turn, this mean that \cref{eq:same_intv_cases_genK_aI_bI} can only be an equality if $\intvvala = \intvvalb$.
Hence, different interaction cases must strictly model different distributions.
\end{proof}

\subsubsection{Alignment of interaction variables - 2 variables}
\label{sec:proof_steps_align_intv_cases}

\paragraph{Setup}
In the previous section, we have proven that any representation needs to model the same interaction cases. 
The next step is to show that the interaction cases further need to align, \ie{} the interaction variables must be equivalent up to permutation and sign flips.
For this, consider an alternative representation, $\hat{C}_1,\hat{C}_2$, which is the result of an invertible change-of-variables operation.
We denote the corresponding interaction variables by $\hat{I}_1,\hat{I}_2$.
Overall, we can write their probability distribution as:
\begin{align}
    p(C_1^{t},C_2^{t}|C^{t-1},I^{t}_1,I^{t}_2) &= p(\hat{C}_1^{t},\hat{C}_2^{t}|C^{t-1},\hat{I}^{t}_1,\hat{I}^{t}_2) \cdot |\det \bm{J}| \\
    \label{eq:proof_intv_case_align_orig_eq}
    p(C^{t}_1|C^{t-1},I^{t}_1) \cdot p(C^{t}_2|C^{t-1},I^{t}_2) &= 
    p(\hat{C}^{t}_1|C^{t-1},\hat{I}^{t}_1) \cdot 
    p(\hat{C}^{t}_2|C^{t-1},\hat{I}^{t}_2) \cdot |\det \bm{J}|
\end{align}
For simplicity, we write the conditioning of $\hat{C}_1^{t},\hat{C}_2^{t}$ still in terms of $C^{t-1}$, since $C^{t-1}$ and $\hat{C}^{t-1}$ contain the same information.
Further, $\bm{J}$ represents the Jacobian of the invertible transformation of $C^{t}_1,C^{t}_2$ to $\hat{C}^{t}_1,\hat{C}^{t}_2$.

\paragraph{Cases to consider}
Now, our goal is to show that $\hat{I^t}_1,\hat{I^t}_2$ must be equivalent to $I_1,I_2$ up to permutation and sign flip.
As an example, consider a value of $C^{t-1}$ under which we may have four possible values of $R^{t}$ which give us the following interactions:
\begin{center}
    \begin{tabular}{ccccc}
        \toprule
        $\bm{R}$ & $\bm{I_1}$ & $\bm{I_2}$ & $\bm{\hat{I}_1}$ & $\bm{\hat{I}_2}$\\
        \midrule
        $r_1$ & 0 & 0 & 0 & 0 \\
        $r_2$ & 1 & 0 & 1 & 0 \\
        $r_3$ & 0 & 1 & 1 & 1 \\
        $r_4$ & 1 & 1 & 0 & 1 \\
        \bottomrule
    \end{tabular}
\end{center}
with $r_1,r_2,r_3,r_4\in\mathcal{R}, r_1\neq r_2\neq r_3\neq r_4$.
In the notation of intervention design, one can interpret these different interactions are different \textit{experiments}, \ie{} different sets of variables that are jointly intervened.
We will denote them with $E_1,...,E_4$ where $E_i=[I_1,I_2]$ for a given $r_i$.
Similarly, we will use $\hat{E}_i=[\hat{I}_1,\hat{I}_2]$ to denote the same set for the alternative representation.

In this setup, we say that $I_2$ aligns with $\hat{I}_2$, since they are equal in all experiments $E_1,...,E_4$ / for all values of $R^t$. 
However, $I_1$ does not align with any interaction variable of $\hat{C}$, because $I_1\neq \hat{I}_1$ and $I_1\neq 1-\hat{I}_1$, and same for $\hat{I}_2$.
Thus, we are aiming to derive that this setup contradicts \cref{eq:proof_intv_case_align_orig_eq}.

\paragraph{Single-target vs Joint interaction}
We start the analysis by writing down all distributions to compare:
\begin{align}
    \label{eq:proof_intv_case_align_0}
    p(C^{t}_1|C^{t-1},I^{t}_1=0) \cdot p(C^{t}_2|C^{t-1},I^{t}_2=0) &= 
    p(\hat{C}^{t}_1|C^{t-1},\hat{I}^{t}_1=0) \cdot 
    p(\hat{C}^{t}_2|C^{t-1},\hat{I}^{t}_2=0) \cdot |\det \bm{J}| \\
    \label{eq:proof_intv_case_align_1}
    p(C^{t}_1|C^{t-1},I^{t}_1=1) \cdot p(C^{t}_2|C^{t-1},I^{t}_2=0) &= 
    p(\hat{C}^{t}_1|C^{t-1},\hat{I}^{t}_1=1) \cdot 
    p(\hat{C}^{t}_2|C^{t-1},\hat{I}^{t}_2=0) \cdot |\det \bm{J}| \\
    \label{eq:proof_intv_case_align_2}
    p(C^{t}_1|C^{t-1},I^{t}_1=0) \cdot p(C^{t}_2|C^{t-1},I^{t}_2=1) &= 
    p(\hat{C}^{t}_1|C^{t-1},\hat{I}^{t}_1=1) \cdot 
    p(\hat{C}^{t}_2|C^{t-1},\hat{I}^{t}_2=1) \cdot |\det \bm{J}| \\
    \label{eq:proof_intv_case_align_3}
    p(C^{t}_1|C^{t-1},I^{t}_1=1) \cdot p(C^{t}_2|C^{t-1},I^{t}_2=1) &= 
    p(\hat{C}^{t}_1|C^{t-1},\hat{I}^{t}_1=0) \cdot 
    p(\hat{C}^{t}_2|C^{t-1},\hat{I}^{t}_2=1) \cdot |\det \bm{J}|
\end{align}
Our overall proof strategy is to derive relations between individual variables, \eg{} $C_1$ and $\hat{C}_1$.
Since the invertible transformation between $C$ and $\hat{C}$ must be independent of $R$, $I$ and $\hat{I}$, the relations we derive must hold across all the experiments.
By dividing the sets of equations, we obtain:
\begin{align}
    \label{eq:proof_intv_case_align_10}
    \text{Eq \ref{eq:proof_intv_case_align_1} / Eq \ref{eq:proof_intv_case_align_0}}&: \hspace{3mm} 
    \frac{p(C^{t}_1|C^{t-1},I^{t}_1=1)}{p(C^{t}_1|C^{t-1},I^{t}_1=0)} = \frac{p(\hat{C}^{t}_1|C^{t-1},\hat{I}^{t}_1=1)}{p(\hat{C}^{t}_1|C^{t-1},\hat{I}^{t}_1=0)} \\
    \label{eq:proof_intv_case_align_20}
    \text{Eq \ref{eq:proof_intv_case_align_2} / Eq \ref{eq:proof_intv_case_align_0}}&: \hspace{3mm} 
    \frac{p(C^{t}_2|C^{t-1},I^{t}_2=1)}{p(C^{t}_2|C^{t-1},I^{t}_2=0)} = 
    \frac{p(\hat{C}^{t}_1|C^{t-1},\hat{I}^{t}_1=1)}{p(\hat{C}^{t}_1|C^{t-1},\hat{I}^{t}_1=0)}
    \frac{p(\hat{C}^{t}_2|C^{t-1},\hat{I}^{t}_2=1)}{p(\hat{C}^{t}_2|C^{t-1},\hat{I}^{t}_2=0)} \\
    \label{eq:proof_intv_case_align_30}
    \text{Eq \ref{eq:proof_intv_case_align_3} / Eq \ref{eq:proof_intv_case_align_0}}&: \hspace{3mm} 
    \frac{p(C^{t}_1|C^{t-1},I^{t}_1=1)}{p(C^{t}_1|C^{t-1},I^{t}_1=0)}
    \frac{p(C^{t}_2|C^{t-1},I^{t}_2=1)}{p(C^{t}_2|C^{t-1},I^{t}_2=0)} = 
    \frac{p(\hat{C}^{t}_2|C^{t-1},\hat{I}^{t}_2=1)}{p(\hat{C}^{t}_2|C^{t-1},\hat{I}^{t}_2=0)}
\end{align}
Note that the Jacobian, $|\det \bm{J}|$, cancels out in all distributions since it is independent of the interactions and thus identical for all equations above.
As a next step, we replace $\hat{C}_2$ in \cref{eq:proof_intv_case_align_20} with the result of \cref{eq:proof_intv_case_align_30} and rearrange the terms:
\begin{align}
    \label{eq:proof_intv_case_align_2030}
    \frac{p(C^{t}_1|C^{t-1},I^{t}_1=0)}{p(C^{t}_1|C^{t-1},I^{t}_1=1)} = \frac{p(\hat{C}^{t}_1|C^{t-1},\hat{I}^{t}_1=1)}{p(\hat{C}^{t}_1|C^{t-1},\hat{I}^{t}_1=0)}
\end{align}
Similarly, replacing $\hat{C}_1$ in \cref{eq:proof_intv_case_align_10} with the new result in \cref{eq:proof_intv_case_align_2030}, we obtain:
\begin{align}
    \label{eq:proof_intv_case_align_102030}
    \frac{p(C^{t}_1|C^{t-1},I^{t}_1=1)}{p(C^{t}_1|C^{t-1},I^{t}_1=0)} = \frac{p(C^{t}_1|C^{t-1},I^{t}_1=0)}{p(C^{t}_1|C^{t-1},I^{t}_1=1)}
\end{align}
This equation can obviously only hold if both fractions are equal to 1.
However, as shown in \cref{sec:proof_steps_same_intv_cases}, this contradicts our assumptions of the theorem.
Thus, we have shown that the interaction variables $\hat{I}_1,\hat{I}_2$ cannot model the same distribution as $I_1,I_2$.
For the specific example of two causal variables and four experiments, it turns out that there exists no other set of interaction variables that would not align to $I_1,I_2$.
Hence, in this case, any other valid representation $\hat{C}$ which fulfills \cref{eq:proof_intv_case_align_orig_eq} must have interaction variables that align with the true model.

\paragraph{Conclusion}
This example is meant to communicate the general intuition behind our proof strategy for showing that the interaction variables between the true causal model $\mathcal{M}$ and a learned representation $\widehat{\mathcal{M}}$ align.
We note that this example does not cover all possible models with two causal variables $C_1,C_2$, since our assumptions only require $\lfloor \log_2 2\rfloor + 2=3$ experiments/different values of $R^t$, while we considered here four for simplicity.
For this smaller amount of experiments, it becomes difficult to distinguish between models that model the true interaction variables $I_1,...,I_K$ and possible linear combinations of such.
This can be prevented by ensuring sufficient variability either in the dynamics (condition (A) - \cref{theo:main_theorem_no_gaus}) or over time (condition (B) - \cref{theo:main_theorem_time_var}), which we show in the next two subsections.

\subsubsection{Alignment of interaction variables - Multi-variable case (condition (A) - \cref{theo:main_theorem_no_gaus})}
\label{sec:proof_steps_align_intv_cases_multivar_no_gaus}

We start with showing the interaction variable alignment under condition (A) of \cref{theo:main_theorem_no_gaus}.
The goal is to prove the following lemma:

\begin{lemma}[Dynamics Variability]
    \label{lemma:proof_align_interventions_option_1}
    For any variable $C_k$ ($k=1,...,K$) with interaction variable $I_k$, there exist exactly one variable $\hat{C}_l$ with interaction variable $\hat{I}_l$, which models the same interaction pattern:
    \begin{equation*}
        \forall C^{t}\colon I^{t}_k=\hat{I}^{t}_l\hspace{2mm}\text{or}\hspace{2mm}I^{t}_k=1-\hat{I}^{t}_l
    \end{equation*}
    if the second derivative of the log-difference between the observational and the interactional distribution is not constantly zero:
    \begin{equation*}
        \forall C^{t}_k, \exists C^{t-1}\colon \frac{\partial^2 \intvdiv(C^{t}_k|C^{t-1})}{\partial (C^{t}_k)^2} \neq 0
    \end{equation*}
\end{lemma}

\begin{proof}
We structure the proof in four main steps. 
First, we generalize our analysis of the relations between interaction equations from the two variables to the multi-variable case.
We then take a closer look at them from two sides: a variable from the true causal model, $C^{t}_m$, and a variable from the alternative representation, $\hat{C}^{t}_l$.
The intuition behind the proof is that a change in $C^{t}_m$ must correspond to a change in $\hat{C}^{t}_l$ which appears in the same set of equations.
This inherently requires that $C^{t}_m$ and $\hat{C}^{t}_l$ share the same unique interaction pattern.
With this intuition in mind, the following paragraphs detail these individual proof steps.

\pseudoparagraph{Equations sets implied by interactions}
Firstly, we consider a set of $Q$ true interaction experiments $E_1,...,E_Q$, \ie{} $Q$ different values of $R^{t}$ which cause different sets of interaction variable values $I^{t}_1,...,I^{t}_K$, and similarly the $Q$ values of $R^{t}$ in the alternative representation space $\hat{I}$ with experiments $\hat{E}_1,...,\hat{E}_Q$.
In the previous example of the two variables, the experiments would be $E_1=[0,0], E_2=[1,0], E_3=[0,1], E_4=[1,1]$ and $\hat{E}_1=[0,0], \hat{E}_2=[1,0], \hat{E}_3=[1,1], \hat{E}_4=[0,1]$.
We will denote the interaction variable value $I_k$ of the causal variable $C_k$ in the experiment $E_i$ with $E_i^{k}$, \ie{} $E_{2}^1=1$ in the previous example.

For any two experiments $E_i,E_j$, there exists a set of variables for which the interaction targets differ.
We summarize the indices of these variables as $\mathcal{V}_{ij}$, and similarly for the alternative representation $\mathcal{\hat{V}}_{ij}$.
Taking the two-variable example again, $\mathcal{V}_{12}=\{1\}$, \ie{} the interaction variable of the causal variable $C_1$ differs between $E_1$ and $E_2$.
Using this notation, we can write the division of two experiments $E_i,E_j$ as:
\begin{equation}
    \label{eq:proof_intv_case_align_general_eq_div}
    \prod_{k\in\mathcal{V}_{ij}} \frac{p(C_k^{t}|C^{t-1},I_k^{t}=E_i^{k})}{p(C_k^{t}|C^{t-1},I_k^{t}=E_j^{l})} = \prod_{l\in\mathcal{\hat{V}}_{ij}} \frac{p(\hat{C}_l^{t}|C^{t-1},\hat{I}_l^{t}=\hat{E}_i^{l})}{p(\hat{C}_l^{t}|C^{t-1},\hat{I}_l^{t}=\hat{E}_j^{l})}
\end{equation}

\pseudoparagraph{Analyzing equations for individual causal variables}
The experiments imply a set of $\frac{(Q-1)(Q-2)}{2}$ equations. 
Our next step is to analyze what these equations imply for an individual causal variable $C^{t}_m$.
First, we take the log on both sides to obtain:
\begin{align}
    \sum_{k\in\mathcal{V}_{ij}} \log \frac{p(C_k^{t}|C^{t-1},I_k^{t}=E_i^{k})}{p(C_k^{t}|C^{t-1},I_k^{t}=E_j^{k})} & = \sum_{l\in\mathcal{\hat{V}}_{ij}} \log \frac{p(\hat{C}_l^{t}|C^{t-1},\hat{I}_l^{t}=\hat{E}_i^{l})}{p(\hat{C}_l^{t}|C^{t-1},\hat{I}_l^{t}=\hat{E}_j^{l})}
\end{align}
For readability, we adapt our notation of $\intvdiv(C^{t}_k|C^{t-1})$ here by having:
\begin{align}
    \intvdiv_{ij}(C^{t}_k|C^{t-1}) & = \log \frac{p(C_k^{t}|C^{t-1},I_k^{t}=E_i^{k})}{p(C_k^{t}|C^{t-1},I_k^{t}=E_j^{k})} \\
    \intvdiv_{ij}(\hat{C}^{t}_l|C^{t-1}) & = \log \frac{p(\hat{C}_l^{t}|C^{t-1},\hat{I}_l^{t}=\hat{E}_i^{l})}{p(\hat{C}_l^{t}|C^{t-1},\hat{I}_l^{t}=\hat{E}_j^{l})}
\end{align}
which gives us
\begin{align}
    \sum_{k\in\mathcal{V}_{ij}} \intvdiv_{ij}(C^{t}_k|C^{t-1}) & = \sum_{l\in\mathcal{\hat{V}}_{ij}} \intvdiv_{ij}(\hat{C}^{t}_l|C^{t-1})
\end{align}
Now consider a single variable $C^{t}_m$, for which $m\in \mathcal{V}_{ij}$.
If we take the derivative with respect to $C^{t}_m$, we get:
\begin{align}
    \label{eq:proof_align_deriv_1}
    \frac{\partial \intvdiv_{ij}(C^{t}_m|C^{t-1})}{\partial C^{t}_m} & = \sum_{l\in\mathcal{\hat{V}}_{ij}} \frac{\partial \intvdiv_{ij}(\hat{C}^{t}_l|C^{t-1})}{\partial C^{t}_m}
\end{align}
The sum on the left drops away since we know that $C^{t}_k\independent C^{t}_m\mid C^{t-1},I^t$, and therefore $\frac{\partial \intvdiv_{ij}(C^{t}_k|C^{t-1})}{\partial C^{t}_m}$ if $k\neq m$.

For each variable $C^{t}_m$, we obtain at least $Q-1$ equations ($Q$ being the number of overall interaction experiments) since every experiment $E_i$ must have at least one experiment $E_j$ for which $E_i^{m}\neq E_j^{m}$, since otherwise the interaction variable $I_m$ must be equal in all experiments and thus a constant, violating our distinct interaction pattern assumption. 
In other words, we obtain a set of experiment pairs which differ in the interaction variable of $C^{t}_m$, \ie{} $\bm{\mathcal{V}}_m = \{\mathcal{\hat{V}}_{ij} \mid i,j\in\range{1}{Q}, E_i^{m}\neq E_j^{m}\}$ with $|\bm{\mathcal{V}}_m|\geq Q-1$.

For two experiment equations, $\mathcal{\hat{V}}_{ij}, \mathcal{\hat{V}}_{sr} \in \bm{\mathcal{V}}_m$, we have the following equality following from \cref{eq:proof_align_deriv_1}:
\begin{align}
    \frac{\partial \intvdiv_{ij}(C^{t}_m|C^{t-1})}{\partial C^{t}_m} = \sum_{l\in\mathcal{\hat{V}}_{ij}} \frac{\partial \intvdiv_{ij}(\hat{C}^{t}_l|C^{t-1})}{\partial C^{t}_m} & = \sum_{w\in\mathcal{\hat{V}}_{sr}} \frac{\partial \intvdiv_{sr}(\hat{C}^{t}_w|C^{t-1})}{\partial C^{t}_m}
\end{align}
Using $\intvdiv_{ij}(C^{t}_k|C^{t-1}) = -\intvdiv_{ji}(C^{t}_k|C^{t-1})$, we can align the equations above via:
\begin{align}
    \intvdiv(C^{t}_k|C^{t-1}) & = \frac{p(C_k^{t}|C^{t-1},I_k^{t}=1)}{p(C_k^{t}|C^{t-1},I_k^{t}=0)} \\
    \label{eq:proof_align_deriv_2}
    \sum_{l\in\mathcal{\hat{V}}_{ij}} \frac{\partial \intvdiv(\hat{C}^{t}_l|C^{t-1})}{\partial C^{t}_m} & = (-1)^{\mathbbm{1}[E^m_i = E^m_s]}\sum_{w\in\mathcal{\hat{V}}_{sr}} \frac{\partial \intvdiv(\hat{C}^{t}_w|C^{t-1})}{\partial C^{t}_m}
\end{align}

\pseudoparagraph{Analyzing equations for a single variable of alternative representation}
As the next step, we analyze the derivatives of individual variables of the alternative representation $\hat{C}$ in \cref{eq:proof_align_deriv_2}. 
Consider a variable $\hat{C}^{t}_l, l\in \mathcal{\hat{V}}_{ij}$.
Taking the derivative of \cref{eq:proof_align_deriv_2} with respect to $\hat{C}^{t}_l$, the left-hand side simplifies to only the $\hat{C}^{t}_l$ since for all other variables, we have that $\hat{C}^{t}_l\independent \hat{C}^{t}_{l'}\mid C^{t-1},\hat{I}^t$.
The right-hand side, however, has two options:
\begin{align}
    \frac{\partial^2 \intvdiv(C^{t}_m|C^{t-1})}{\partial C^{t}_m \partial \hat{C}^{t}_l} = \frac{\partial^2 \intvdiv(\hat{C}^{t}_l|C^{t-1})}{\partial C^{t}_m \partial \hat{C}^{t}_l} & = (-1)^{\mathbbm{1}[E^m_i = E^m_s]}\sum_{w\in\mathcal{\hat{V}}_{sr}}\begin{cases}
        0 & \text{if } w \neq l \\
        \frac{\partial^2 \intvdiv(\hat{C}^{t}_l|C^{t-1})}{\partial C^{t}_m \partial \hat{C}^{t}_l} & \text{if } w = l
    \end{cases} \\
    & = \begin{cases}
        0 & \text{if } l \not\in \mathcal{\hat{V}}_{sr} \\
        (-1)^{\mathbbm{1}[E^m_i = E^m_s]}\frac{\partial^2 \intvdiv(\hat{C}^{t}_l|C^{t-1})}{\partial C^{t}_m \partial \hat{C}^{t}_l} & \text{if } l \in \mathcal{\hat{V}}_{sr}
    \end{cases}
\end{align}
If $E^m_i \neq E^m_s$, we have an equation similar to $c = -c$, which can only be solved via $c=0$. 
Therefore, we can further simplify the equation to:
\begin{align}
    \label{eq:proof_align_deriv_3}
    \frac{\partial^2 \intvdiv(\hat{C}^{t}_l|C^{t-1})}{\partial C^{t}_m \partial \hat{C}^{t}_l} & = \begin{cases}
        0 & \text{if } l \not\in \mathcal{\hat{V}}_{sr} \text{ or } E^m_i \neq E^m_s\\
        \frac{\partial^2 \intvdiv(\hat{C}^{t}_l|C^{t-1})}{\partial C^{t}_m \partial \hat{C}^{t}_l} & \text{otherwise}
    \end{cases}
\end{align}

\pseudoparagraph{Plugging everything together}
From \cref{eq:proof_align_deriv_3}, we can make the following conclusions: for any variable $\hat{C}^{t}_l$ which is not in \textit{all} experiment pairs of $\bm{\mathcal{V}}_m$, its second derivative $\frac{\partial^2 \intvdiv(\hat{C}^{t}_l|C^{t-1})}{\partial C^{t}_m \partial \hat{C}^{t}_l}$ must be zero.
This is an important insight, since we know that all second derivatives equations must still equal to $\frac{\partial^2 \intvdiv(C^{t}_m|C^{t-1})}{\partial C^{t}_m \partial \hat{C}^{t}_l}$.
Using the chain rule, we can relate these second derivatives even further:
\begin{align}
    \frac{\partial^2 \intvdiv(C^{t}_m|C^{t-1})}{\partial C^{t}_m \partial \hat{C}^{t}_l} &= \frac{\partial^2 \intvdiv(C^{t}_m|C^{t-1})}{\partial^2 C^{t}_m}\bm{J}^{-1}_{ml} \\
    \frac{\partial^2 \intvdiv(C^{t}_m|C^{t-1})}{\partial^2 C^{t}_m}\bm{J}^{-1}_{ml} &= \frac{\partial^2 \intvdiv(\hat{C}^{t}_l|C^{t-1})}{\partial C^{t}_m \partial \hat{C}^{t}_l}
\end{align}
where $\bm{J}^{-1}_{ml}$ is the $ml$-th entry of the inverse of the Jacobian, \ie{} $\bm{J}^{-1}_{ml}=\frac{\partial C^{t}_m}{\hat{C}^{t}_l}$. 
From our assumptions, we know that $\frac{\partial^2 \intvdiv(C^{t}_m|C^{t-1})}{\partial^2 C^{t}_m}$ cannot be constant zero for all values of $C^{t}_m$.
Therefore, if $\frac{\partial^2 \intvdiv(\hat{C}^{t}_l|C^{t-1})}{\partial C^{t}_m \partial \hat{C}^{t}_l}$ is zero following \cref{eq:proof_align_deriv_3}, then this must strictly imply that $\bm{J}^{-1}_{ml}$ must be constantly zero, \ie{} $C^{t}_m$ and $\hat{C}^{t}_l$ are independent.

However, at the same time, we know that $\bm{J}^{-1}_{ml}$ cannot be constantly zero for all $l$ since otherwise, $\bm{J}^{-1}$ (and therefore $\bm{J}$) has a zero determinant and thus the transformation between $C$ and $\hat{C}$ cannot be invertible.
Therefore, in order for $\hat{C}$ to be a valid transformation, there must exist at least one variable $\hat{C}_l$ which is in all experiment sets $\bm{\mathcal{V}}_m$.
This implies that for this variable $\hat{C}_l$ and our original causal variable $\hat{C}_m$, the following relations must hold:
\begin{align}
    E_i^k = E_j^k & \Leftrightarrow \hat{E}_i^l = \hat{E}_j^l
\end{align}
This inherently implies that for any variable $C_k$, there must exist at least one variable $\hat{C}_l$, for which the following must hold:
\begin{equation}
    \forall i, E_i^k = \hat{E}_i^l\hspace{5mm}\text{or}\hspace{5mm}\forall i, E_i^k = 1-\hat{E}_i^l
\end{equation}
Finally, since for every variable $C_k$, the set of experiments is unique, \ie{} no deterministic function between $I_k$ and any other interaction variable $I_j$, and the alternative representation has the same number of variables, it implies that there exists a 1-to-1 match between an interaction variable $I_k$ and in the alternative representation $\hat{I}_k$.
This proves our original lemma.
\end{proof}

\subsubsection{Alignment of interaction variables - Multi-variable case  (condition (B) - \cref{theo:main_theorem_time_var})}
\label{sec:proof_steps_align_intv_cases_multivar_time_var}

\begin{lemma}[Time-variability]
    \label{lemma:proof_align_interventions_option_2}
    For any variable $C_k$ ($k=1,...,K$) with interaction variable $I_k$, there exist exactly one variable $\hat{C}_l$ with interaction variable $\hat{I}_l$, which models the same interaction pattern:
    \begin{equation*}
        \forall C^{t}\colon I^{t}_k=\hat{I}^{t}_l\hspace{2mm}\text{or}\hspace{2mm}I^{t}_k=1-\hat{I}^{t}_l
    \end{equation*}
    if for any $C^{t}$, there exist $K+1$ different values $c^1,...,c^{K+1}$ for $C^{t}$ for which the vectors $v_{1},...,v_{K}$ of the following structure are linearly independent:
    \begin{equation*}
        v_{i} = \begin{bmatrix}
            \frac{\partial \intvdiv\left(C^{t}_i|C^{t-1}=c^1\right)}{\partial C^{t}_i} & 
            \frac{\partial \intvdiv\left(C^{t}_i|C^{t-1}=c^2\right)}{\partial C^{t}_i} & 
            \cdots &
            \frac{\partial \intvdiv\left(C^{t}_i|C^{t-1}=c^{K+1}\right)}{\partial C^{t}_i} \\ 
        \end{bmatrix}^T \in \mathbb{R}^K
    \end{equation*}
\end{lemma}

\begin{proof}
    We follow the same proof as for Lemma~\ref{lemma:proof_align_interventions_option_1} up until \cref{eq:proof_align_deriv_1}, where for each variable $C_m$, we have obtained the following equation:
    \begin{align}
        \label{eq:proof_align_deriv_1_option_2}
        \frac{\partial \intvdiv_{ij}(C^{t}_m|C^{t-1})}{\partial C^{t}_m} & = \sum_{l\in\mathcal{\hat{V}}_{ij}} \frac{\partial \intvdiv_{ij}(\hat{C}^{t}_l|C^{t-1})}{\partial C^{t}_m}
    \end{align}
    Here, we rewrite the derivative $\frac{\partial \intvdiv_{ij}(\hat{C}^{t}_l|C^{t-1})}{\partial C^{t}_m}$ using the chain rule to:
    \begin{align}
        \frac{\partial \intvdiv_{ij}(\hat{C}^{t}_l|C^{t-1})}{\partial C^{t}_m} & = \frac{\partial \intvdiv_{ij}(\hat{C}^{t}_l|C^{t-1})}{\partial \hat{C}^{t}_l}\frac{\partial \hat{C}^{t}_l}{\partial C^{t}_m}\\
        & = \frac{\partial \intvdiv_{ij}(\hat{C}^{t}_l|C^{t-1})}{\partial \hat{C}^{t}_l}\bm{J}_{lm}
    \end{align}
    Therefore, we obtain:
    \begin{align}
        \label{eq:proof_align_deriv_1_option_2_ext}
        \frac{\partial \intvdiv_{ij}(C^{t}_m|C^{t-1})}{\partial C^{t}_m} & = \sum_{l\in\mathcal{\hat{V}}_{ij}} \frac{\partial \intvdiv_{ij}(\hat{C}^{t}_l|C^{t-1})}{\partial \hat{C}^{t}_l}\bm{J}_{lm}
    \end{align}
    Note hereby that $\bm{J}_{lm}$ is independent of the time index $t$ and particularly $C^{t-1}$, which will become important in the next steps of the proof.

    \pseudoparagraph{Alternative representation having linear independent vectors}
    Now consider the $K$ different vectors $v^1,...,v^{K}$, which are linearly independent. 
    For each of these individual vectors, we have at least $K$ equations of the form of \cref{eq:proof_align_deriv_1_option_2}, namely for each $C^{t}_1,...,C^{t}_K$.
    We can also express this in the form of a matrix product. 
    For that, we first stack the vectors $v^1,...,v^{K}$ into $\bm{V}\in\R^{(K+1)\times K}$:
    \begin{align}
        \label{eq:proof_align_option_2_matrix}
        \bm{V}=\begin{bmatrix}
            \vert &  & \vert \\
            v_1 & \cdots & v_{K}\\
            \vert &  & \vert \\
        \end{bmatrix} = 
        \begin{bmatrix}
            \frac{\partial \intvdiv\left(C^{t}_1|C^{t-1}=c^1\right)}{\partial C^{t}_1} & 
            \frac{\partial \intvdiv\left(C^{t}_2|C^{t-1}=c^1\right)}{\partial C^{t}_2} & \cdots & 
            \frac{\partial \intvdiv\left(C^{t}_K|C^{t-1}=c^1\right)}{\partial C^{t}_K}\\
            \frac{\partial \intvdiv\left(C^{t}_1|C^{t-1}=c^2\right)}{\partial C^{t}_1} & 
            \frac{\partial \intvdiv\left(C^{t}_2|C^{t-1}=c^2\right)}{\partial C^{t}_2} & \cdots & 
            \frac{\partial \intvdiv\left(C^{t}_K|C^{t-1}=c^2\right)}{\partial C^{t}_K}\\
            \vdots & \vdots & \ddots & \vdots\\
            \frac{\partial \intvdiv\left(C^{t}_1|C^{t-1}=c^{K+1}\right)}{\partial C^{t}_1} & 
            \frac{\partial \intvdiv\left(C^{t}_2|C^{t-1}=c^{K+1}\right)}{\partial C^{t}_2} & \cdots & 
            \frac{\partial \intvdiv\left(C^{t}_K|C^{t-1}=c^{K+1}\right)}{\partial C^{t}_K}\\ 
        \end{bmatrix}
    \end{align}
    We denote $\hat{\bm{V}}$ as the same matrix as in \cref{eq:proof_align_option_2_matrix}, just with each $C^{t}_i$ replaced with $\hat{C}^{t}_i$.
    Finally, we need to represent the factors of \cref{eq:proof_align_deriv_1_option_2_ext}. 
    Since these depend on a specific pair of experiments $E_i,E_j$, we pick for each variable $C_m^{t+1}$ an arbitrary pair of experiments, where $E^m_i=1$ and $E^m_j=0$.
    With this in mind, we can express the factors of \cref{eq:proof_align_deriv_1_option_2_ext} as:
    \begin{align}
        \bm{L} = \bm{J} \odot \begin{bmatrix}
            \delta^{E}_{11} & \delta^{E}_{12} & \cdots & \delta^{E}_{1K} \\
            \delta^{E}_{21} & \delta^{E}_{22} & \cdots & \delta^{E}_{2K} \\
            \vdots & \vdots & \ddots & \vdots \\
            \delta^{E}_{K1} & \delta^{E}_{K2} & \cdots & \delta^{E}_{KK} \\
        \end{bmatrix}
    \end{align}
    where $\bm{L}\in\R^{K\times K}$, $\odot$ is the Hamard product/element-wise product, and $\delta^{E}_{lm}=1$ if $\hat{E}^l_i=1,\hat{E}^l_j=0$ for the experiment pair $E_i,E_j$ picked for $C_m$, $0$ if $\hat{E}^l_i=\hat{E}^l_j$, and $-1$ otherwise.
    With that, we can express \cref{eq:proof_align_deriv_1_option_2_ext} in matrix form as:
    \begin{align}
        \label{eq:proof_align_option_2_simple_matrix_eq}
        \bm{V} = \hat{\bm{V}}\bm{L}
    \end{align}
    Since $\bm{V}$ has linearly independent columns and, based on \cref{eq:proof_align_option_2_simple_matrix_eq}, is equal to linear combinations of the columns of $\bm{\hat{V}}$, it directly follows that $\bm{\hat{V}}$ must also have linearly independent columns. 

    \pseudoparagraph{Solution to linear independent system} At the same time, we know that for each variable $C_m$, there exist pairs of experiments $E_i,E_j$ for which $E^m_i=E^m_j$. 
    We denote this set of experiment pairs by $\bm{\mathcal{\bar{V}}}^m = \{\mathcal{\bar{V}}_{ij} \mid i,j\in\range{1}{Q}, E_i^{m}= E_j^{m}\}$ with its size denoted as $Q_m=|\bm{\mathcal{\bar{V}}}^m|$.
    Each of these implies an equation like the following:
    \begin{equation}
        \label{eq:proof_align_deriv_1_option_2_ext_zero}
        0 = \sum_{l\in\mathcal{\bar{V}}_{ij}} \frac{\partial \intvdiv_{ij}(\hat{C}^{t}_l|C^{t-1})}{\partial \hat{C}^{t}_l}\bm{J}_{lm}
    \end{equation}
    We can, again, write it in matrix form to show this set of equations over the different temporal values $c^1,...,c^{K+1}$:
    \begin{align}
        \bm{0} = \hat{\bm{V}}\bm{\bar{L}}^m
    \end{align}
    where 
    \begin{align}
        \bm{\bar{L}}^m = \begin{bmatrix}
            \vert & & \vert \\
            \bm{J}_{\cdot m} & \cdot & \bm{J}_{\cdot m} \\
            \vert & & \vert \\
        \end{bmatrix} 
        \odot
        \begin{bmatrix}
            \mathbbm{1}[1\in \bm{\mathcal{\bar{V}}}^m_1] & \mathbbm{1}[1\in \bm{\mathcal{\bar{V}}}^m_2] & \cdots & \mathbbm{1}[1\in \bm{\mathcal{\bar{V}}}^m_{Q_m}] \\
            \mathbbm{1}[2\in \bm{\mathcal{\bar{V}}}^m_1] & \mathbbm{1}[2\in \bm{\mathcal{\bar{V}}}^m_2] & \cdots & \mathbbm{1}[2\in \bm{\mathcal{\bar{V}}}^m_{Q_m}] \\
            \vdots & \vdots & \ddots & \vdots \\
            \mathbbm{1}[K\in \bm{\mathcal{\bar{V}}}^m_1] & \mathbbm{1}[K\in \bm{\mathcal{\bar{V}}}^m_2] & \cdots & \mathbbm{1}[K\in \bm{\mathcal{\bar{V}}}^m_{Q_m}] \\
        \end{bmatrix}
    \end{align}
    Intuitively, $\bm{\bar{L}}^m$ lists out the $Q_m$ different equations of \cref{eq:proof_align_deriv_1_option_2_ext_zero} for variable $C_m$, duplicated for all possible temporal values $c^1,...,c^{K+1}$.
    Since all columns of $\bm{\hat{V}}$ are linearly independent, the only solution to the system is that $\bm{\bar{L}}^m=\bm{0}$, or in index form $\bm{\bar{L}}^m_{ij}=0$ for all $i=1,...,Q_m; j=1,...,K$.

    \pseudoparagraph{Matching of interaction variables}
    The fact that $\bm{\bar{L}}^m_{ij}=0$ must be zero means that one of its two matrix elements must have a zero entry. 
    Thus, for each variable $\hat{C}_l$, $\bm{J}_{lm}$ can only be non-zero if $l$ is not in any sets of $\bm{\mathcal{\bar{V}}}^m$.
    This implies that either $\hat{C}_l$ must follow the exact same interaction pattern as $C_m$, \ie{} $I_m=\hat{I}_l$ or $I_m=1-\hat{I}_l$, or $\hat{I}_l$ is a constant value.
    However, the constant value case can directly be excluded since this would imply $\bm{L}$ to have a zero determinant ($\delta^E_{l\cdot}=0$), which is not possible with $\det \bm{V} \neq 0$.
    At the same time, at least one value of $\bm{J}_{\cdot m}$ must be non-zero to ensure $\bm{J}$ to be invertible.
    Therefore, each variable $C_m$ must have one variable $\hat{C}_l$ for which $I_m=\hat{I}_l$ or $I_m=1-\hat{I}_l$.
    Finally, this match of $C_m,\hat{C}_l$ must be a unique since every variable $C_m$ has a different interaction pattern, and we are limited to $K$ variables $\hat{C}_1,...,\hat{C}_K$.
    With that, we have proven the initial lemma.

    \pseudoparagraph{Non-zero elements in Jacobian}
    Additionally to the lemma, this proof also shows that $\bm{J}$ must have exactly one non-zero value in each column and row, \ie{} being a permuted diagonal matrix.
\end{proof}

\subsubsection{Equivalence up to component-wise invertible transformations and permutation}
\label{sec:proof_steps_invertible_transformations}

With both Lemma~\ref{lemma:proof_align_interventions_option_1} and Lemma~\ref{lemma:proof_align_interventions_option_2}, we have shown that the two representations $C$ and $\hat{C}$ need to have the same interaction patterns.
Now, we are ready to prove the identifiability of the individual causal variables.
Since most of the results have been already shown in the previous proofs, we skip the intuition on the two-variable case and directly jump to the multi-variable case:

\begin{lemma}
    \label{lemma:proof_invertible_transformation}
    For any variable $C_k$ ($k=1,...,K$) with interaction variable $I_k$, there exist exactly one variable $\hat{C}_l$ with an invertible transformation $T_l$ for which the following holds:
    $$C^{t}_k = T_l(\hat{C}^{t}_l)$$
\end{lemma}

\begin{proof}
    We start with reiterating the initial result of \cref{sec:proof_steps_setup} stating that there exist an invertible transformation between $C$ and $\hat{C}$: $C=T(\hat{C})$.
    This also gives us the change of variables distribution:
    \begin{equation}
        \label{eq:proof_equiv_orig_prod}
        p(C^{t}|C^{t-1},I^t) = p(\hat{C}^{t}|C^{t-1},\hat{I}^t)|\det \bm{J}|
    \end{equation}
    Our goal is to show it follows for each $C_k$, there exists a $\hat{C}_l$ for which the following holds:
    \begin{equation}
        p(C_k^{t}|C^{t-1},I_k^t) = p(\hat{C}^{t}_l|C^{t-1},\hat{I}_l^t) |\bm{J}_{lk}|
    \end{equation}
    This change-of-variable equation implies that there exist an invertible transformation between $C_k$ and $\hat{C}_l$ with the scalar Jacobian $\bm{J}_{lk}$.

    \pseudoparagraph{Intermediate proof step based on Lemma~\ref{lemma:proof_align_interventions_option_1}}
    To prove this based on Lemma~\ref{lemma:proof_align_interventions_option_1}, we reuse our final results of the proof in \cref{sec:proof_steps_align_intv_cases_multivar_no_gaus}.
    Specifically, we have shown before that the inverse of the Jacobian $\bm{J}^{-1}_{kj}$ for the transformation from $C_k$ to $\hat{C}_j$ must be constantly zero if $C_k$ and $\hat{C}_j$ do not share the same interaction pattern. 
    Further, we have shown that for each variable $C_k$, there exists exactly one variable $\hat{C}_l$ for which $\bm{J}^{-1}_{kl}\neq 0$. 
    Given that $\bm{J}^{-1}_{k\cdot}$ is zero except for entry $l$, and that this entry index is different for every $k$, it follows that $\bm{J}^{-1}_{kl}$ must be a permuted diagonal matrix:
    \begin{equation}
        \bm{J}^{-1} = \bm{D}\bm{P}
    \end{equation}
    where $\bm{D}$ is a diagonal matrix and $\bm{P}$ is a permutation matrix.
    The diagonal elements of $\bm{D}^{-1}$ are the non-zero values of $\bm{J}^{-1}$, \ie{} where $\bm{J}^{-1}_{kl}\neq 0$.
    Inverting both sides gives us:
    \begin{equation}
        \bm{J} = \bm{P}^T\bm{D}^{-1}
    \end{equation}
    Inverting the diagonal matrix $\bm{D}$ gives us yet another diagonal matrix, just with inverted values.
    Therefore, we have that $\bm{J}_{kl}=\frac{1}{\bm{J}^{-1}_{lk}}$ if $\bm{J}^{-1}_{kl}\neq 0$, and $0$ otherwise.

    \pseudoparagraph{Intermediate proof step based on Lemma~\ref{lemma:proof_align_interventions_option_2}}
    In the proof of Lemma~\ref{lemma:proof_align_interventions_option_2} (\cref{sec:proof_steps_align_intv_cases_multivar_time_var}), we have already shown that $\bm{J}$ must be a permuted diagonal matrix.

    \pseudoparagraph{Joint final step}
    With having $\bm{J}$ identified as a permuted diagonal matrix, we can derive the originally stated component-wise invertible transformation.
    For clarity, we denote the indices at which the Jacobian is non-zero as $f(l)=\arg\max_{(l,k)} |\bm{J}_{lk}|$, \ie{} $f(l)$ returns the index $(l,k)$ for which $\bm{J}_{lk}\neq 0$.
    Using these indices, we can write the determinant of $\bm{J}$ as the product of the individual diagonal elements:
    \begin{equation}
        |\det \bm{J}| = \prod_{l=1}^{K} |\bm{J}_{f(l)}|
    \end{equation}
    Inherently, we can use this to rewrite \cref{eq:proof_equiv_orig_prod} to:
    \begin{align}
        p(C^{t}|C^{t-1},I^t) & = p(\hat{C}^{t}|C^{t-1},\hat{I}^t)\prod_{l=1}^{K} |\bm{J}_{f(l)}| \\
        & = \prod_{l=1}^{K} p(\hat{C}_l^{t}|C^{t-1},\hat{I}^t) |\bm{J}_{f(l)}|
    \end{align}
    Therefore, for a pair of variables $C_k,\hat{C}_l$ with $f(l)=(l,k)$, it follows that:
    \begin{equation}
        p(C_k^{t}|C^{t-1},I_k^t) = p(\hat{C}^{t}_l|C^{t-1},\hat{I}_l^t) |\bm{J}_{lk}|
    \end{equation}
    This shows that for every variable $C_k$, there exist one variable $\hat{C}_l$ with an invertible transformation $C^{t}_k = T_l(\hat{C}^{t}_l)$ which has the Jacobian of $|\bm{J}_{lk}|$.
\end{proof}

\subsubsection{Putting everything together}
\label{sec:proof_steps_final}

Having proven Lemma~\ref{lemma:proof_setup}, \ref{lemma:proof_intervention_cases}, \ref{lemma:proof_align_interventions_option_1}, \ref{lemma:proof_align_interventions_option_2}, and \ref{lemma:proof_invertible_transformation}, we have now all components to prove the original theorem:

\begin{theorem}
    \label{theo:appendix_theorem_no_gaus}
    \label{theo:appendix_theorem_time_var}
    An estimated model $\mathcal{\widehat{M}}=\langle \hat{g},\hat{f},\hat{\omega},\hat{\mathcal{C}} \rangle$ identifies the true causal model $\mathcal{M}=\langle g,f,\omega,\mathcal{C} \rangle$ if:
    \begin{compactenum}
        \item (\textbf{Observations}) $\mathcal{\widehat{M}}$ and $\mathcal{M}$ model the same likelihood:
        $$
            p_{\widehat{\mathcal{M}}}(X^t|X^{t-1},R^t)=p_{\mathcal{M}}(X^t|X^{t-1},R^t);
        $$
        \item (\textbf{Distinct Interaction Patterns}) Each variable $C_i$ in $\mathcal{M}$ has a distinct interaction pattern (Definition \ref{def:theory_distinct_intvs});
    \end{compactenum}
    and one of the following two conditions holds for $\mathcal{M}$:
    \begin{compactenum}
        \item[A.] (\textbf{Dynamics Variability}) Each variable's log-likelihood difference is twice differentiable and not always zero:
        $$
            \forall C^t_i ,\exists C^{t-1} \colon \frac{\partial^2 \intvdiv(C^{t}_i|C^{t-1})}{\partial (C^t_i)^2} \neq 0;
        $$
        \item[B.] (\textbf{Time Variability}) For any $C^{t}\in\mathcal{C}$, there exist $K+1$ different values of $C^{t-1}$ denoted with $c^1,...,c^{K+1}\in\mathcal{C}$, for which the vectors $v_1,...,v_{K}\in\R^{K+1}$ with
        \ifinappendix
        $$
            v_{i} = \begin{bmatrix}
                \frac{\partial \intvdiv\left(C^{t}_i|C^{t-1}=c^1\right)}{\partial C^{t}_i} & 
                \frac{\partial \intvdiv\left(C^{t}_i|C^{t-1}=c^2\right)}{\partial C^{t}_i} & 
                \cdots &
                \frac{\partial \intvdiv\left(C^{t}_i|C^{t-1}=c^{K+1}\right)}{\partial C^{t}_i} \\ 
            \end{bmatrix}^T \in \mathbb{R}^{K+1}
        $$
        \else
        $$
            v_{i} = \begin{bmatrix}
                \frac{\partial \intvdiv\left(C^{t}_i|C^{t-1}=c^1\right)}{\partial C^{t}_i} & 
                \cdots &
                \frac{\partial \intvdiv\left(C^{t}_i|C^{t-1}=c^{K+1}\right)}{\partial C^{t}_i} \\ 
            \end{bmatrix}^T
        $$
        \fi
        are linearly independent.
    \end{compactenum}
\end{theorem}

\begin{proof}
    Based on Lemma~\ref{lemma:proof_setup}, we have shown that there exists an invertible transformation between the latent spaces of $\mathcal{M}$ and $\widehat{\mathcal{M}}$.
    Further, we have shown in Lemma~\ref{lemma:proof_intervention_cases} with Lemma~\ref{lemma:proof_align_interventions_option_1} (for condition (A)) or Lemma~\ref{lemma:proof_align_interventions_option_2} (for condition (B)) that $\widehat{\mathcal{M}}$ must model the same interaction cases and patterns as $\mathcal{M}$.
    Finally, this resulted in the proof of Lemma~\ref{lemma:proof_invertible_transformation}, namely that the invertible transformation $T$ has a Jacobian with the structure of a permuted diagonal matrix.
    This shows that there exist component-wise invertible transformations between the latent spaces of $\mathcal{M}$ and $\widehat{\mathcal{M}}$, effectively \textit{identifying} the causal variables of $\mathcal{M}$.
\end{proof}

\subsection{Extension to Longer Temporal Dependencies}
\label{sec:proof_longer_temporal_deps}

The shown proof demonstrates the identifiability results for causal relations between $C^t$ and its previous time step $C^{t-1}$.
In case the ground truth system contains longer temporal dependencies, e.g. $C^{t-\tau}\to C^{t}$ for any $\tau>1$, we can easily obtain the same identifiability results by extending our conditioning set of the distribution $p(C^t|C^{t-1},R^{t})$ to include $\tau \ge 1$ additional time steps $p(C^t|C^{t-1},C^{t-2},...,C^{t-\tau},R^{t})$. 

The key property of the identifiability proof in \cref{sec:appendix_proofs}, that allows for this simple extension, is that we use $C^{t-1}$ only to ensure that the causal variables in a time step $t$ remain conditionally independent given $R^t$:
\begin{equation}
    C^t_i \perp\hspace{-2mm}\perp C^t_j \hspace{1mm}|\hspace{1mm} C^{t-1},R^t
\end{equation}
In order to extend this to longer dependencies up to $t-\tau$, we can instead consider:
\begin{equation}
    C^t_i \perp\hspace{-2mm}\perp C^t_j \hspace{1mm}|\hspace{1mm} C^{t-1},...,C^{t-\tau},R^t
\end{equation}
Furthermore, the conditioning set can also be extended with any other observable information, e.g. environment parameters, as long as the conditional independencies hold.
In the proof, this corresponds to replacing $C^{t-1}$ with the set of time steps that may have a causal relation to $C^t$, $\left\{C^{t-\tau}|\tau=1,...,T_D\right\}$, with $T_D$ denoting the maximal temporal length of causal relations.

Our learning algorithm, \OurApproach{}, can be similarly adapted to longer temporal relations. Specifically, we need to condition its prior $p_{\omega}$ and interaction MLP $\text{MLP}^{\hat{I}_i}_{\omega}$ on more time steps, i.e. changing $p_{\omega}(z^t|z^{t-1},R^t)$ and $\text{MLP}^{\hat{I}_i}_{\omega}(R^t,z^{t-1})$ to $p_{\omega}(z^t|z^{t-1},z^{t-2},...,z^{t-\tau},R^t)$ and $\text{MLP}^{\hat{I}_i}_{\omega}(R^t,z^{t-1},...,z^{t-\tau})$.

\subsection{Identifying the Temporal Causal Graph}
\label{sec:proof_causal_graph}

In our setting, we assume that the relations between underlying causal variables are limited to edges in a causal graph that go from a variable $C^{t-1}_i$ at time step $t-1$ to another variable $C^t_j$ in the following time step $t$. As summarized in Figure 2, the edges between the interactions variables $I^t_i$ and the relevant causal variables $C^t_i$ are fixed, as are the edges between the regime $R^t$ and the interaction variables $I^t_i$. 

If the true causal variables were observed, by additionally assuming also the standard causal Markov and faithfulness assumptions (which are not otherwise necessary in our setup), one could easily learn the causal relations between $C^{t-1}_i$ and $C^{t}_j$ by checking which of these causal variables are still dependent when conditioning on $C^{t-1} \setminus C^{t-1}_i$ and $R^{t-1}$. This is a trivial modification of known results for causal discovery on time series \cite{peters2013causal}, proving that the causal graph in this setting is identifiable. 

In our setting, we do not know the true causal variables, but we learn the causal variables up to permutation and component-wise transformations. By applying any appropriate causal discovery algorithm for this time series setting, \eg{} as described by \citet{assaad2022survey}, we can then identify the causal structure, again up to permutation of the nodes. We provide an example of discovering the causal graph between the learned causal variables of \OurApproach{} in \cref{sec:appendix_voronoi_results}.

\subsection{Relation to Previous Identifiability Results}
\label{sec:proof_relation_to_prev_results}

\begin{table}[t!]
    \centering
    \caption{Overview of the setups under which identifiability results have been derived in causal representation learning. As additional observations, we distinguish between counterfactuals, intervention targets, and regime variable. Further, we show which causal relations (over time and/or instantaneous) are supported by the respective identifiability results. \OurApproach{} shares a similar setup with iVAE, LEAP and DMS.}
    \label{tab:appendix_related_work_comparison}
    \begin{tabular}{lcccccc}
        \toprule
        \textbf{Method} & \multicolumn{3}{c}{\textbf{Observations / Inputs}} & & \multicolumn{2}{c}{\textbf{Causal Relations}}\\
        & \textit{\small Counterfactuals} & \textit{\small Intervention targets} & \textit{\small Regime} & & \textit{\small Temporal} & \textit{\small Instantaneous} \\
        \midrule
        \rowcolor{white!85!gray} \citet{locatello2020weakly} & \cmark & \xmark & \xmark & & \xmark & \xmark \\
        LCM \cite{brehmer2022weakly} & \cmark & \xmark & \xmark & & \xmark & \cmark \\
        \rowcolor{white!85!gray} \citet{ahuja2022weakly} & \cmark & \xmark & (\cmark) & & \xmark & \xmark \\
        CITRIS \cite{lippe2022citris} & \xmark & \cmark & \xmark & & \cmark & \xmark \\
        \rowcolor{white!85!gray} iCITRIS \cite{lippe2022icitris} & \xmark & \cmark & \xmark & & \cmark & \cmark \\
        iVAE \cite{khemakhem2020variational} & \xmark & \xmark & \cmark & & (\cmark) & \xmark \\ 
        \rowcolor{white!85!gray} LEAP \cite{yao2021learning} & \xmark & \xmark & \cmark & & \cmark & \xmark \\
        DMS \cite{lachapelle2022partial} & \xmark & \xmark & \cmark & & \cmark & \xmark \\
        \rowcolor{white!85!gray} BISCUIT (ours) & \xmark & \xmark & \cmark & & \cmark & \xmark \\
        \bottomrule
    \end{tabular}
\end{table}

We compare our identifiability results to various previous works in causal representation learning in terms of the different inputs/observations the methods require, and the causal relations they support in \cref{tab:appendix_related_work_comparison}. The related works which have the most similar setups to ours, \ie{} using a regime variable and focuses on temporal causal relations, are iVAE \cite{khemakhem2020variational}, LEAP \cite{yao2021learning} and DMS \cite{lachapelle2021disentanglement}. In summary, iVAE and LEAP require a stronger form of both our dynamics and time variability assumptions, which excludes common models like additive Gaussian noise models. DMS requires that no two causal variables share the same parents, limiting the allowed temporal graph structures. We give a more detailed discussion below, which we will add in short to the main paper (Section 3 and 5) and in its full form in the appendix of our paper.

In iVAE \cite{khemakhem2020variational}, the regime variable $u$ is assumed to contain any additional information that makes the causal variables conditionally independent. In our case, $u$ would include both $R^t$ and the previous time step $C^{t-1}$. Under this setup, the iVAE theorem states that an additive Gaussian noise model can only be identified up to a linear transformation, e.g. $\hat{C}^t=AC^t+c$ with $A\in\mathbb{R}^{K\times K}, c\in\mathbb{R}$. In comparison, the identifiability class in BISCUIT for the additive Gaussian noise model is much stronger, since the causal variables are identified up to component-wise invertible transformations. In order to gain a similar identifiability class as BISCUIT, iVAE requires the distributions $p(C^t_i|u)$ to have a sufficient statistic that is either not monotonic, or of size greater than 2 (similar to our dynamics variability assumption). Additionally, iVAE requires $2K+1$ different regimes with linearly independent effects on the causal variables, which is similar to the linear independence in our time variability assumption.

The identifiability theorem of LEAP \cite{yao2021learning} has similar differences to ours as iVAE.
In short, it requires a stronger form of our dynamics and time variability assumption. Specifically, the theorem requires that for all regimes $R^t$ and time steps $C^t$, there exist $2K+1$ values of $C^{t-1}$ for which the first and second-order derivative of the log-likelihood differences (in our notation $\partial \Delta (C^t_i|C^{t-1}) / \partial C^t_i$ and $\partial \Delta (C^t_i|C^{t-1})^2 / \partial^2 C^t_i$) are linearly independent. Thus, it requires our time variability assumption for both the first and second derivative, as well as $2K+1$ points instead of $K+1$ as in BISCUIT. This excludes common models such as additive Gaussian noise models.

Finally, the Disentanglement via Mechanism Sparsity method (DMS) \cite{lachapelle2021disentanglement} is based on the same concepts as iVAE. In order to allow for additive Gaussian noise models, the following assumptions need to be taken: (1) each causal variable has a unique set of parents, (2) the distribution functions for each causal variable need to vary sufficiently over time, and (3) there exist $K+1$ values of $C^{t-1}$ that change the distributions for each causal variable in a linearly independent manner (similar to our time variability assumption). The assumption of unique parents may be violated in settings where causal variables strongly interact, especially when we have context-dependent interactions (e.g. the egg being cooked by the stove, only if it is in the pan). In comparison, BISCUIT can be applied to any graph structure and does not require additional variability assumptions, making it overall wider applicable.

We note that the stronger identifiability results of BISCUIT are only possible by taking the assumption of binary interactions. In situations where the interactions between the regime variable and the causal variables cannot be described by binary variables, iVAE, LEAP or DMS may still be applicable.

\subsection{Detailed Example for Additive Gaussian Noise}
\label{sec:proof_additive_gaussian_noise}

In this section, we discuss the additive Gaussian Noise example of \cref{sec:theory_intuition} in detail, including the specific mechanisms used in \cref{fig:main_gaussian_example}.
Consider an additive Gaussian noise model with two variables $C_1,C_2$ and their dynamics function $C^t_i=\mu_i(C^{t-1}, I^t_i) + \epsilon_i, \epsilon_i \sim \mathcal{N}(0,\sigma^2)$. For simplicity, in this example we assume the regime variable can be represented as $R^t=[I^t_1, I^t_2]$ and the mean function $\mu_i$ to be of the following form:
\begin{equation}
    \mu_i(C^{t-1}, I^t_i)=\begin{cases} 1 & \text{if } I^t_i=1 \\ 0 & \text{otherwise}\end{cases}
\end{equation}
A common difficulty in additive Gaussian noise models is to identify the true causal variables, $C_1,C_2$, in contrast to possible rotated representations, \ie{} $\hat{C}_1=\cos(\theta)C_1 - \sin(\theta) C_2, \hat{C}_1=\sin(\theta)C_1 + \cos(\theta) C_2$ for an angle $\theta\in[0,2\pi)$.
Thus, for simplicity, we limit the possible representation space here to rotations of the form $\hat{C}_1,\hat{C}_2$.
We aim to show that if we enforce the dependency of the regime variable $R^t$ with each causal variable ($\hat{C}_1,\hat{C}_2$) to be expressed by a binary interaction variable ($\hat{I}_1,\hat{I}_2$), $\hat{C}_1$ and $\hat{C}_2$ must identify the true causal variables up to permutation and element-wise invertible transformations.

Firstly, in the general case, we can write the mean functions for the rotated representation $\hat{C}_1,\hat{C}_2$ as:
\begin{equation}
    \hat{\mu}_1(\hat{C}^{t-1}, R^t)=\begin{cases} 
    \cos(\theta) - \sin(\theta) & \text{if } R^t=[1,1] \\
    \cos(\theta) & \text{if } R^t=[1,0] \\
    -\sin(\theta) & \text{if } R^t=[0,1] \\
    0 & \text{if } R^t=[0,0]
    \end{cases}
\end{equation}
\begin{equation}
    \hat{\mu}_2(\hat{C}^{t-1}, R^t)=\begin{cases} 
    \sin(\theta) + \cos(\theta) & \text{if } R^t=[1,1] \\ 
    \sin(\theta) & \text{if } R^t=[1,0] \\
    \cos(\theta) & \text{if } R^t=[0,1] \\
    0 & \text{if } R^t=[0,0]
    \end{cases}
\end{equation}
Since the Gaussian distribution is rotational invariant, the fact that we can determine the mean from $C^{t-1}, R^t$ for the new representation implies that $C_1,C_2$ and $\hat{C}_1,\hat{C}_2$ model the same data likelihood, i.e. $\prod_{i=1}^2 p_i(C^{t}_i|C^{t-1},R^{t}) = \prod_{i=1}^2 \hat{p}_i(\hat{C}^{t}_i|\hat{C}^{t-1},R^{t})$.
Therefore, without constraints on how $R^t$ is used in the distributions for the individual causal variables, we cannot distinguish the true causal variables from an arbitrarily rotated representation simply from the data likelihood under different interactions.

This changes when making use of binary interaction variables $\hat{I}_1,\hat{I}_2$. In this setting, both mean functions need to be reduced to the form:
\begin{equation}
    \hat{\mu}_i(\hat{C}^{t-1}, \hat{I}^t_i)=\begin{cases} 
    a(\hat{C}^{t-1}) & \text{if } \hat{I}^t_i=1 \\\
    b(\hat{C}^{t-1}) & \text{otherwise}
    \end{cases}
\end{equation}
where $a,b$ can be arbitrary functions.
In other words, we need to reduce the mean functions for the general setting from 4 cases to 2 cases per causal variables.
Hereby, it becomes clear that we can only do so by setting either $\sin(\theta)$ or $\cos(\theta)$ to zero, which is at $\theta\in\{0,\pi/2,\pi,3\pi/2\}$.
Any of these rotations are a multiple of 90 degrees, which means that $\hat{C}_1,\hat{C}_2$ are identical to $C_1,C_2$ up to permutation and/or sign-flips.
Therefore, by enforcing the effect of $R^t$ to be described by binary interaction variables, we have successfully identified the causal variables according to Definition 3.1.

\newpage

\section{Experimental Details}
\label{sec:appendix_experimental_details}

In this section, we provide details on the experimental setup, including datasets and hyperparameters. 
\cref{sec:appendix_voronoi}, \ref{sec:appendix_causalworld}, and \ref{sec:appendix_ithor} discuss the Voronoi, CausalWorld, and iTHOR experiments respectively.

\subsection{Voronoi}
\label{sec:appendix_voronoi}

\begin{figure}[t!]
	\centering
        \footnotesize
	\begin{subfigure}[b]{0.35\columnwidth}
		\centering
		\begin{tabular}{ccc}
			\includegraphics[width=0.25\textwidth]{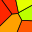} & 
			\includegraphics[width=0.25\textwidth]{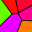} & 
			\includegraphics[width=0.25\textwidth]{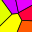} \\
			$t=0$ & $t=1$ & $t=2$ \\
		\end{tabular}
		\caption{Example sequence}
	\end{subfigure}
	\hfill
	\begin{subfigure}[b]{0.25\columnwidth}
		\centering
		\includegraphics[width=\textwidth]{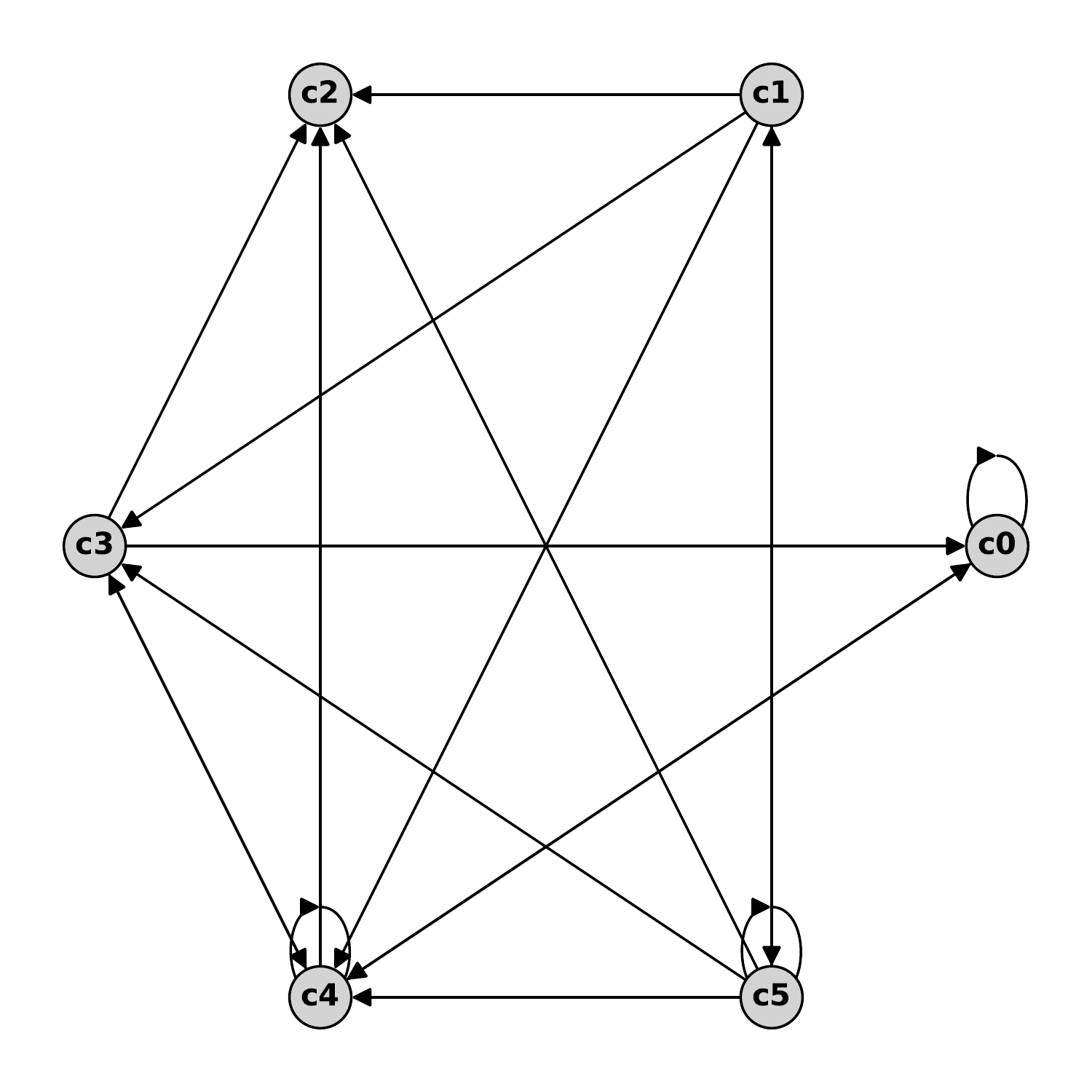}
		\caption{Example temporal causal graph}
	\end{subfigure}
	\hfill
	\begin{subfigure}[b]{0.3\columnwidth}
		\centering
		\begin{tabular}{cc}
			\includegraphics[width=0.3\textwidth]{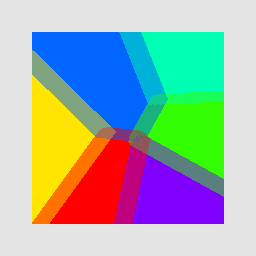} & 
			\includegraphics[width=0.3\textwidth]{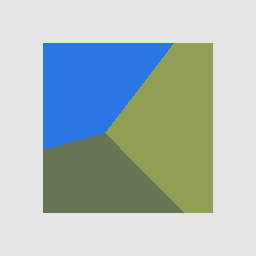}  \\
			Robotic Arm & Minimal Interactions \\
		\end{tabular}
		\caption{Example interaction maps}
	\end{subfigure}
	\caption{(a) Example sequence in the Voronoi dataset \cite{lippe2022icitris} for six variables. The six colors in the images represent an entangled view of the true causal variables. (b) An example graph sampled in the Voronoi dataset for six variables. Each edge $C_i\to C_j$ represents a causal relation from $C_i^{t-1}$ to $C_j^t$. (c) Example interaction maps. The two spatial dimensions represents the two dimensions of $R^t$, with different colors representing different interventional experiments. Gray represents the observational regime. For more examples and predictions, see \cref{fig:appendix_voronoi_interaction_variables}.}
	\label{fig:appendix_voronoi}
\end{figure}

\subsubsection{Dataset Setup}
\label{sec:appendix_voronoi_dataset_setup}

The Voronoi dataset \cite{lippe2022icitris} represents a synthetic benchmark for testing the model on various distributions, graph structures and variable sizes.
We adapt the original setup of \citet{lippe2022icitris} by removing instantaneous effects and restrict the distributions to additive Gaussian noise models:
\begin{equation}
	C^t_i = \text{MLP}_i\left(C^{t-1} \odot M_i\right) + \epsilon, \hspace{5mm}\epsilon\sim\mathcal{N}(0,0.4)
\end{equation}
where $M_i\in\{0,1\}^K$ is a binary mask according to the sampled causal graph (\ie{} $M_{ij}=1$ if $C^{t-1}_j\to C^{t}_i$, and $M_{ij}=0$ otherwise). 
The network $\text{MLP}_i$ is a randomly initialized 3-layer MLP with BatchNorm layers in between.
Under interventions, we set the output of the MLP to zero, effectively performing a perfect intervention while keeping the same noise distribution.
For the graph structures, we sample each edge independently with a chance of $0.4$, \ie{} $M_{ij}\sim\text{Bern}(0.4)$, while ensuring each variable having at least one parent. 
For a graph with six variables, this gives $2.4$ parents per variable in expectation, and $3.6$ for nine variables. 
The \robotstate{} is a two-dimensional continuous variable in the range $R^t\in[-1.5,1.5]^2$ and is sampled uniformly at each time step.
If any dimension has an absolute value greater than 1, the data point is considered observational, \ie{} no interventions.
For all other values in the \textit{Robotic Arm} setup, we assign each voronoi cluster to one causal variable randomly.
If a circle with radius $1/16$ around $R^t$ touches a voronoi cluster, we perform an intervention on this respective causal variable.
On boundaries, this results in performing interventions on multiple variables simultaneously.
In the \textit{Minimal Interactions} setup, we use the same strategy but limit the number of voronoi clusters to $\lfloor \log_2 K\rfloor +2$ as in \citet{lippe2022intervention}, which is four for six variables and five for nine variables.
Further, we do not allow for any overlap.
Numerating the clusters by $c=1,...,\lfloor \log_2 K\rfloor +2$, we perform an intervention on variable $C_i$ if $\lfloor\frac{(i+1)}{2^{c-1}}\rfloor\mod 2 = 0$. 
This gives each causal variable a unique pattern in its interaction variable.
An example setup for both cases is shown in \cref{fig:appendix_voronoi}.
For each dataset, we sample 150k training frames in a single sequence, and 25k samples for the held-out test set.

\subsubsection{Hyperparameters and Implementation Details}
\label{sec:appendix_voronoi_hyperparams}

The hyperparameters shared across all methods are shown in \cref{sec:appendix_voronoi_hyperparams}.
The CNN used for the encoder and decoder is the same as in \citet{lippe2022icitris}.
The number of latents is twice as the number of causal variables, which represents a rough overestimation of the true number of causal variables in the environment.
Further, the learning rate was finetuned separately for each method in the range [1e-4, 1e-3].
However, none of the tested methods showed to be sensitive to these hyperparameters, and 4e-4 showed to work well across all methods.
In the paragraph below, we discuss further implementation details specific to the individual methods.

\paragraph{\OurApproach{}}
In the prior, \MLPzi{} is implemented by a 2-layer MLP with hidden dimensionality 32 and SiLU \cite{ramachandran2017searching} activations. 
\MLPI{} shares the same structure. 
The output activation function of \MLPI{} is implemented as $f(x)=\tanh(x \cdot \tau)$ where $\tau$ represents the inverse of the temperature and is scaled linearly in the range $[1,5]$ from start to end of the training.
With $\tau=\infty$, the activation function becomes a step function: $f(x)=\tanh(x \cdot \infty)=1 \text{ if }x > 0$, $-1$ otherwise. 
As regularizer on the logits, we use an L2 regularizer for values that are above -1: $\ell_2=\max(x+1, 0)^2$.
This regularizer gives the interaction variables a bias towards being 0 / -1 for the default, observational case.
We add this regularizer with a weight of 5e-4 to the negative log-likelihood loss of \OurApproach{}.

\paragraph{iVAE}
As conditioning variable $u$, we use the concatenation of the latents of the previous time step, $z^{t-1}$, and the current \robotstate{} $R^t$.
To account for the parameter difference to \OurApproach{} and the other baselines, we use a 2-layer MLP with hidden size 256. 
The prior outputs a Gaussian per latent with learnable mean and standard deviation.

\paragraph{LEAP}
We base our implementation of LEAP \cite{yao2021learning} on the publicly released code\footnote{\url{https://github.com/weirayao/leap}}.
To allow for LEAP to work both on single images and sequences, we do not use an RNN structure in the encoder as this showed to give poor performance on this dataset.
We use the same MLP architecture as in \OurApproach{} to model the prior network, and use affine conditional normalizing flows for the mapping of the noise to the prior distributions.
The discriminator is implemented with a 2-layer MLP with hidden dimension 64.
We finetuned the hyperparameters for sparsity and discriminator weight on the NLL weight, for which we found $0.01$ and $0.1$ respectively.

\paragraph{DMS}
We base our implementation of DMS \cite{lachapelle2021disentanglement} on the publicly released code\footnote{\url{https://github.com/slachapelle/disentanglement_via_mechanism_sparsity}}.
In DMS, we use $R^t$ as two action variables that are concatenated to the inputs.
We differ from the original implementation by only training the model on the conditional distribution, \ie{} $p(X^t|X^{t-1},R^{t})$, and not include the loss for the first element, $p(X^0|R^{0})$.
The reason for this is that we observed significantly worse performance on this dataset when including this prior loss, and to be fair with \OurApproach{}, we align the implementation to focus on the conditional distribution too.
As a hyperparameter, we finetuned the sparsity weight in the range $[0.001,0.1]$ with a final value of $0.02$, which lead to a learned graph density of $\sim 0.2$.

\begin{table}[t!]
	\centering
	\caption{Shared hyperparameters across all methods. The learning rate was separately finetuned for each method based on the NLL loss. For individual hyperparameters, see \cref{sec:appendix_voronoi_hyperparams}.}
	\begin{tabular}{l|cccc}
		\toprule
		\textbf{Hyperparameter} & \textbf{iVAE} & \textbf{LEAP} & \textbf{DMS} & \textbf{\OurApproach{}} \\
		\midrule
		Encoder architecture & \multicolumn{4}{c}{5-layer CNN} \\
		Decoder architecture & \multicolumn{4}{c}{5-layer CNN} \\
		Channel size & \multicolumn{4}{c}{32} \\
		Number of latents & \multicolumn{4}{c}{$2\cdot K$} \\
		Learning rate & 4e-4 & 4e-4 & 4e-4 & 4e-4 \\
		Optimizer & \multicolumn{4}{c}{Adam \cite{kingma2015adam}} \\
		Batch size & \multicolumn{4}{c}{256} \\
		Number of epochs & \multicolumn{4}{c}{100} \\
		\bottomrule
	\end{tabular}
\end{table}

\subsubsection{Results}
\label{sec:appendix_voronoi_results}

\paragraph{Result Table} 
As supplement to \cref{fig:experiments_voronoi_bar_plot} of the main plot, we provide the $R^2$ scores with standard deviations in \cref{tab:appendix_voronoi_results}.
Besides the $R^2$ score, we also report the Spearman correlation on the same latent variables to the ground truth causal variables.
We created five datasets with different graphs and mechanisms for each setup.
Each model was trained on these five datasets with two different seeds each. 
This gives us 10 results per model and data setup.

\begin{table}[t!]
    \centering
    \caption{Result table for the Voronoi results visualized in \cref{fig:experiments_voronoi_bar_plot}. All results are reported over 10 experiments, with standard deviation listed next to the mean.}
    \label{tab:appendix_voronoi_results}
    \resizebox{0.8\textwidth}{!}{
    \begin{tabular}{ccccccc}
        \toprule
        \textbf{\# variables} & \textbf{Interactions} & \textbf{Method} & \textbf{$\bm{R^2}$-diag} & \textbf{$\bm{R^2}$-sep} & \textbf{Spearman-diag} & \textbf{Spearman-sep} \\
        \midrule
        \multirow{4}{*}{6 variables} & \multirow{4}{*}{Robotic arm} & \OurApproach{} & $0.99 \pm 0.00$ & $0.02 \pm 0.03$ & $0.99 \pm 0.00$ & $0.07 \pm 0.04$ \\
        & & DMSVAE & $0.63 \pm 0.09$ & $0.31 \pm 0.05$ & $0.77 \pm 0.07$ & $0.53 \pm 0.05$ \\
        & & LEAP & $0.62 \pm 0.08$ & $0.36 \pm 0.06$ & $0.76 \pm 0.07$ & $0.57 \pm 0.05$ \\
        & & iVAE & $0.48 \pm 0.06$ & $0.23 \pm 0.04$ & $0.64 \pm 0.06$ & $0.45 \pm 0.04$ \\
        \midrule
        \multirow{4}{*}{9 variables} & \multirow{4}{*}{Robotic arm} & \OurApproach{} & $0.98 \pm 0.00$ & $0.02 \pm 0.02$ & $0.99 \pm 0.00$ & $0.08 \pm 0.04$ \\
        & & DMSVAE & $0.67 \pm 0.07$ & $0.31 \pm 0.03$ & $0.79 \pm 0.05$ & $0.53 \pm 0.03$ \\
        & & LEAP & $0.65 \pm 0.06$ & $0.35 \pm 0.06$ & $0.78 \pm 0.06$ & $0.57 \pm 0.05$ \\
        & & iVAE & $0.38 \pm 0.05$ & $0.23 \pm 0.04$ & $0.54 \pm 0.06$ & $0.46 \pm 0.03$ \\
        \midrule
        \multirow{4}{*}{6 variables} & \multirow{4}{*}{Minimal inter.} & \OurApproach{} & $0.99 \pm 0.00$ & $0.02 \pm 0.02$ & $0.99 \pm 0.00$ & $0.09 \pm 0.03$ \\
        & & DMSVAE & $0.62 \pm 0.14$ & $0.26 \pm 0.05$ & $0.75 \pm 0.11$ & $0.48 \pm 0.05$ \\
        & & LEAP & $0.59 \pm 0.07$ & $0.31 \pm 0.04$ & $0.74 \pm 0.05$ & $0.53 \pm 0.04$ \\
        & & iVAE & $0.47 \pm 0.08$ & $0.25 \pm 0.04$ & $0.63 \pm 0.07$ & $0.46 \pm 0.04$ \\
        \midrule
        \multirow{4}{*}{9 variables} & \multirow{4}{*}{Minimal inter.} & \OurApproach{} & $0.98 \pm 0.00$ & $0.00 \pm 0.01$ & $0.99 \pm 0.00$ & $0.05 \pm 0.02$ \\
        & & DMSVAE & $0.62 \pm 0.07$ & $0.26 \pm 0.03$ & $0.77 \pm 0.06$ & $0.48 \pm 0.03$ \\
        & & LEAP & $0.60 \pm 0.07$ & $0.30 \pm 0.04$ & $0.74 \pm 0.07$ & $0.52 \pm 0.04$ \\
        & & iVAE & $0.37 \pm 0.05$ & $0.22 \pm 0.03$ & $0.52 \pm 0.07$ & $0.45 \pm 0.03$ \\
        \bottomrule
    \end{tabular}
    }
\end{table}

\begin{figure}[t!]
    \centering
    \footnotesize
    \begin{subfigure}[b]{0.4\columnwidth}
        \begin{tabular}{cc}
            \includegraphics[width=0.4\textwidth]{figures/experiments/voronoi/interaction_maps/6_vars_robarm_seed42_gt.png} & 
            \includegraphics[width=0.4\textwidth]{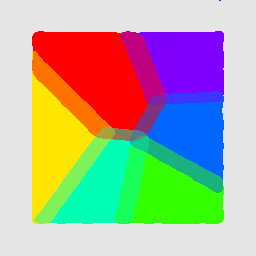} \\
            Ground Truth & Learned Interactions \\
        \end{tabular}
        \caption{6 variables, robotic arm}
    \end{subfigure}
    \hspace{5mm}
    \begin{subfigure}[b]{0.4\columnwidth}
        \begin{tabular}{cc}
            \includegraphics[width=0.4\textwidth]{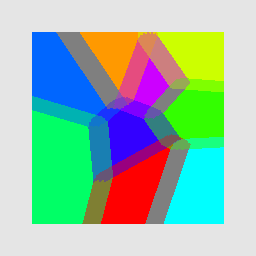} & 
            \includegraphics[width=0.4\textwidth]{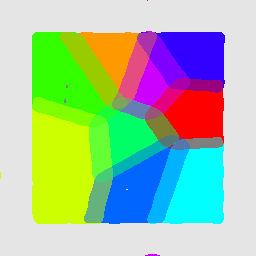} \\
            Ground Truth & Learned Interactions \\
        \end{tabular}
        \caption{9 variables, robotic arm}
    \end{subfigure}
    \begin{subfigure}[b]{0.4\columnwidth}
        \begin{tabular}{cc}
            & \\
            \includegraphics[width=0.4\textwidth]{figures/experiments/voronoi/interaction_maps/6_vars_min_seed42_gt.png} & 
            \includegraphics[width=0.4\textwidth]{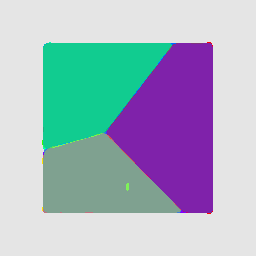} \\
            Ground Truth & Learned Interactions \\
        \end{tabular}
        \caption{6 variables, minimal interactions}
    \end{subfigure}
    \hspace{5mm}
    \begin{subfigure}[b]{0.4\columnwidth}
        \begin{tabular}{cc}
            & \\
            \includegraphics[width=0.4\textwidth]{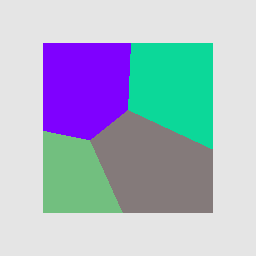} & 
            \includegraphics[width=0.4\textwidth]{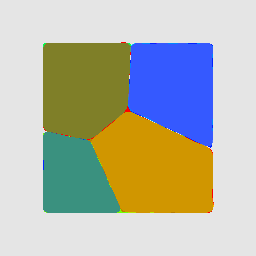} \\
            Ground Truth & Learned Interactions \\
        \end{tabular}
        \caption{9 variables, minimal interactions}
    \end{subfigure}
    \caption{Visualizing the ground truth (left) and learned (right) interaction variables of \OurApproach{} on the Voronoi dataset. The $x$ and $y$ dimension represent the two dimensions of the robotic arm position $R^t\in[-1.5,1.5]^2$, and the different colors correspond to interactions with different causal variables, \ie{} interaction variables being equal to 1 (not to confuse with the actual colors in the observations). Gray represents the observational regime. (a,b) Under the robotic arm interaction setup, the interaction map corresponds to the same Voronoi structure as the observation. \OurApproach{} identifies the same structure well up to minor errors on the boundaries. Note that the ground truth and prediction matches up to permutation of the colors, \ie{} the same permutation as latent to causal variables. (c,d) Under minimal interactions, \OurApproach{} models the same interaction structure as the ground truth. The different colors are due to the several overlaps of interaction variables.}
    \label{fig:appendix_voronoi_interaction_variables}
\end{figure}

\paragraph{Learned Interaction Variables}
To verify that the learned interaction variables are matching the ground truth, we plot them over the different values of $R^t$ in \cref{fig:appendix_voronoi_interaction_variables}.
The x and y dimensions of the images correspond to the two dimensions of $R^t$, and different colors show different interaction variables.
Overall, we find that \OurApproach{} learned the same underlying structure of the interaction variables, but, as expected, with an arbitrary permutation.

\begin{figure}
    \centering
    \begin{subfigure}{0.48\textwidth}
        \includegraphics[width=\textwidth]{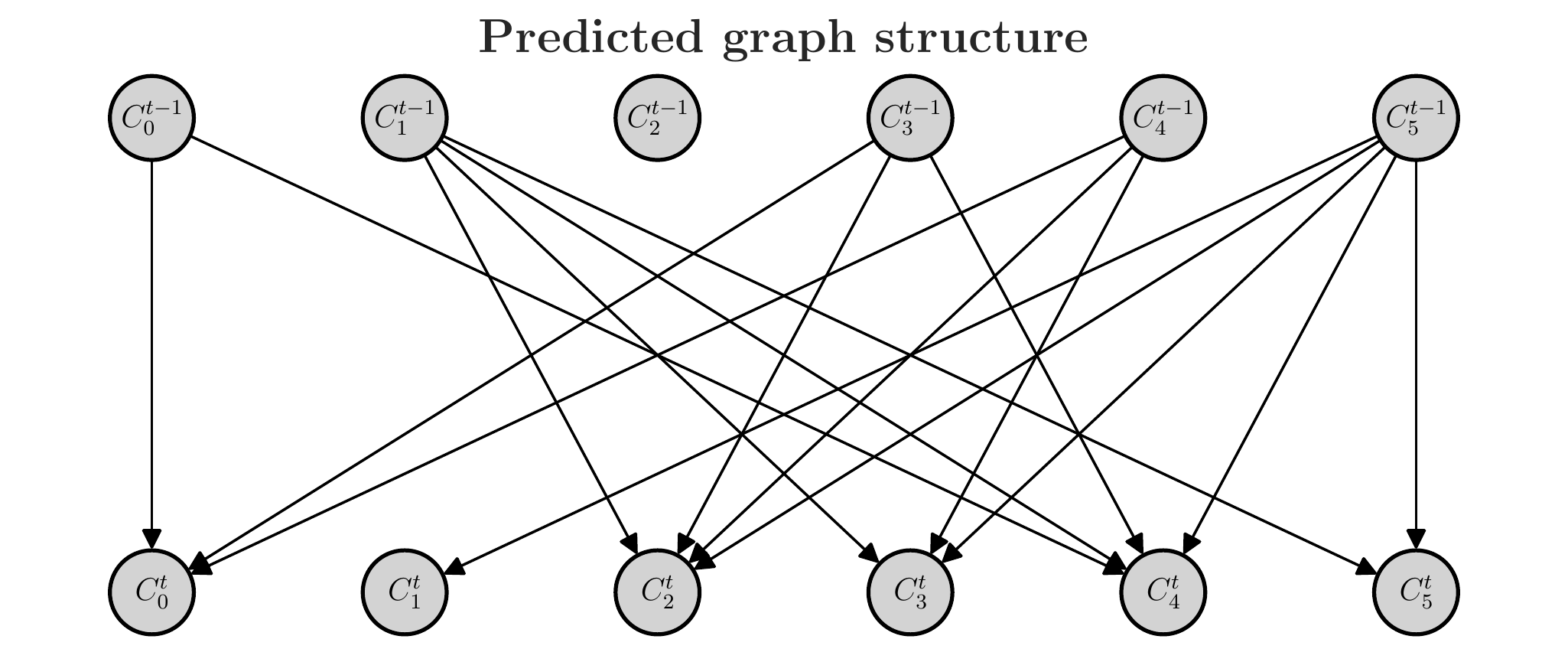}
        \caption{BISCUIT Learned Graph (6 vars)}
    \end{subfigure}
    \begin{subfigure}{0.48\textwidth}
        \includegraphics[width=\textwidth]{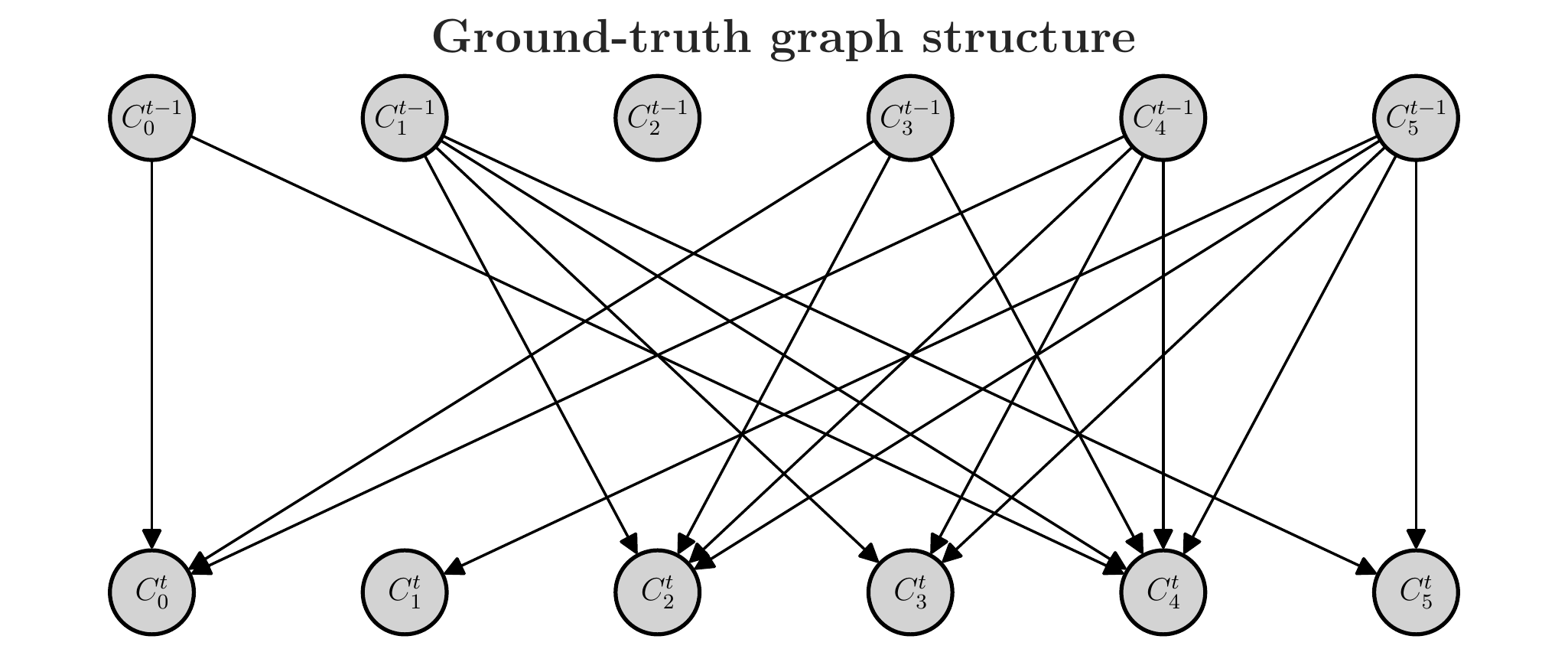}
        \caption{Ground-Truth Graph (6 vars)}
    \end{subfigure}
    \begin{subfigure}{0.48\textwidth}
        \includegraphics[width=\textwidth]{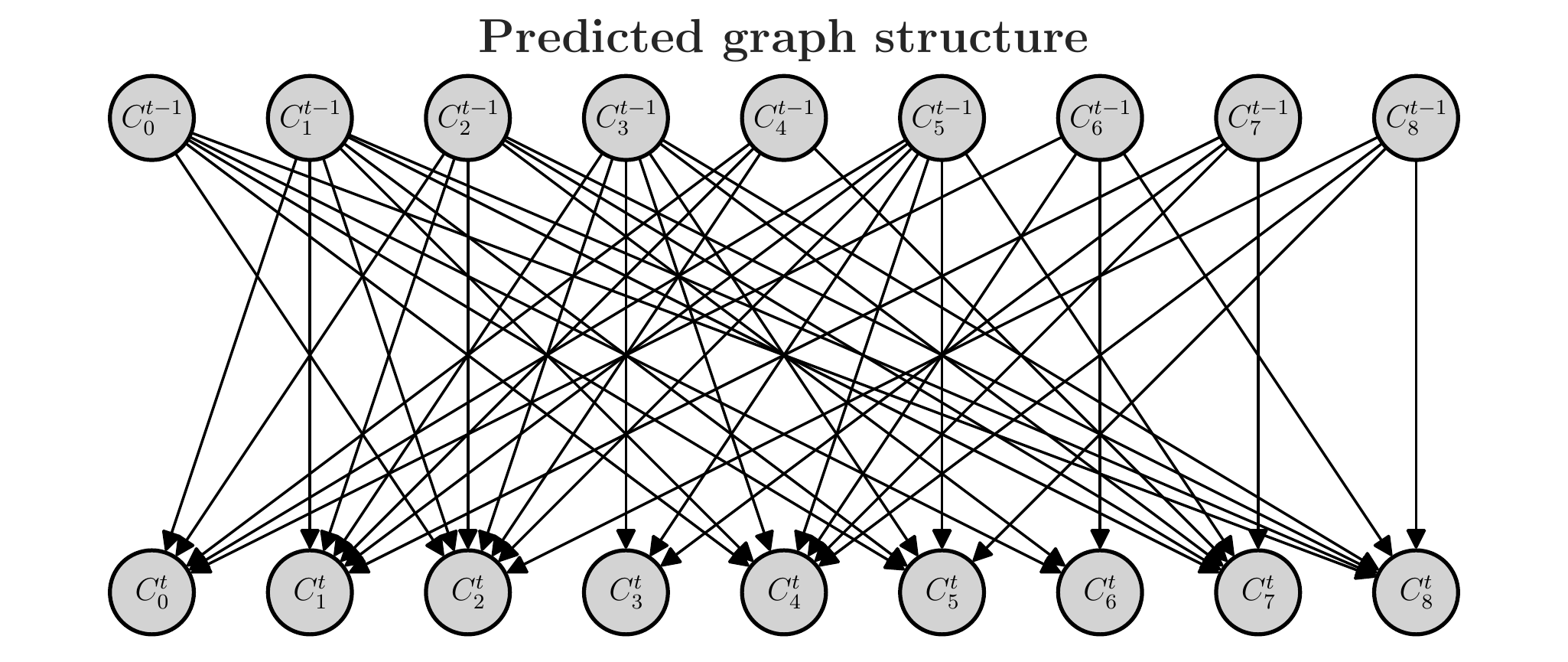}
        \caption{BISCUIT Learned Graph (9 vars)}
    \end{subfigure}
    \begin{subfigure}{0.48\textwidth}
        \includegraphics[width=\textwidth]{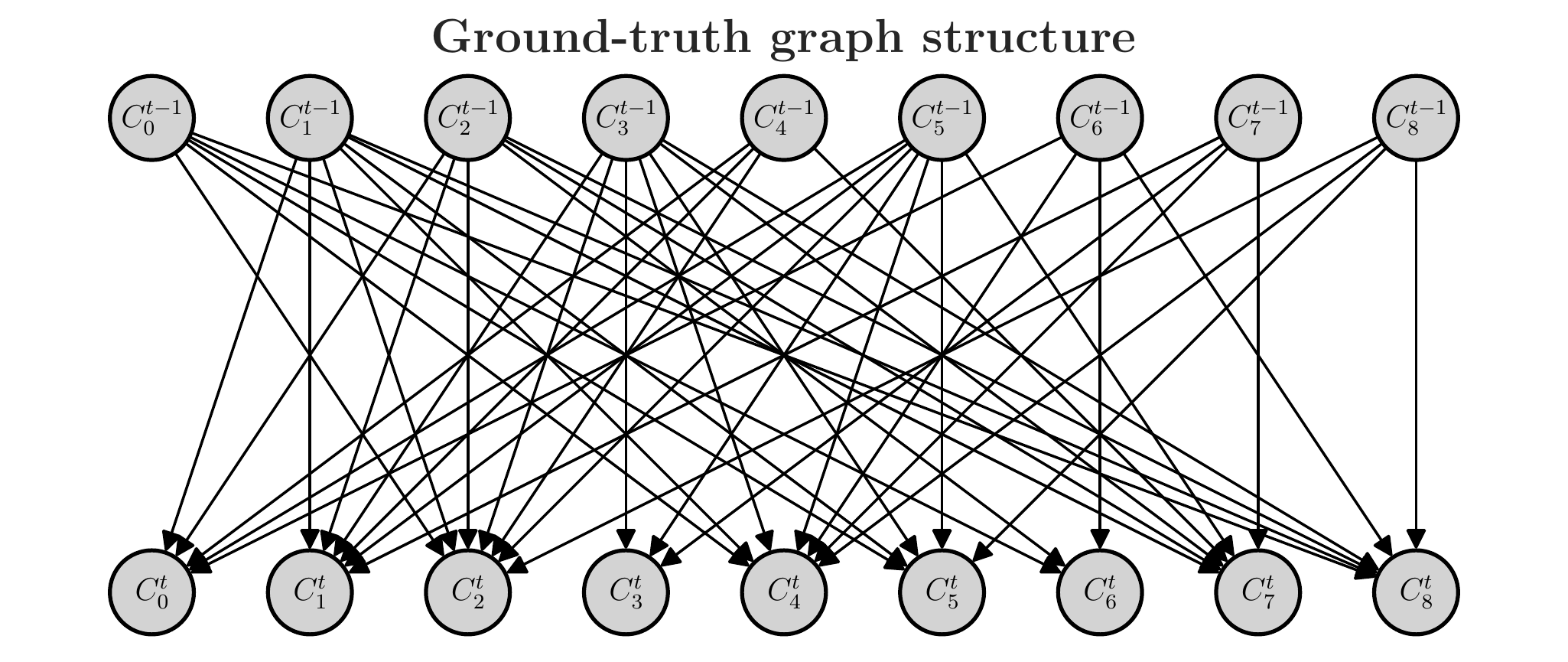}
        \caption{Ground-Truth Graph (9 vars)}
    \end{subfigure}
    \caption{Discovering the causal graph among learned causal variables. Subfigure (a) and (c) show the causal structure that has been discovered between the learned causal variables of BISCUIT on a Voronoi dataset of 6 and 9 variables, respectively. Subfigures (b) and (d) show the corresponding ground truth causal structures. BISCUIT identifies the graph up to an SHD error of 1 for the 6 variables and 0 for 9 variables.}
    \label{fig:appendix_voronoi_graph_discovery}
\end{figure}

\paragraph{Learned Causal Structure} We provide results for estimating the causal graph between the learned causal variables of BISCUIT in the Voronoi dataset. For estimating $p(z^{t}\_i|z^{t-1}, R^t)$, we start from a fully-connected graph from $z^{t-1}$ to $z^t$, and applying a sparsity regularizer to remove edges. This is similar to NOTEARS \cite{zheng2018dags} without the acyclicity regularizer, since the directions of all edges are known. Alternatively, intervention-based causal discovery algorithms like DCDI \cite{brouillard2020differentiable} or ENCO \cite{lippe2022enco} could be used with the learned interaction variables. The results in \cref{fig:appendix_voronoi_graph_discovery} show that the identified causal graph matches the ground truth graph, with an SHD of 0 for the 9 variable graph and an SHD of 1 for the 6 variable graph.

\subsection{CausalWorld}
\label{sec:appendix_causalworld}

\begin{figure}[t!]
    \centering
    \begin{tabular}{ccc}
        \includegraphics[width=0.3\textwidth]{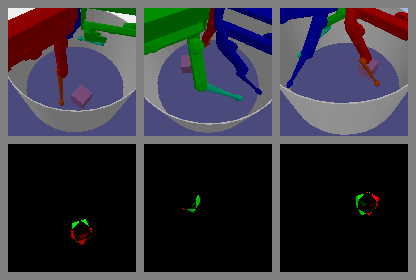} & 
        \includegraphics[width=0.3\textwidth]{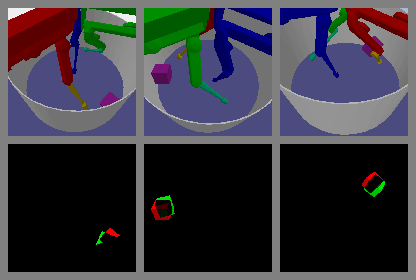} & 
        \includegraphics[width=0.3\textwidth]{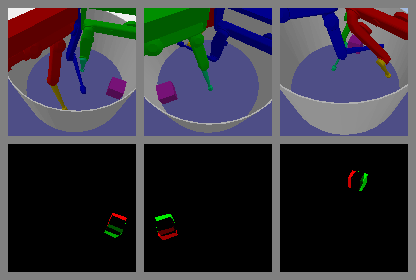} \\
        $t=0$ & $t=1$ & $t=2$ \\
    \end{tabular}
    \caption{Example sequence in the CausalWorld dataset \cite{ahmed2020causalworld}. Each observation has 3 images from different camera perspectives, with each image being an RGB with additional velocity channel of the object. For visualization, the velocity channel is shown with positive values in red and negative in green. The overall observation dimension is $64\times 64\times 12$.}
    \label{fig:appendix_causal_world_sequence}
\end{figure}

\subsubsection{Dataset Setup}
\label{sec:appendix_causalworld_dataset_setup}

We set up the CausalWorld environment \cite{ahmed2020causalworld} to contain a single cube as an object with a tri-finger system to interact with it.
The environment is observed by three cameras positioned around the arena, each returning an RGB image of dimensions $128\times 128\times 3$.
Additionally, to perceive the velocity, a difference of frames on the cube is concatenated to each RGB image.
This gives a combined observation size of $128\times 128\times 12$.
To reduce the computation cost of training all models on this dataset, we bi-linearly downscale the images to a resolution of $64\times 64\times 12$.

The \robotstate{} consists of the three rotation angles of each arm of the tri-finger at the current and previous time step, giving overall $R^t\in\R^{18}$.
This provides both location and velocity information about the tri-finger robotic system.
Furthermore, in this setup, the \robotstate{} across time steps has a causal relation, \ie{} $R^{t-1}\to R^t$, since they share one time step and the tri-fingers have a limited distance they can travel within one time step.

The causal graph consists of seven high-level causal variables: the colors of the three tri-fingers, the friction of the canvas, the floor, and the cube, and finally the state of the cube (cube position, velocity and rotation).
The frictions of the different objects are visualized by the colors of the respective objects. 
For all colors and frictions, we use an additive Gaussian noise model as a ground truth causal model, where under no interactions, we have $C^t_i=0.95\cdot C^{t-1}_i + \epsilon_i, \epsilon_i\sim \mathcal{N}(0,0.15)$.
Under interactions, we set $C^t_i=\sigma^{-1}(u), u\sim U(0,1)$ where $\sigma^{-1}$ is an inverse sigmoid.
The causal model of the cube state is based on the physical interactions between the cube, the robot, and the frictions of the floor, stage, and cube.
An interaction of one of the tri-fingers with the cube causes an intervention on the color of the touched tri-finger and the cube state (position, velocity, and rotation). 
Further, the floor friction is randomly re-sampled if all three tri-fingers touch the floor.
Similarly, the canvas friction changes if all tri-finger rotations of the first arm element are above a certain threshold.
Finally, the robot interacts with the cube friction if the fingers touch in the center of the arena.

We generate 200 sequences of each 1000 frames for training, and 25 sequences for testing.
Examples are shown in \cref{fig:appendix_causal_world_sequence}.

\subsubsection{Hyperparameters and Implementation Details}
\label{sec:appendix_causalworld_hyperparams}

\paragraph{\OurApproach{}}
We apply the autoencoder + normalizing flow setup of \OurApproach{}.
For this, we first train an autoencoder with mean-squared error loss on individual observations to map the $64\times 64\times 12$ input to a $32$ dimensional latent space.
During the autoencoder training, we apply a Gaussian noise of $0.05$ on the latents, and add an L2 regularizer with a weight of 1e-5 to limit the scale of the latent variables.
As architecture, we use a convolutional ResNet \cite{he2016deep} with two convolutional layers with consecutive GroupNorm normalization \cite{wu2018group} per ResNet block. 
After each two ResNet blocks, we reduce the spatial dimensionality using a convolution with stride 2 until we reach a spatial size of $4\times 4$.
At this point, the feature map is flattened, and two linear layers map it to the $32$ latent dimensions.
The decoder is a mirrored version of the encoder, replacing stride convolutions with up-scaling layers using bi-linear interpolation.
We add $R^t$ to the decoder by concatenating it with the latent vector before passing it to the first linear layer of the decoder.
Both encoder and decoder use a channel size of 128.
We train this network with a batch size of 128 and learning rate of 4e-4 with cosine scheduling for 500 epochs.

The normalizing flow follows the architecture used by \citet{lippe2022citris}, namely six autoregressive affine coupling layers \cite{dinh2017density} with 1x1 convolutions and activation normalization \cite{kingma2018glow} in between.
The prior uses the same setup and hyperparameters as for the Voronoi dataset, except using a hidden dimension of 64 instead of 32 in the MLPs.
We use a batch size of 512 and learning rate of 1e-3, and train for 100 epochs.

\paragraph{Baselines}
For all baselines, we use the same convolutional ResNet architecture for the encoder and decoder as in the autoencoder of \OurApproach{}.
To account for the additional parameters introduced by the normalizing flow of \OurApproach{}, we increase the hidden dimensions of the prior networks correspondingly.
However, in general, we found no noticeable gain from increasing the prior dimensions of the baselines further beyond 64 for LEAP and DMS per latent, and 51sarta2 for iVAE.
We performed a small hyperparameter search over the learning rates \{2e-4, 4e-4, 1e-3\} and sparsity regularizers \{0.001, 0.01, 0.1\}.
For LEAP and iVAE, we picked learning 4e-4, and 2e-4 for DMS.
For the sparsity regularizers, very high regularizer lead to early sparsification of the graph, while too small regularizers did not change the graph much.
Thus, we picked 0.01 for both models, although it did not show much of an impact on the final identification result.
We use a batch size of 64 and train for 250 epochs with cosine learning rate scheduling.
Longer trainings did not show any improvements.
 
\subsubsection{Results}
\label{sec:appendix_causalworld_results}

\begin{figure}[t!]
    \centering
    \begin{subfigure}[b]{0.4\columnwidth}
        \centering
        \includegraphics[width=\textwidth]{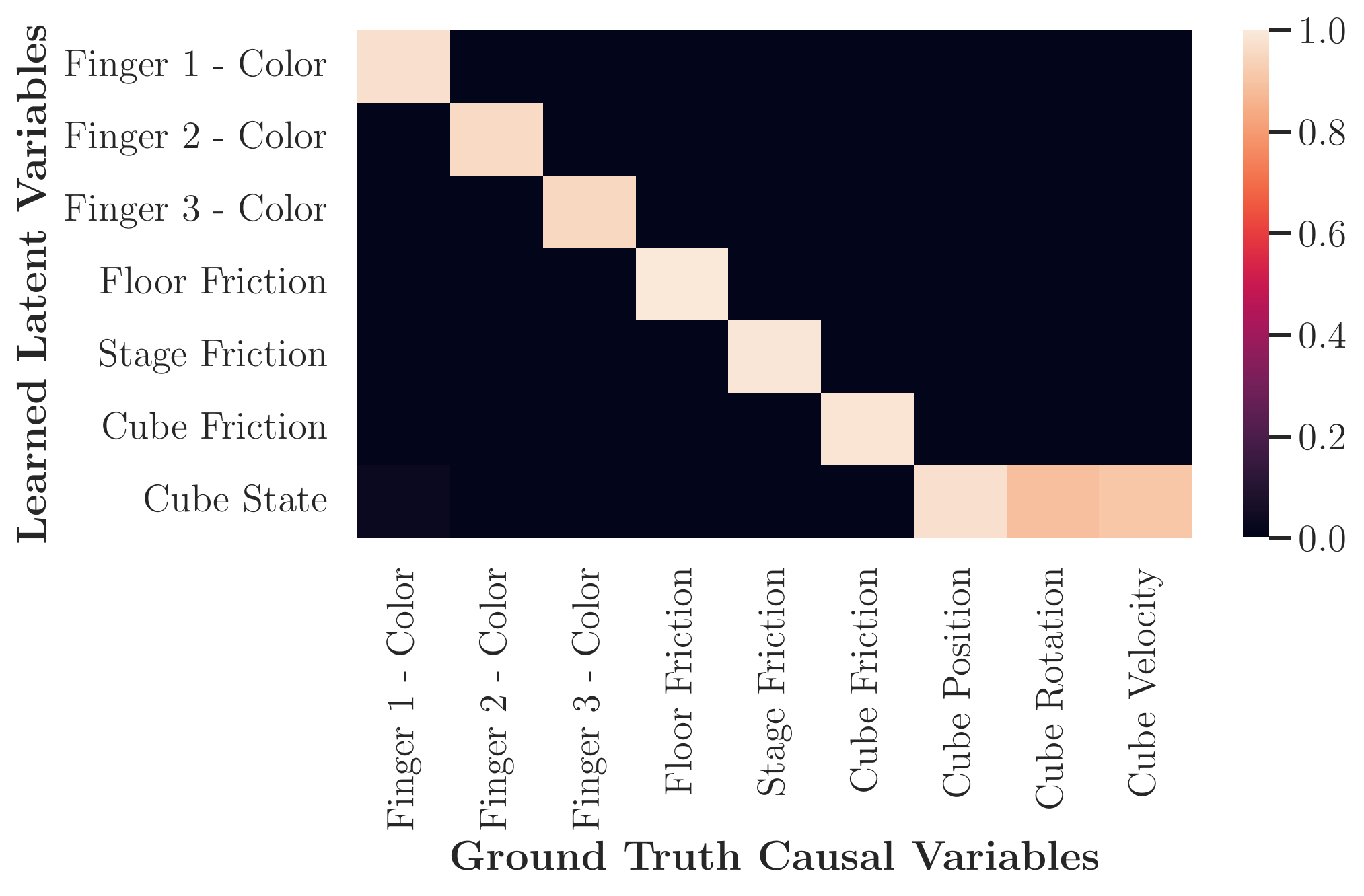}
        \caption{\OurApproach{}}
    \end{subfigure}
    \hspace{5mm}
    \begin{subfigure}[b]{0.4\columnwidth}
        \centering
        \includegraphics[width=\textwidth]{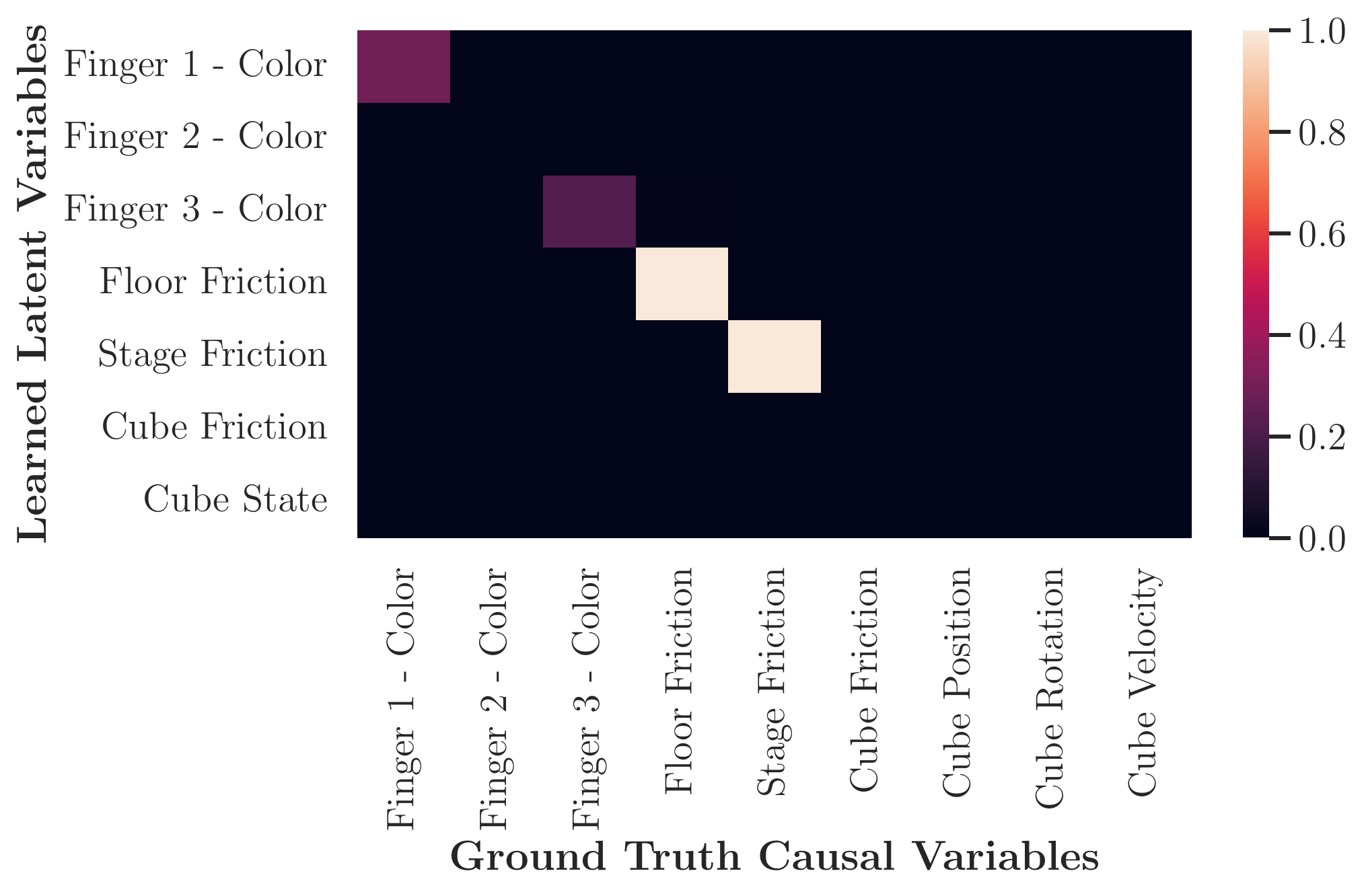}
        \caption{DMS \cite{lachapelle2021disentanglement}}
    \end{subfigure}
    \begin{subfigure}[b]{0.4\columnwidth}
        \vspace{2mm}
        \centering
        \includegraphics[width=\textwidth]{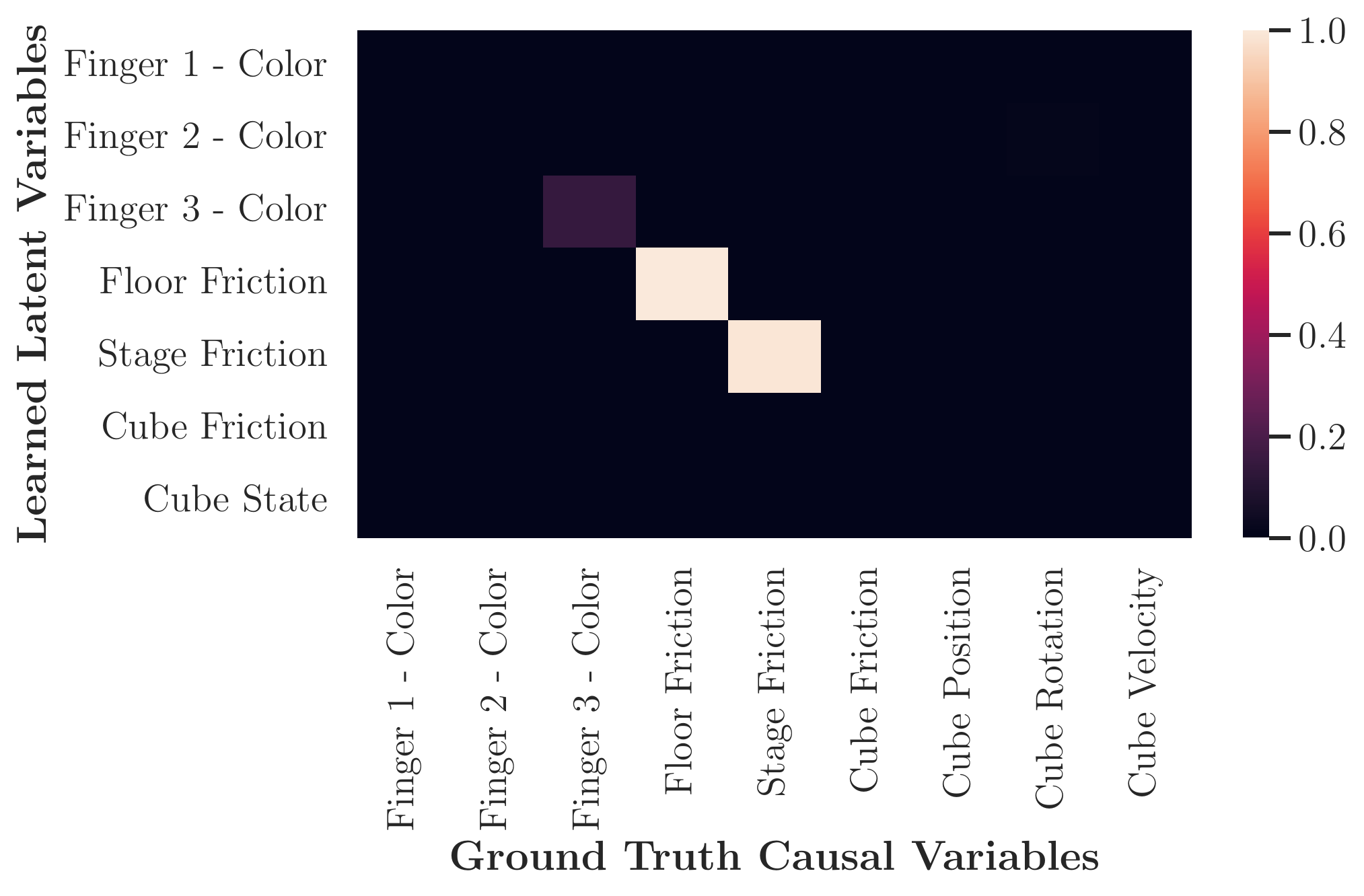}
        \caption{LEAP \cite{yao2021learning}}
    \end{subfigure}
    \hspace{5mm}
    \begin{subfigure}[b]{0.4\columnwidth}
        \vspace{2mm}
        \centering
        \includegraphics[width=\textwidth]{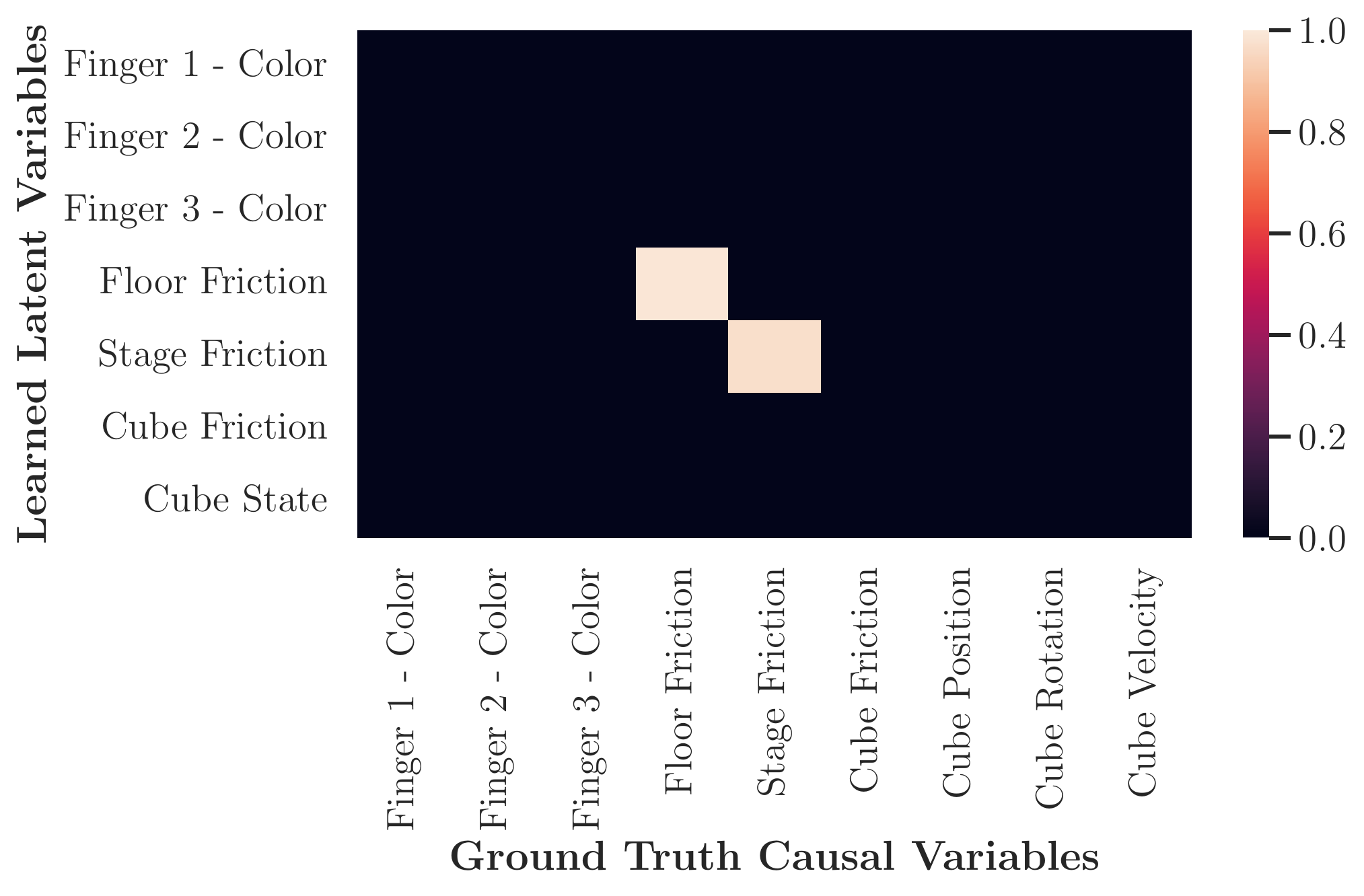}
        \caption{iVAE \cite{khemakhem2020variational}}
    \end{subfigure}
    \caption{$R^2$ matrix for learned models on the CausalWorld environment \cite{ahmed2020causalworld}: (a) \OurApproach{} (ours), (b) DMS \cite{lachapelle2021disentanglement}, (c) LEAP \cite{yao2021learning}, (d) iVAE \cite{khemakhem2020variational}.}
    \label{fig:appendix_causalworld_r2_matrix}
\end{figure}

\begin{table}[t!]
    \centering
    \caption{Result table for the CausalWorld experiments over 3 seeds, with standard deviation listed next to the mean.}
    \label{tab:appendix_causalworld_results}
    \footnotesize
    \begin{tabular}{ccccc}
        \toprule
        \textbf{Method} & \textbf{$\bm{R^2}$-diag} & \textbf{$\bm{R^2}$-sep} & \textbf{Spearman-diag} & \textbf{Spearman-sep} \\
        \midrule
        \OurApproach{} & $0.97 \pm 0.01$ & $0.01 \pm 0.00$ & $0.98 \pm 0.00$ & $0.09 \pm 0.02$ \\
        DMSVAE & $0.32 \pm 0.03$ & $0.00 \pm 0.00$ & $0.44 \pm 0.09$ & $0.02 \pm 0.00$ \\
        LEAP & $0.30 \pm 0.02$ & $0.00 \pm 0.00$ & $0.36 \pm 0.06$ & $0.02 \pm 0.00$ \\
        iVAE & $0.28 \pm 0.00$ & $0.00 \pm 0.00$ & $0.29 \pm 0.00$ & $0.02 \pm 0.00$ \\
        \bottomrule
    \end{tabular}
\end{table}

\paragraph{Correlation Evaluation} We show the results of all methods on the dataset in \cref{tab:appendix_causalworld_results}.
Similar to the results on the Voronoi dataset, we include the standard deviation over multiple seeds.
We performed each experiment only for three seeds to limit the computational cost, and the standard deviation was much lower than the significant differences.
As a second metric, we also report the Spearman correlation, which shows a very similar trend.
Finally, we show example $R^2$ matrices learned by all methods in \cref{fig:appendix_causalworld_r2_matrix}.

\begin{figure}[t!]
    \centering
    \begin{subfigure}{\textwidth}
        \centering
        \includegraphics[width=0.4\textwidth]{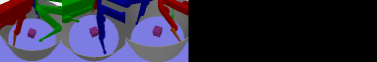}
        \caption{Ground Truth}
    \end{subfigure}
    \begin{subfigure}{0.4\textwidth}
        \centering
        \includegraphics[width=\textwidth]{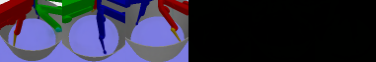}
        \caption{iVAE Reconstruction}
    \end{subfigure}
    \begin{subfigure}{0.4\textwidth}
        \centering
        \includegraphics[width=\textwidth]{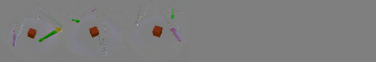}
        \caption{iVAE Reconstruction - Difference to GT}
    \end{subfigure}
    \begin{subfigure}{0.4\textwidth}
        \centering
        \includegraphics[width=\textwidth]{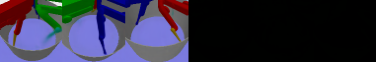}
        \caption{LEAP Reconstruction}
    \end{subfigure}
    \begin{subfigure}{0.4\textwidth}
        \centering
        \includegraphics[width=\textwidth]{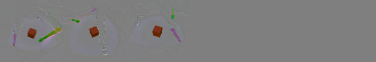}
        \caption{LEAP Reconstruction - Difference to GT}
    \end{subfigure}
    \begin{subfigure}{0.4\textwidth}
        \centering
        \includegraphics[width=\textwidth]{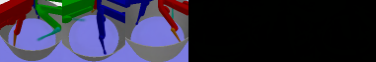}
        \caption{DMS Reconstruction}
    \end{subfigure}
    \begin{subfigure}{0.4\textwidth}
        \centering
        \includegraphics[width=\textwidth]{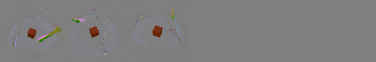}
        \caption{DMS Reconstruction - Difference to GT}
    \end{subfigure}
    
    \caption{Reconstructions on the CausalWorld dataset by the baselines (b) iVAE, (d) LEAP and (f) DMS for an (a) ground truth example. The differences between the prediction of iVAE, LEAP and DMS to the ground truth is shown in (c), (e), (f), respectively. For better visual representation, we scale the difference from $[-1,1]$ back to $[0,1]$, \ie{} no difference being gray. All baselines struggle to reconstruct the robotic arms, in particular their color, and the cube. }
    \label{fig:appendix_causalworld_baseline_reconstructions}
\end{figure}

\begin{figure}[t!]
    \centering
    \includegraphics[width=0.7\textwidth]{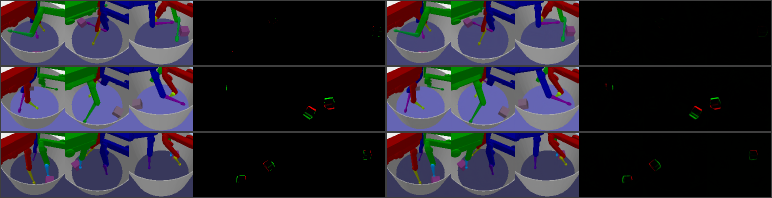}
    \caption{Reconstructions of BISCUIT on the CausalWorld dataset. \textbf{Left} is shown the ground truth observation of the environment, and \textbf{right} the reconstruction of the autoencoder in \OurApproach{}-NF. \OurApproach{} can reconstruct the images accurately up to minor smoothing artifacts.}
    \label{fig:appendix_causalworld_biscuit_reconstructions}
\end{figure}

\begin{figure}[t!]
    \centering
    \includegraphics[width=0.55\textwidth]{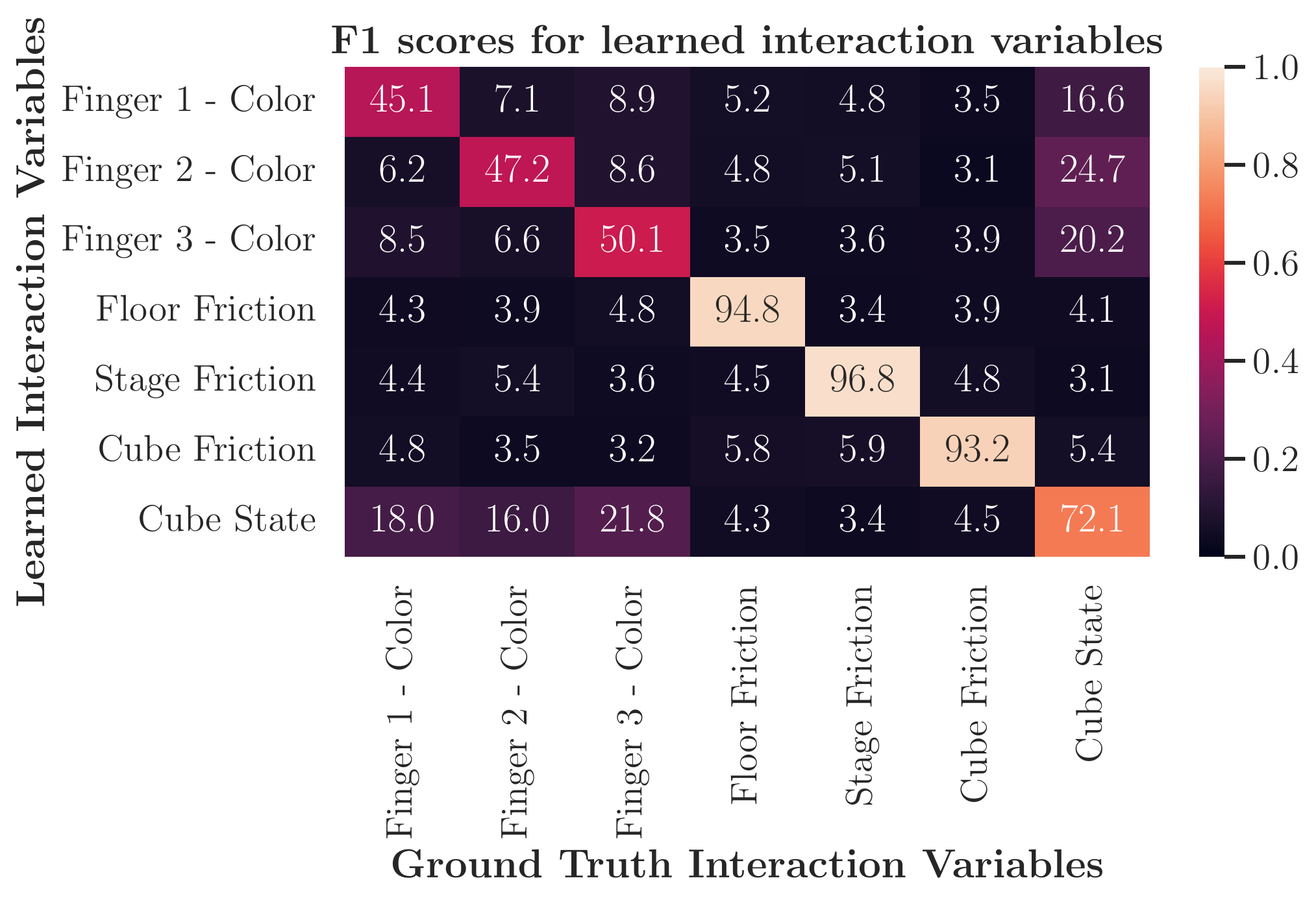}
    \caption{F1 scores between the learned interaction variables by \OurApproach{} and ground truth interactions on the CausalWorld dataset. \OurApproach{} closely models the interactions between the causal variables.}
    \label{fig:appendix_causalworld_f1_interaction_vars}
\end{figure}

\paragraph{Reconstructions} The identification of the baselines suffers from the poor reconstruction of the models.
We show an example of the reconstructions in \cref{fig:appendix_causalworld_baseline_reconstructions}.
In general, all baselines miss the cube as well as the colors of the tips. 
This is because the robotic arms and the cube move over time and appear at different positions in different frames, requiring the VAE to be accurately modeling their positions before learning the colors.
However, the gradient signal is often too small to overcome the KL regularizer on the latent space. 
Common tricks like using a KL scheduler did not show to improve the results.
In comparison, \OurApproach{}-NF can accurately reconstruct the images in \cref{fig:appendix_causalworld_biscuit_reconstructions}.
Since it uses an autoencoder with an unregularized latent space, it is much easier for the model to map all information in the latent space.

\paragraph{Interaction Variables} To analyze the learned interaction variables in \OurApproach{}, we recorded in the simulator the time steps on the test set in which there is a collision between an arm and the cube, and convert it to a binary signal (0 - no collision, 1 - collision). 
After training \OurApproach{}, we compare the learned binary interaction variables to the recorded collisions in terms of F1 score, similar to the Voronoi experiments.
We follow a similar procedure for the remaining causal variables as well.
We plot the F1 matrix of learned vs ground truth interactions in \cref{fig:appendix_causalworld_f1_interaction_vars}.
The learned interaction variables for each arm have an F1 score of about 50\%.
Since collisions only happen in approx. 5\% of the frames, a score of 50\% indicates a high similarity between the learned interaction and the ground truth collisions.
The mismatches are mostly due to the learned interaction being more conservative, i.e. being 1 already a frame too early sometimes.

\subsection{iTHOR}
\label{sec:appendix_ithor}

\begin{figure}[t!]
    \centering
    \footnotesize
    \begin{tabular}{rcccc}
        \textbf{Frames} & \includegraphics[align=c,width=0.18\textwidth]{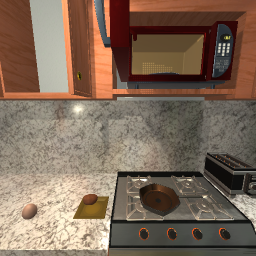} & 
        \includegraphics[align=c,width=0.18\textwidth]{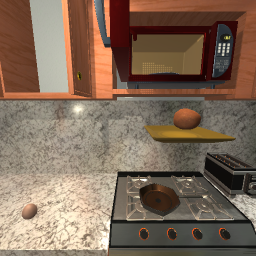} & 
        \includegraphics[align=c,width=0.18\textwidth]{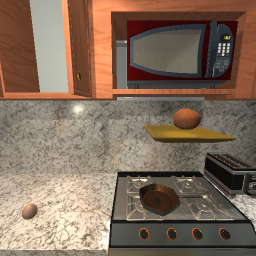}  & 
        \includegraphics[align=c,width=0.18\textwidth]{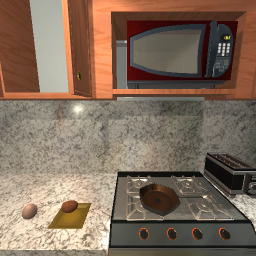} \\
        & & & & \\[-2mm]
        \textbf{Performed action} & - & PickupObject - Plate & CloseObject - Microw. & PutObject - Counter \\
        \textbf{Time step} & $t=0$ & $t=1$ & $t=2$ & $t=3$ \\
    \end{tabular}
    \caption{Example sequence in the iTHOR dataset \cite{kolve2017ai}. At each time step, we perform one action in the environment and use a randomly sampled pixel location of the interacted object as \robotstate{}. The resolution of each frame is $256\times 256\times 3$.}
    \label{fig:appendix_ithor_seq}
\end{figure}

\subsubsection{Dataset Setup}
\label{sec:appendix_ithor_dataset_setup}

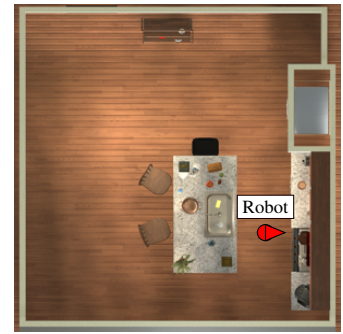
\begin{wrapfigure}{r}{0.25\textwidth}
    \centering
    \vspace{-2mm}
    \begin{tikzpicture}
        \node at (0,0) {\includegraphics[width=\linewidth]{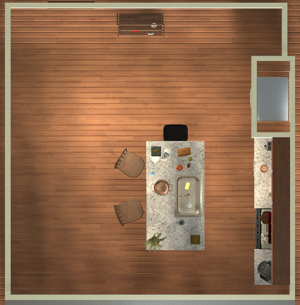}};
        \draw[draw=black,fill=red] (0.27\linewidth,-0.2\linewidth) circle (0.025\linewidth);
        \coordinate (r0) at (0.27\linewidth,-0.175\linewidth);
        \coordinate (r1) at (0.27\linewidth,-0.225\linewidth);
        \coordinate (r2) at (0.33\linewidth,-0.2\linewidth);
        \filldraw[draw=black,fill=red] (r0) -- (r1) -- (r2) -- cycle;
        \node[draw=black!90!red,fill=white,inner sep=2pt] (RB) at (0.27\linewidth,-0.12\linewidth) {\scriptsize Robot};
    \end{tikzpicture}
    \caption{The floor plan of the \texttt{FloorPlan10} environment \cite{kolve2017ai}. The robot's position and orientation are shown in red.}
    \label{fig:appendix_ithor_floor_plan}
    \vspace{-2mm}
\end{wrapfigure}

The iTHOR dataset \cite{kolve2017ai} is based on the \texttt{FloorPlan10} environment, which is the default kitchen environment.
We show the overall floor plan in \cref{fig:appendix_ithor_floor_plan}.
In it, we position the robot in front of the kitchen counter, and keep its position fixed.
As a first step in the environment, we place two movable objects (a plate with a potato and an egg) randomly on the counter, as well as the pan on the stove.
We remove all remaining movable objects from the robot's view.
Then, at each time step, we perform a randomly chosen action on one of the objects.
An overview of all objects and actions is shown in \cref{tab:appendix_ithor_object_list}.
Note that not all actions are always possible.
For instance, objects can only be opened when they are closed, and vice versa.
Further, we can perform the action \texttt{ToggleObject} on the microwave only if the microwave is closed, and the action \texttt{OpenObject} when the microwave is turned off.
For the movable objects, we can only pick up one of the two objects at once.
When an object is picked up, we can interact with a remaining object on the counter with the action \texttt{MoveObject} which moves it to a new random position on the counter.
The action \texttt{PutObject} randomly chooses one of the available receptacles to put the object down on.
For the plate, this includes the counter and the microwave if it is open.
For the egg, this includes the counter and the pan.
If the egg is put into the pan, it is automatically broken and cannot be picked up or moved anymore.
When any object is put down to the counter, we randomly sample a location on the counter which does not overlap with any other object on the counter.
Besides these interactions, we add the \texttt{NoOp} action, which does not interact with any object and represents the observational regime.

\begin{table}[t!]
    \centering
    \caption{List of the objects in the iTHOR environment with their corresponding possible actions to performed on.}
    \label{tab:appendix_ithor_object_list}
    \footnotesize
    \begin{tabular}{rl}
        \toprule
        \textbf{Objects} & \textbf{Possible actions} \\
        \midrule
        Plate & PickupObject, PutObject, MoveObject \\
        Egg & PickupObject, PutObject, MoveObject \\
        Cabinet & OpenObject, CloseObject \\
        Microwave & OpenObject, CloseObject, ToggleObject \\
        Toaster & ToggleObject \\
        Stove-Knob (4$\times$) & ToggleObject \\
        \bottomrule
    \end{tabular}
\end{table}

\begin{figure}[t!]
    \centering
    \footnotesize
    \begin{tabular}{rccccc}
        \textbf{Frames} & \includegraphics[align=c,width=0.13\textwidth]{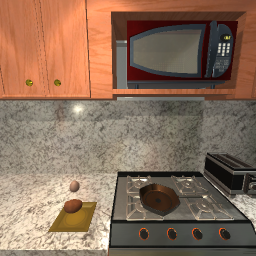} & 
        \includegraphics[align=c,width=0.13\textwidth]{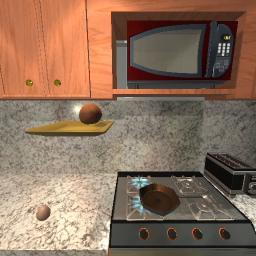} & 
        \includegraphics[align=c,width=0.13\textwidth]{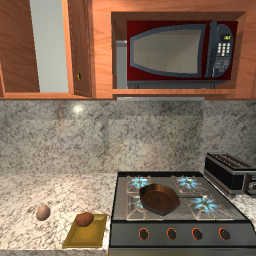} & 
        \includegraphics[align=c,width=0.13\textwidth]{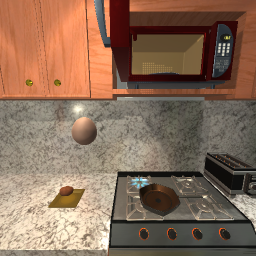}  & 
        \includegraphics[align=c,width=0.13\textwidth]{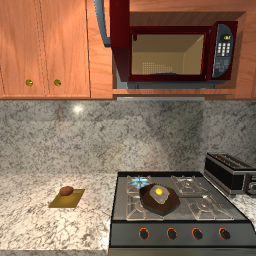} \\
        & & & & \\[-2mm]
        \textbf{Interaction maps} & \includegraphics[align=c,width=0.13\textwidth]{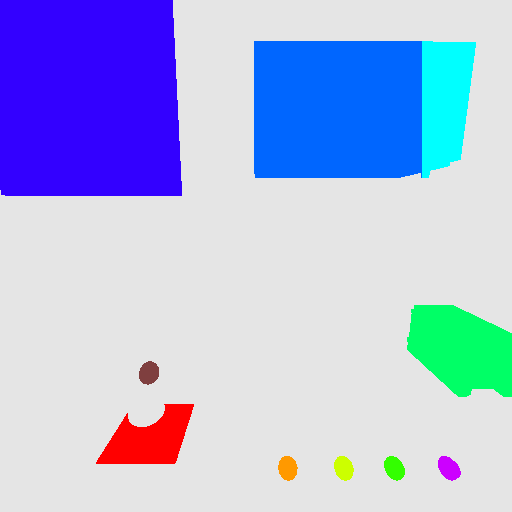} & 
        \includegraphics[align=c,width=0.13\textwidth]{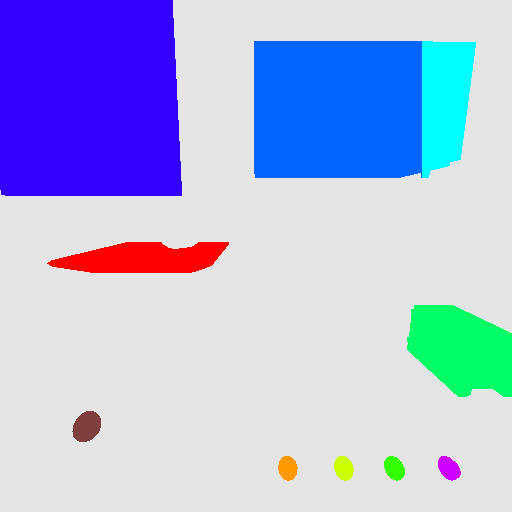} & 
        \includegraphics[align=c,width=0.13\textwidth]{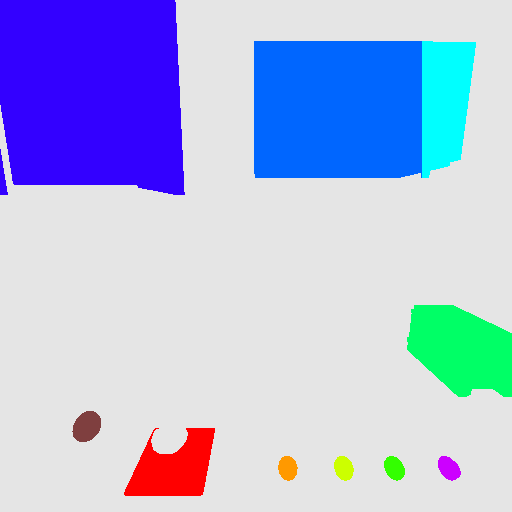} & 
        \includegraphics[align=c,width=0.13\textwidth]{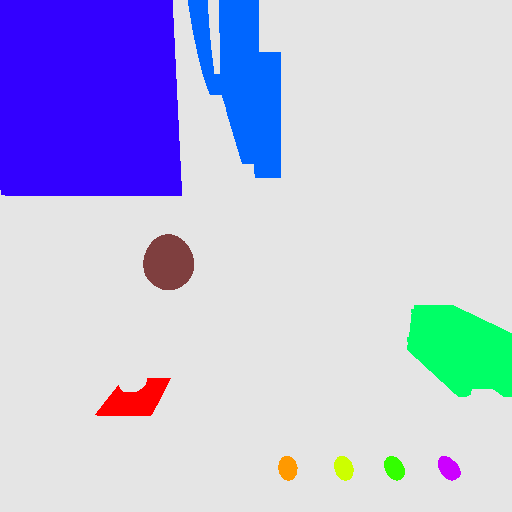} & 
        \includegraphics[align=c,width=0.13\textwidth]{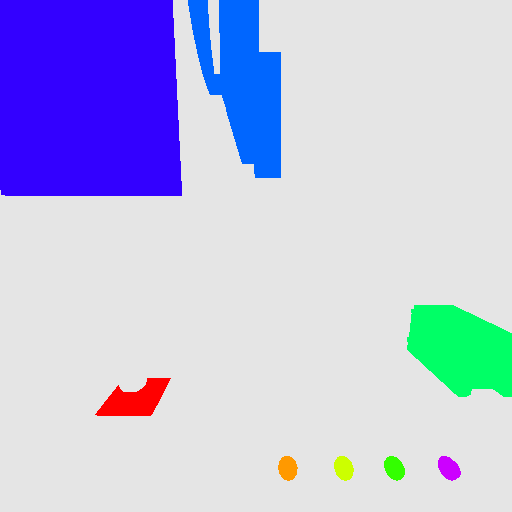} \\
    \end{tabular}
    \caption{Example ground truth interaction maps in the iTHOR dataset \cite{kolve2017ai}. The colors are aligned with the predicted interaction maps of \OurApproach{} in \cref{fig:experiments_ithor_interaction_maps}.}
    \label{fig:appendix_ithor_true_interaction_maps}
\end{figure}

The \robotstate{} $R^t\in[0,1]^2$ represents the click-location of a user on the image to select which object to interact with.
Specifically, after choosing the action-object pair to perform at a time step, we use the object segmentation mask of the iTHOR simulator to identify the set of pixels that show the object in question.
We then randomly sample one of these pixels, and set $R^t$ as the location of this pixel in the frame.
After that, the actual action is performed.
As location for the \texttt{NoOp} action, we sample a pixel location which does not belong to any object in the current frame.
For the microwave, we split the object in two halves, where the left part, \ie{} the door, performs the action \texttt{OpenObject}, and the right part, \ie{} the buttons and display, activates the microwave.
Additionally, when the microwave is open, we sample the pixel location from the open door.
Finally, the stoves are controlled by the knobs, such that their interactions are sampled from the knobs positions.
Examples of the interaction maps are shown in \cref{fig:appendix_ithor_true_interaction_maps}.

The causal variables in this environment align with the action-object pairs at hand. 
For each object that has the action \texttt{ToggleObject}, we have a binary causal variable indicating whether it is active or not.
Similarly, for each object that has the action \texttt{OpenObject}, we have a binary causal variable indicating whether it is open or not.
For each object that has the action \texttt{PickupObject}, we have one binary causal variable indicating whether it is picked up or not, and three causal variables for the object's x-y-z position in the 3D environment.
Additionally, for the egg, we have two binary causal variables indicating whether it is broken, \ie{} in the pan, and cooked, which happens instantaneously when the stove is being turned on and the egg is in the pan.
This results in overall 18 causal variables: 
\begin{compactitem}
    \item Cabinet-Open
    \item Egg-Broken, Egg-Pos-x, Egg-Pos-y, Egg-Pos-z, Egg-Cooked, Egg-PickedUp
    \item Microwave-Open, Microwave-Active
    \item Plate-Pos-x, Plate-Pos-y, Plate-Pos-z, Plate-PickedUp
    \item Stove1-Active, Stove2-Active, Stove3-Active, Stove4-Active
    \item Toaster-Active
\end{compactitem}

\begin{wrapfigure}{r}{0.26\textwidth}
    \centering
    \vspace{-2mm}
    \begin{tabular}{cc}
        \begin{tikzpicture}
            \clip (0.1\textwidth,-0.12\textwidth) rectangle + (0.1\textwidth, 0.1\textwidth);
            \node at (0,0) {\includegraphics[width=0.4\textwidth]{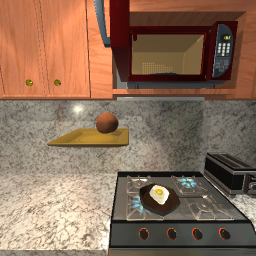}};
        \end{tikzpicture} & 
        \begin{tikzpicture}
            \clip (0.1\textwidth,-0.12\textwidth) rectangle + (0.1\textwidth, 0.1\textwidth);
            \node at (0,0) {\includegraphics[width=0.4\textwidth]{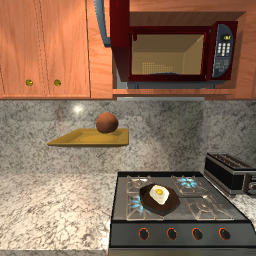}};
        \end{tikzpicture} \\
        Toaster - Off & Toaster - On \\
    \end{tabular}   
    \caption{The Toaster being turned off or on changes very few pixels in the observation.}
    \label{fig:appendix_ithor_toaster_active}
    \vspace{-2mm}
\end{wrapfigure}

We generate the frames with a resolution of $512\times 512\times 3$, and reduce the dimension to $256\times 256\times 3$ via bi-linearly interpolation afterward.
The high resolution is required since some states are only shown by fine details in the image.
For instance, \cref{fig:appendix_ithor_toaster_active} shows that the difference between the toaster being active or not is only a few pixels.
In the meantime, actions like the cabinet being opened change about $10\%$ of the image.
This makes it challenging for reconstruction-based methods to fully capture all variables fairly in the latent space.
Besides, the environment uses a 3D physics engine, showing some interactions that we are not able to fully capture by the causal variables defined.
For instance, when the stove is turned on, the flame slowly grows over time and slightly fluctuates once it reached its maximum.
Further, when the egg is broken, it slides into the pan over the next three frames.

We generate 1500 sequences of each 100 frames, and 250 sequences for testing. Examples are shown in \cref{fig:appendix_ithor_seq}.

\subsubsection{Hyperparameters and Implementation Details}
\label{sec:appendix_ithor_hyperparams}

\paragraph{\OurApproach{}}
We slightly adapt the setup from the experiments of the CausalWorld environment by:
\begin{compactenum}
    \item we reduce the channel size to 64 and batch size to 64 due to the higher resolution;
    \item we increase the latent dimension to 40 due to the larger number of causal variables;
    \item we reduce the number of epochs to 150 for the autoencoder;
    \item we reduce the learning rate to 2e-4;
    \item we do not use $R^t$ as input to the decoder.
\end{compactenum}
In general, we did not find \OurApproach{} to be particularly hyperparameter sensitive in these experiments.

\paragraph{Baselines}
We adapt the experimental setup of the baselines in the same way as \OurApproach{}.
We set the learning rate to 2e-4 for all baselines, which was more stable across models.

\subsubsection{Results}
\label{sec:appendix_ithor_results}

\begin{figure}[t!]
    \centering
    \begin{subfigure}[b]{0.48\columnwidth}
        \centering
        \includegraphics[width=\textwidth]{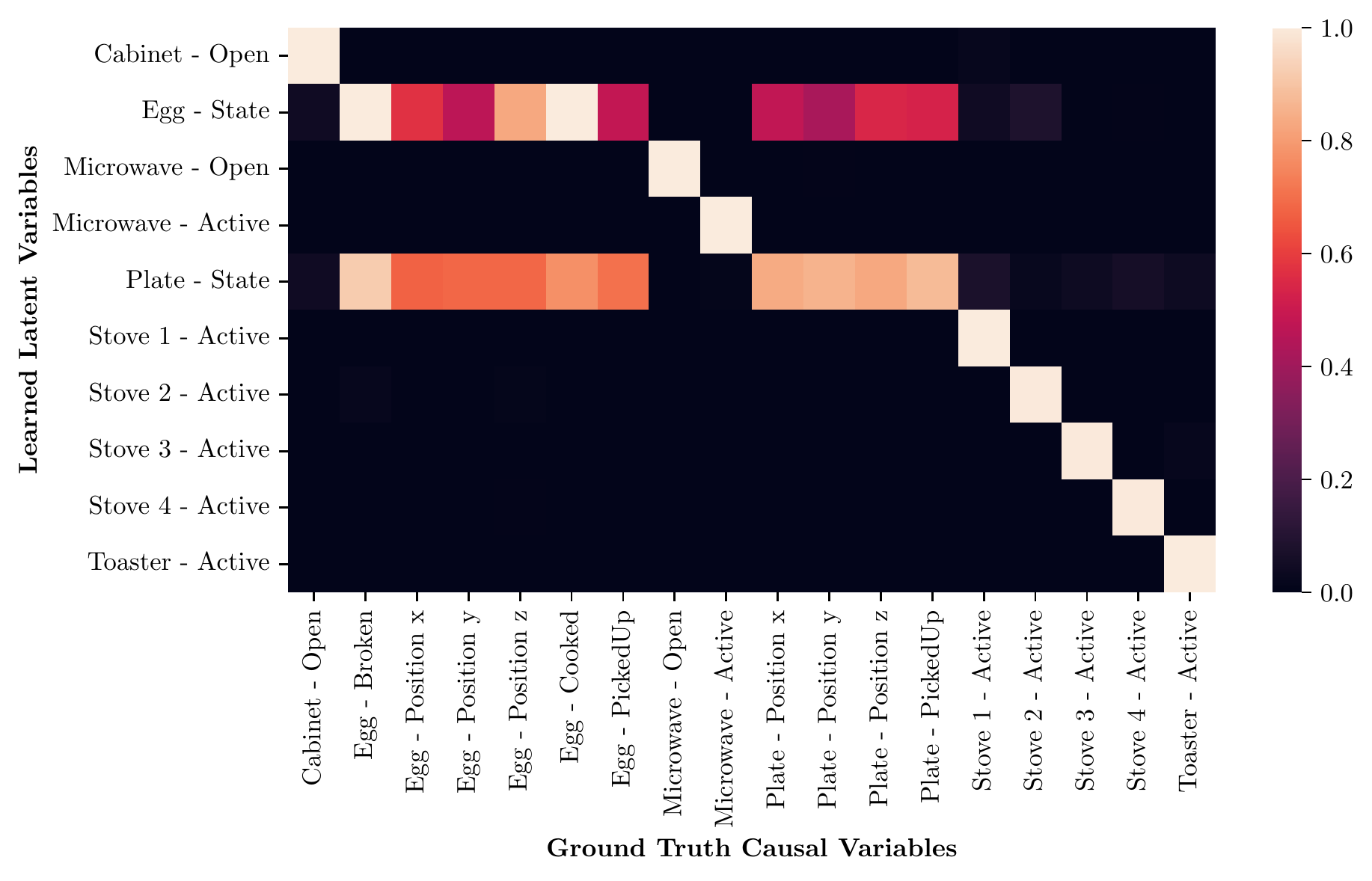}
        \caption{\OurApproach{}}
    \end{subfigure}
    \hfill
    \begin{subfigure}[b]{0.48\columnwidth}
        \centering
        \includegraphics[width=\textwidth]{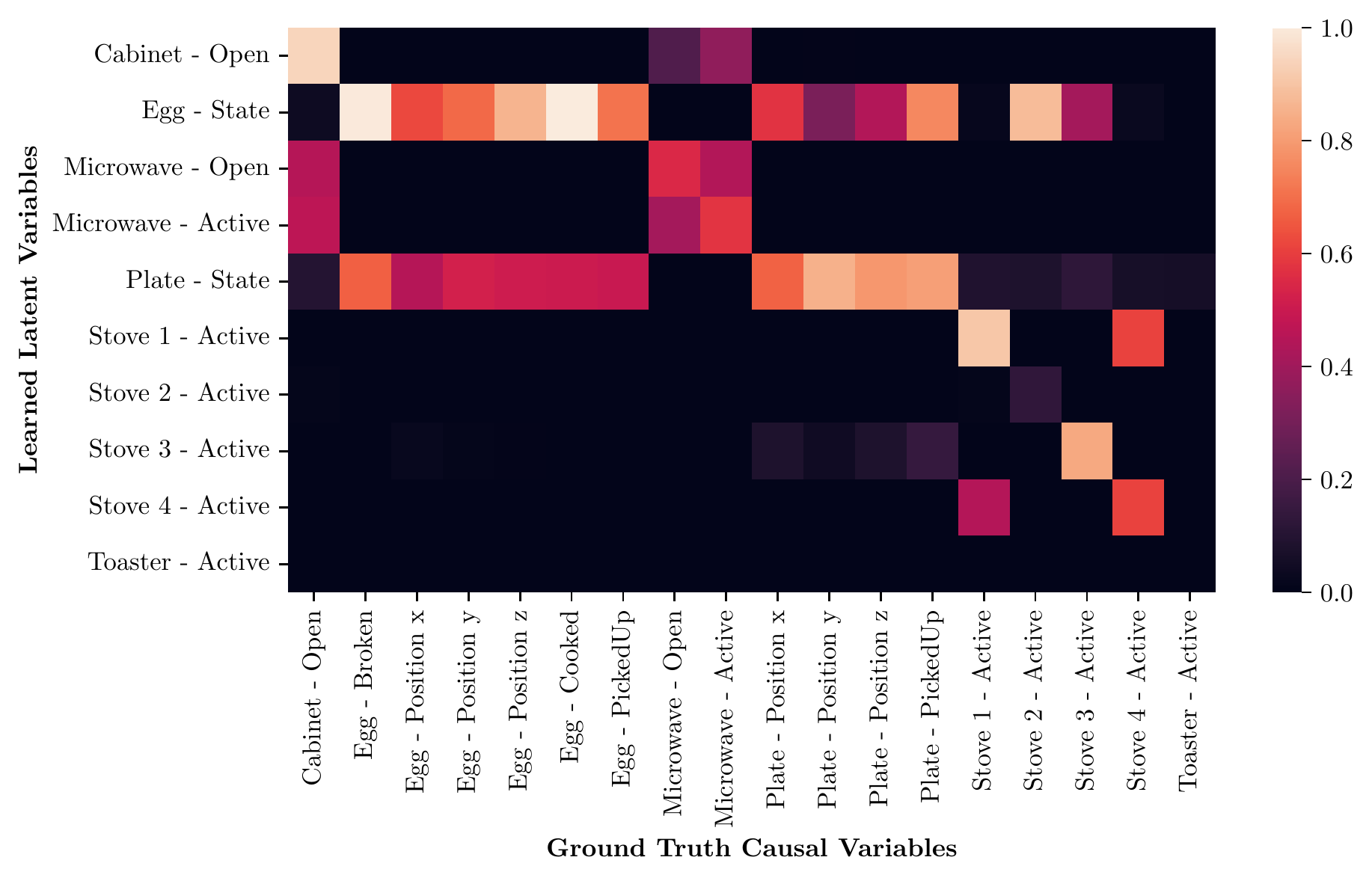}
        \caption{DMS \cite{lachapelle2021disentanglement}}
    \end{subfigure}
    \begin{subfigure}[b]{0.48\columnwidth}
        \vspace{5mm}
        \centering
        \includegraphics[width=\textwidth]{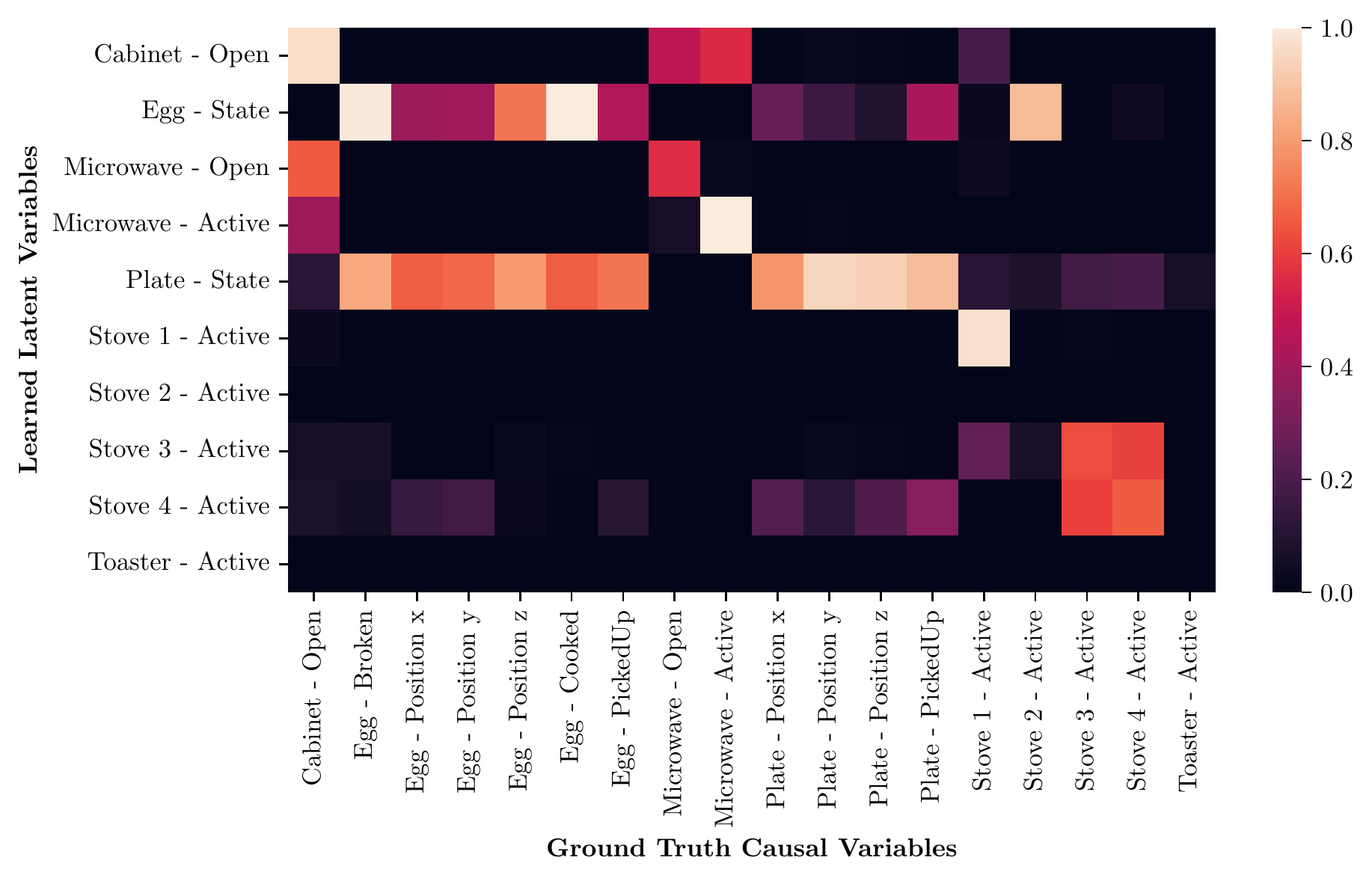}
        \caption{LEAP \cite{yao2021learning}}
    \end{subfigure}
    \hfill
    \begin{subfigure}[b]{0.48\columnwidth}
        \vspace{5mm}
        \centering
        \includegraphics[width=\textwidth]{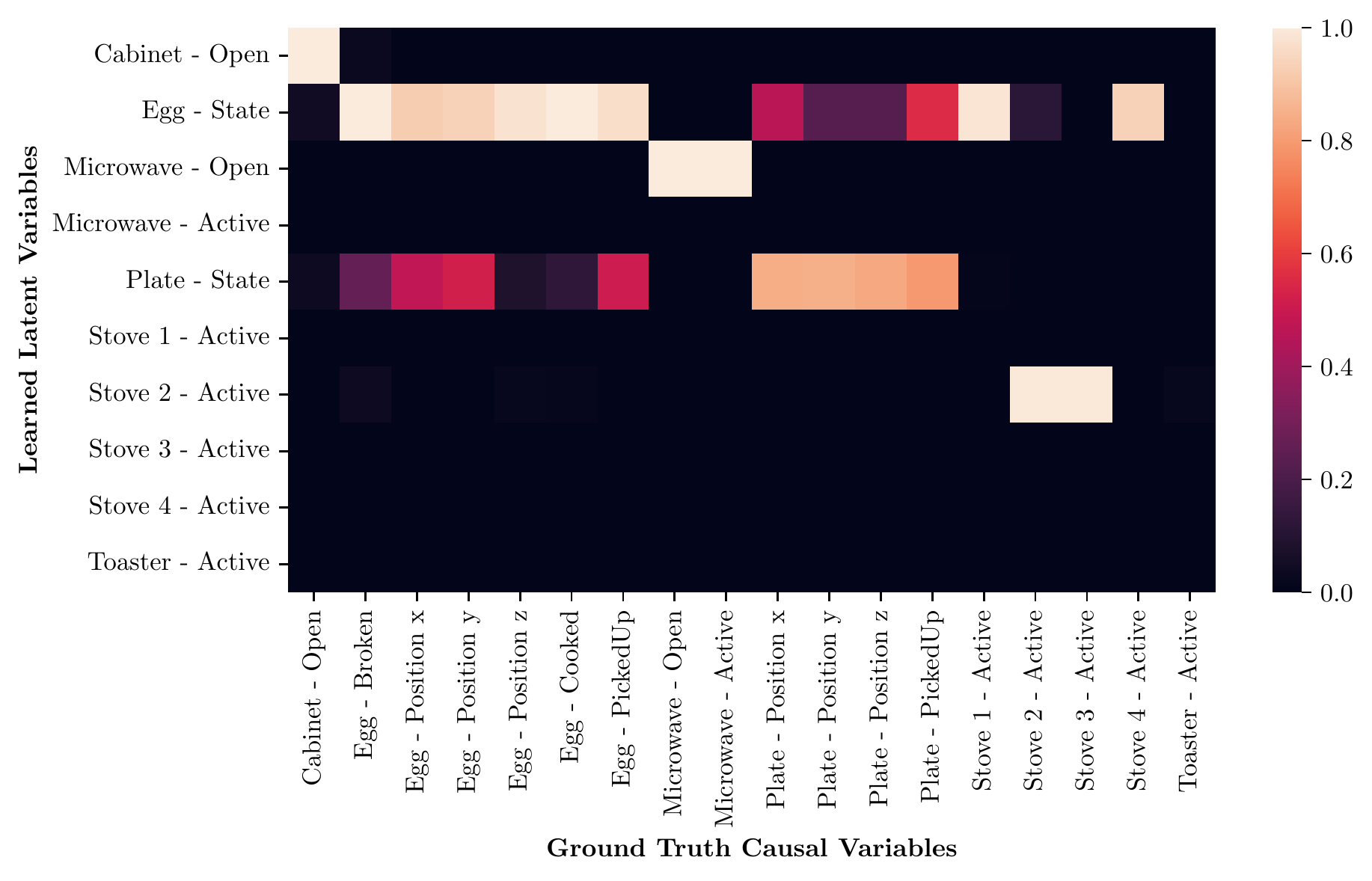}
        \caption{iVAE \cite{khemakhem2020variational}}
    \end{subfigure}
    \caption{$R^2$ matrix for learned models on the iTHOR environment \cite{kolve2017ai}: (a) \OurApproach{} (ours), (b) DMS \cite{lachapelle2021disentanglement}, (c) LEAP \cite{yao2021learning}, (d) iVAE \cite{khemakhem2020variational}.}
    \label{fig:appendix_ithor_r2_matrix}
\end{figure}

\begin{table}[t!]
    \centering
    \caption{Result table for the iTHOR experiments over 3 seeds, with standard deviation listed next to the mean.}
    \label{tab:appendix_ithor_results}
    \footnotesize
    \begin{tabular}{ccccc}
        \toprule
        \textbf{Method} & \textbf{$\bm{R^2}$-diag} & \textbf{$\bm{R^2}$-sep} & \textbf{Spearman-diag} & \textbf{Spearman-sep} \\
        \midrule
        \OurApproach{} & $0.96 \pm 0.01$ & $0.15 \pm 0.01$ & $0.96 \pm 0.00$ & $0.22 \pm 0.02$\\
        DMSVAE & $0.61 \pm 0.05$ & $0.40 \pm 0.04$ & $0.65 \pm 0.07$ & $0.48 \pm 0.05$ \\
        LEAP & $0.63 \pm 0.04$ & $0.45 \pm 0.04$ & $0.65 \pm 0.06$ & $0.49 \pm 0.03$ \\
        iVAE & $0.48 \pm 0.04$ & $0.35 \pm 0.02$ & $0.48 \pm 0.04$ & $0.38 \pm 0.02$ \\
        \bottomrule
    \end{tabular}
\end{table}

\paragraph{Correlation Evaluation}
\cref{tab:appendix_ithor_results} supplements the results of \cref{tab:experiments_causalworld_ithor} in the main text by including standard deviations over three seeds, as well as the Spearman scores on the same variables.
For the multidimensional causal variables 'Egg state' and 'Plate state', where each dimension is highly correlated and shares the same interactions, we follow \citet{lippe2022citris} by averaging the variable's dimensions in the $R^2$-diag calculation before taken the diagonal average.
An example $R^2$ matrix for each method is shown \cref{tab:appendix_ithor_results}.
\OurApproach{} identifies and separates all causal variables up to the two movable objects (plate and egg).
Meanwhile, for both DMS and LEAP, we find that the models diversely entangle the causal variables.
The only causal variable that has not been modeled at all is the toaster, which is likely due to its small pixel footprint in the image (see \cref{fig:appendix_ithor_toaster_active}).
Interestingly, iVAE showed a very different trend.
While it has less diverse entanglement than LEAP and DMS, it often models several binary causal variables in the same latent dimension.
Since its latent variables are all conditioned on full $R^t$, there is no difference to model two binary variable in one or two separate dimensions. 
In contrast, this does not happen in \OurApproach{} since these different binary causal variables have different binary interaction variables.

\begin{figure}[t!]
    \centering
    \begin{tabular}{ccc}
        \textbf{Original image} & \textbf{Overlapped image} & \textbf{Interaction map} \\
        \includegraphics[width=0.2\textwidth]{figures/experiments/ithor/interaction_maps/ithor_hard_vq_matches_5_orig_img.png} &
        \includegraphics[width=0.2\textwidth]{figures/experiments/ithor/interaction_maps/ithor_hard_vq_matches_5_augmented.png} &
        \includegraphics[width=0.2\textwidth]{figures/experiments/ithor/interaction_maps/ithor_hard_vq_matches_5.png} \\
        \includegraphics[width=0.2\textwidth]{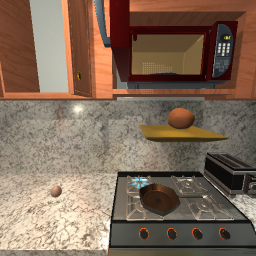} &
        \includegraphics[width=0.2\textwidth]{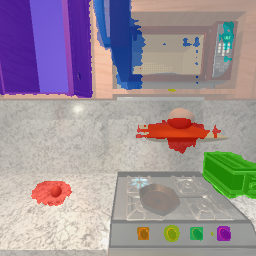} &
        \includegraphics[width=0.2\textwidth]{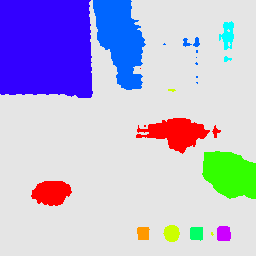} \\
        \includegraphics[width=0.2\textwidth]{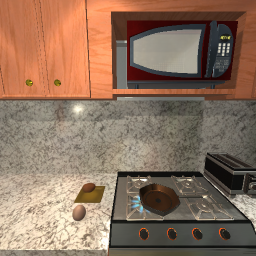} &
        \includegraphics[width=0.2\textwidth]{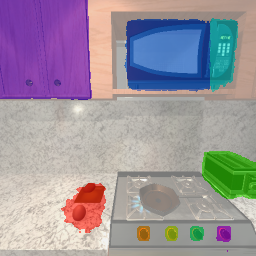} &
        \includegraphics[width=0.2\textwidth]{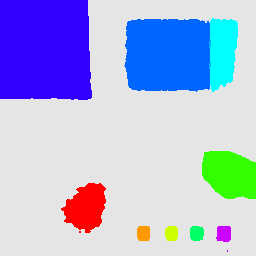} \\
        \includegraphics[width=0.2\textwidth]{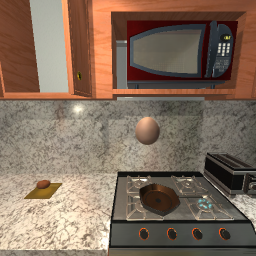} &
        \includegraphics[width=0.2\textwidth]{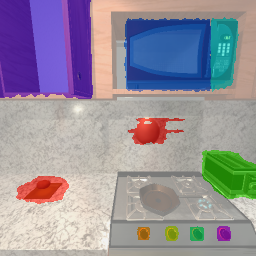} &
        \includegraphics[width=0.2\textwidth]{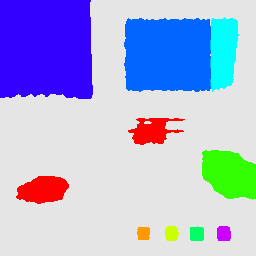} \\
    \end{tabular}
    \caption{Learned interaction maps of \OurApproach{} on iTHOR \cite{kolve2017ai}. The images on the left constitute the input image, the right image the learned interaction variables, and the center the overlap of both. The colors correspond to the following causal variables modeled by the respective latent variable: \textcolor{red}{red} $\to$ plate/egg state, \textcolor{green}{light green} $\to$ Toaster-Active, \textcolor{cyan}{light blue} $\to$ Microwave-Active, \textcolor{blue}{blue} $\to$ Microwave-Open, \textcolor{black!50!blue}{dark blue} $\to$ Cabinet-Open, \textcolor{orange}{orange} $\to$ Stove1-Active, \textcolor{yellow}{yellow} $\to$ Stove2-Active, \textcolor{black!50!green}{green} $\to$ Stove3-Active, \textcolor{purple}{purple} $\to$ Stove4-Active.}
    \label{fig:appendix_ithor_interaction_maps}
\end{figure}

\begin{figure}[t!]
    \centering
    \begin{tabular}{cccc}
        \textbf{Input image 1} & \textbf{Input image 2} & \textbf{Generated Output} & \textbf{Latents from image 2} \\
        \includegraphics[align=c,width=0.2\textwidth]{figures/experiments/ithor/triplets/ithor_triplet_090.png} &
        \includegraphics[align=c,width=0.2\textwidth]{figures/experiments/ithor/triplets/ithor_triplet_099.png} & 
        \includegraphics[align=c,width=0.2\textwidth]{figures/experiments/ithor/triplets/rec_frame_11_23_090_099.png} &
        \begin{tabular}{c}
            Microwave Active \\
            Stove (front-left) \\
        \end{tabular}\\
        & & & \\[-2mm]
        \includegraphics[align=c,width=0.2\textwidth]{figures/experiments/ithor/triplets/ithor_triplet_090.png} &
        \includegraphics[align=c,width=0.2\textwidth]{figures/experiments/ithor/triplets/ithor_triplet_099.png} & 
        \includegraphics[align=c,width=0.2\textwidth]{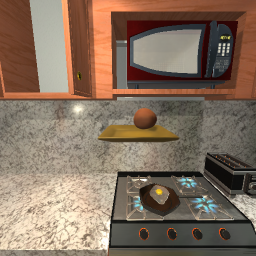} &
        \begin{tabular}{c}
            Stove (front-left) \\
        \end{tabular}\\
        & & & \\[-2mm]
        \includegraphics[align=c,width=0.2\textwidth]{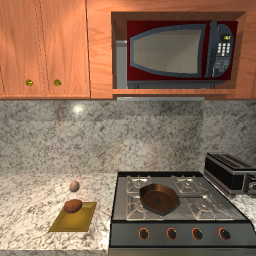} &
        \includegraphics[align=c,width=0.2\textwidth]{figures/experiments/ithor/triplets/ithor_triplet_090.png} & 
        \includegraphics[align=c,width=0.2\textwidth]{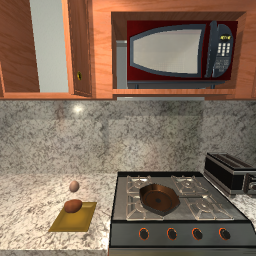} &
        \begin{tabular}{c}
            Microwave Active \\
            Cabinet Open \\
        \end{tabular}\\
        & & & \\[-2mm]
        \includegraphics[align=c,width=0.2\textwidth]{figures/experiments/ithor/triplets/ithor_triplet_000.png} &
        \includegraphics[align=c,width=0.2\textwidth]{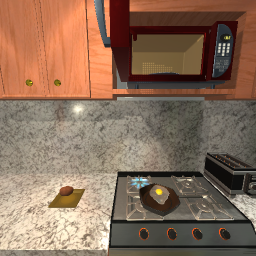} & 
        \includegraphics[align=c,width=0.2\textwidth]{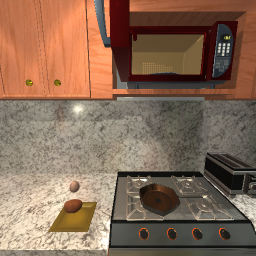} &
        \begin{tabular}{c}
            Microwave Open \\
        \end{tabular}\\
        & & & \\[-2mm]
        \includegraphics[align=c,width=0.2\textwidth]{figures/experiments/ithor/triplets/ithor_triplet_000.png} &
        \includegraphics[align=c,width=0.2\textwidth]{figures/experiments/ithor/triplets/ithor_triplet_090.png} & 
        \includegraphics[align=c,width=0.2\textwidth]{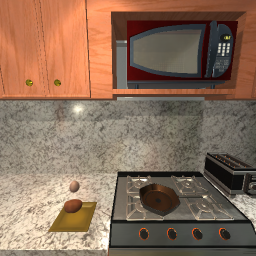} &
        \begin{tabular}{c}
            Toaster Active \\
        \end{tabular}\\
    \end{tabular}
    \caption{Additional triplet predictions of \OurApproach{} on images from the test dataset. The first two images are encoded in the latent space. We then replace the latents of the first image by the second image for the dimensions that correspond to the causal variable listed in the fourth column. The third column shows then the reconstruction of \OurApproach{} for this new latent vector.}
    \label{fig:appendix_ithor_triplet_generations}
\end{figure}

\paragraph{Interaction Maps}
Similarly to \cref{fig:experiments_ithor_interaction_maps} in the main text, we visualize more examples of the learned interaction maps of \OurApproach{} in \cref{fig:appendix_ithor_interaction_maps}.
The first column shows the input image, the third the learned binary interaction variables, and the second column the overlap of both.
One can clearly see how \OurApproach{} adapts its interaction variables to the input image.
For instance, in the second row, one can see how \OurApproach{} identifies the regions of the plate and the egg, visualized in red.
In line with the $R^2$ evaluation, \OurApproach{} maps the interactions with the plate and egg together into one interaction variable instead of separating them.
The interaction variables of the microwave change towards focusing on the opened door for the action \texttt{CloseObject} (blue).
Since it is not possible to activate the microwave at this state, its corresponding interaction variable (light blue) mostly disappears.
It does not fully disappear, since the model has never seen a click in this region when the microwave door is open.
Meanwhile, other interaction variables that are independent of the actual input image, \eg{} the stove knobs or the toaster, are also kept constant throughout the examples by \OurApproach{}.

The images only show 9 out of the 40 learned interaction maps.
The remaining 31 interaction variables are mostly either (1) constants and would give an empty interaction map, or (2) follow a very similar pattern to the red interaction variable, \ie{} the plate and the egg.
While we find that some of these dimension focus slightly more on the plate and some more on the egg, there was none that fully focused on either variable, which follows the insights of the $R^2$ evaluation.

\paragraph{Triplet Generations}
Similar to \cref{fig:experiments_ithor_triplet} in the main text, we visualize additional examples of triplet predictions of \OurApproach{} in \cref{fig:appendix_ithor_triplet_generations}.
The first two images represent the input images, and the third the generated output.
The fourth column specifies which causal variables we tried to replace in the first image by the latents of the second image.
\OurApproach{} consistently generates the correct counterfactual prediction for a variety of causal variables and input images.

\end{document}